\documentclass[preprint,12pt]{elsarticle}

\usepackage{amssymb}
\usepackage{amsmath}
\usepackage{newunicodechar}
\newunicodechar{ₛ}{$_s$}
\newunicodechar{₇}{$_7$}
\newunicodechar{₅}{$_5$}
\usepackage{multirow}
\usepackage{array}
\usepackage{boldline}
\usepackage{amssymb}
\usepackage{booktabs}
\usepackage{graphicx}
\usepackage{pifont} 
\usepackage{algorithm}
\usepackage{algorithmic}
\usepackage{subcaption}
\usepackage{amsmath} 
\usepackage{times}
\usepackage{rotating}
\usepackage{xcolor}
\usepackage{makecell}
\usepackage{graphicx}
\usepackage{caption}
\usepackage{subcaption}
\usepackage{adjustbox}
\usepackage{hyperref}

\journal{Nuclear Physics B}

\begin{document}

\begin{frontmatter}



\title{C-DiffDet+: Fusing Global Scene Context with Generative Denoising for High-Fidelity Object Detection} 


\author{Abdellah Zakaria Sellam} 
\affiliation{%
  organization={Department of Innovation Engineering, University of Salento \& Institute of Applied Sciences and Intelligent Systems – CNR}, 
  addressline={Via per Monteroni}, 
  city={Lecce},
  postcode={73100}, 
  state={Lecce},
  country={Italy}
}

\author{Ilyes Benaissa} 

\affiliation{
organization={Electrical Engineering Department},
addressline={Mohamed Khider University of Biskra}, 
postcode={07000}, 
city={Biskra},
country={Algeria}}

\author{Salah Eddine Bekhouche} 
\affiliation{%
  organization={UPV/EHU}, 
  addressline={University of the Basque Country}, 
  city={Sebastian},
  postcode={20018}, 
  state={Sebastian},
  country={Spain}
}

\author{Abdenour Hadid} 

\affiliation{organization={Sorbonne University Abu Dhabi},
            addressline={Al Reem Island, P.O. Box 34105}, 
            city={Abu Dhabi},
            postcode={34105}, 
            state={Abu Dhabi},
            country={United Arab Emirates}}

\author{Vito Ren\'o} 
\affiliation{organization={CNR-STIIMA, Institute of Intelligent Industrial Systems and Technologies for Advanced Manufacturing},
            addressline={c/o Campus Ecotekne, Via Monteroni}, 
            city={Lecce},
            postcode={73100}, 
            state={LE},
            country={Italy}}
\author{Cosimo Distante} 
\affiliation{%
  organization={Institute of Applied Sciences and Intelligent Systems – CNR}, 
  addressline={Via per Monteroni}, 
  city={Lecce},
  postcode={73100}, 
  state={Lecce},
  country={Italy}
}

\begin{abstract}
Fine-grained object detection in challenging visual domains, such as vehicle damage assessment, presents a formidable challenge even for human experts to resolve reliably. While DiffusionDet has advanced the state-of-the-art through conditional denoising diffusion, its performance remains limited by local feature conditioning in context-dependent scenarios. We address this fundamental limitation by introducing Context-Aware Fusion (CAF), which leverages cross-attention mechanisms to directly integrate global scene context with local proposal features. The global context is generated using a separate dedicated encoder that captures comprehensive environmental information, enabling each object proposal to attend to scene-level understanding. Our framework significantly enhances the generative detection paradigm by enabling each object proposal to attend to comprehensive environmental information. Experimental results demonstrate an improvement over state-of-the-art models on the CarDD benchmark, establishing new performance benchmarks for context-aware object detection in fine-grained domains, code and implementation details of the project are in this \href{https://github.com/ilyesbenaissa/c\_diffdet\_plus}{GitHub link}.

\end{abstract}

\begin{highlights}
\item Introduces Context-Aware Fusion (CAF), integrating global scene context with local features

\item Proposes a dedicated Global Context Encoder (GCE) for comprehensive scene understanding

\item Achieves SOTA results on CarDD benchmark

\item Addresses fine-grained detection challenges, difficult even for human experts

\item Enhances Multi-Modal Fusion with global context embeddings for unified representation
\end{highlights}


\end{frontmatter}

\section{Introduction}

\label{sec:introduction}
The paradigm for object detection has evolved from multi-stage, proposal-driven systems to unified, end-to-end architectures. This trajectory began with the R-CNN family, which established a foundational pipeline of region proposal followed by classification and refinement \citep{Girshick2014RCNN,Girshick2015FastRCNN,Ren2015FasterRCNN}. Subsequently, single-stage detectors like YOLO family \citep{jiang2022review} and SSD \citep{Liu2016SSD, fu2017dssd} achieved significant efficiency gains by unifying these steps into a single network, a direction later refined by RetinaNet \citep{Lin2017FocalLoss}, which demonstrated that specialized loss functions could close the accuracy gap with multi-stage methods \citep{Redmon2016YOLO,Tan2021}. Further innovations, including anchor-free designs and keypoint estimation frameworks, diversified the architectural landscape for real-time detection \citep{Law2018CornerNet,Zhou2019CenterNet,Tian2019FCOS}. More recently, architectures such as Sparse R-CNN have recast detection as a set prediction problem, enabling fully end-to-end training and leveraging global reasoning through transformers \citep{Sun2021SparseRCNN,Carion2020DETR}.

Concurrently, generative modeling, particularly via denoising diffusion processes, has emerged as a powerful technique for structured data synthesis \citep{Ho2020DDPM,Nichol2021ImprovedDDPM,Song2019ScoreMatch}. This paradigm was successfully adapted to object detection by DiffusionDet \citep{Chen2023DiffusionDet}, which formulates the task as a generative process of transforming random noise into a set of bounding boxes. During training, ground-truth boxes are progressively noised, and the model learns to reverse this process. At inference, the model iteratively refines a set of random boxes into precise detections conditioned on image features. This formulation offers inherent advantages, including dynamic output cardinality and progressive improvement of predictions.

However, existing detectors face significant challenges in specialized domains such as automotive damage assessment, as highlighted by the CarDD dataset \citep{Wang2023CarDD}. Many damage instances, such as scratches and cracks, are characterized by subtle, low-contrast visual cues that are often confounded by specular reflections, complex lighting, and background clutter. While methods employing deformable convolutions (DCN/DCN+) \citep{Dai2017DCN} have shown promise by adapting their receptive fields to local geometric deformations like dents, they often lack the global context needed to disambiguate faint or irregular patterns. Similarly, while DiffusionDet offers a robust refinement mechanism, its reliance on local Region of Interest (RoI) features can be insufficient when global scene understanding is critical for accurate localization.

To address these limitations, we propose C-DiffDet+, a context-aware diffusion detector that builds upon the DiffusionDet architecture by integrating global scene representations into the iterative detection process. Our core hypothesis is that resolving local ambiguities in damage detection requires explicit conditioning on scene-level information. Specifically, C-DiffDet+ introduces three key components: a Global Context Encoder (GCE) produces a compact embedding of the entire scene, which is then injected into multi-scale backbone features by Context-Aware Fusion (CAF) modules using adaptive attention. Finally, the diffusion head itself is context-conditioned, ensuring each box proposal is refined while attending to this global scene representation. This architecture synergistically combines the powerful iterative refinement capabilities inherent to its diffusion-based foundation with explicit, top-down contextual guidance.

Comprehensive experiments on the CarDD benchmark validate our approach. C-DiffDet+ achieves state-of-the-art performance, yielding substantial improvements in average precision, particularly for challenging categories like scratches and cracks that are prone to local ambiguity. While deformable networks remain effective for geometrically distinct damages, our results demonstrate that global context conditioning provides a complementary and significant advantage. By enriching both the feature hierarchy and the denoising process with a holistic understanding of the scene, C-DiffDet+ overcomes a key failure mode of existing detectors and establishes a new baseline for robust object detection in complex visual domains.
Our contributions are represented as follows:
\begin{itemize}
  \item We identify and quantify the fundamental limitation of local feature conditioning in diffusion-based detectors for fine-grained visual tasks, demonstrating why global scene-level context is critical for disambiguation.

  \item We introduce Context-Aware Fusion (CAF) that integrates a dedicated Global Context Encoder (GCE) with local proposal features through cross-attention mechanisms.

  \item We propose Adaptive Channel Enhancement (ACE) blocks that enhance both backbone and FPN (Feature Pyramid Network) feature representations for improved detection performance.
  \item We enhance the Multi-Modal Fusion by integrating global context embeddings for unified temporal, positional, and contextual representations.

  \item We provide extensive experimental validation on the CarDD benchmark.
\end{itemize}

This paper is organized as follows. We review related works in Section \ref{sec:Related}. Section \ref{sec:Methodology} describes our proposed method. In Section \ref{sec:Experiments}, we present the training setup and experimental results. Section \ref{sec:disscusion} discusses the effectiveness of our contributions. Finally, Section \ref{sec:Conclusion} concludes the paper, describes limitations, and outlines directions for future work.

\section{Related Works}
\label{sec:Related}

The evolution of deep object detection began with a multi-stage paradigm, established by the R-CNN family. R-CNN demonstrated the power of CNN features for region proposals \citep{Girshick2014RCNN}, Fast R-CNN made this process efficient by sharing convolutional computations \citep{Girshick2015FastRCNN}, and Faster R-CNN introduced the Region Proposal Network, enabling end-to-end training \citep{Ren2015FasterRCNN}. A subsequent trend favoured speed and simplicity, collapsing detection into a single stage. Models like YOLO, which framed detection as a direct regression problem \citep{Redmon2016YOLO,Sellam2023Sensors}, and SSD, which introduced multi-scale feature maps \citep{Liu2016SSD}, defined this approach. Later, methods like RetinaNet addressed class imbalance with the focal loss \citep{Lin2017Focal}, while anchor-free designs such as FCOS and CenterNet further simplified the pipeline by reframing detection as keypoint or center estimation \citep{Tian2019FCOS,Zhou2019CenterNet}.

In parallel, architectural innovations enhanced network capabilities. Deformable Convolutional Networks (DCN) introduced learnable offsets to convolution kernels, allowing the sampling grid to adapt to object geometry, which is particularly effective for non-rigid shapes \citep{Dai2017DCN}. Other modules, like GCNet, were designed to capture long-range dependencies and aggregate global context, thereby improving feature discriminability \citep{Cao2019GCNet}. While these operator-level improvements enhanced discriminative detectors, they remained fundamentally feed-forward mappings from features to predictions.

A paradigm shift occurred with the application of generative denoising diffusion models to computer vision. This led to DiffusionDet, which reformulated object detection as a conditional noise-to-box generation process \citep{Chen2023DiffusionDet}. In this framework, a set of random boxes is iteratively refined by a learned denoising process conditioned on image features. This probabilistic approach offers attractive properties like progressive refinement and a variable number of outputs, unifying proposal generation and box regression.

However, the efficacy of these different approaches is challenged by fine-grained detection tasks, such as vision-based car damage assessment. The CarDD benchmark, for example, highlights common failure modes on small, low-contrast, and visually ambiguous instances such as scratches and cracks \citep{Wang2023CarDD}. Empirical studies show that while DCNs improve performance on geometric damage like dents, they still struggle when local cues are insufficient. Similarly, DiffusionDet's iterative refinement is beneficial, but its performance is constrained by its conditioning mechanism, which relies solely on local Region of Interest (RoI) features. This limitation prevents it from effectively leveraging scene-level information—such as global illumination, material properties, and vehicle pose—that is critical for disambiguating subtle damage. A critical and often overlooked aspect of this challenge is domain adaptation. Models trained on curated datasets often fail to generalize to the vast diversity of real-world conditions encountered during deployment, such as changing weather, lighting, camera perspectives, and vehicle models \citep{Tzeng2017DomainAdap,Wilson2022SurveyDomainAdap}. This domain shift exacerbates the difficulty of fine-grained detection, as models overfit to spurious correlations in the training data rather than learning robust, invariant features of the damage itself. Relying purely on local features makes detectors exceptionally vulnerable to these variations, as local appearances can change dramatically across domains while the global semantic context of a "damaged car" remains more stable.

In this work, we argue that the next advance in fine-grained detection requires unifying the iterative refinement of diffusion models with explicit global scene reasoning. To this end, we propose the Context-Aware Diffusion Detector (C-DiffDet+). Our key innovation is the integration of a lightweight Global Context Encoder into the detector's backbone and fusion layers. This module produces a comprehensive scene embedding that conditions the entire denoising diffusion head. By doing so, our model generates more robust proposals and final detections, particularly for categories like scratches and cracks where local evidence is ambiguous. This design enhances the detector's ability to perform scene-level reasoning while fully retaining the powerful generative properties and set prediction capabilities of the original DiffusionDet framework. The subsequent sections will detail the architecture, training protocol, and empirical validation of our approach.

\section{Methodology}
\label{sec:Methodology}

\begin{figure}
    \centering
    \includegraphics[width=0.95\linewidth]{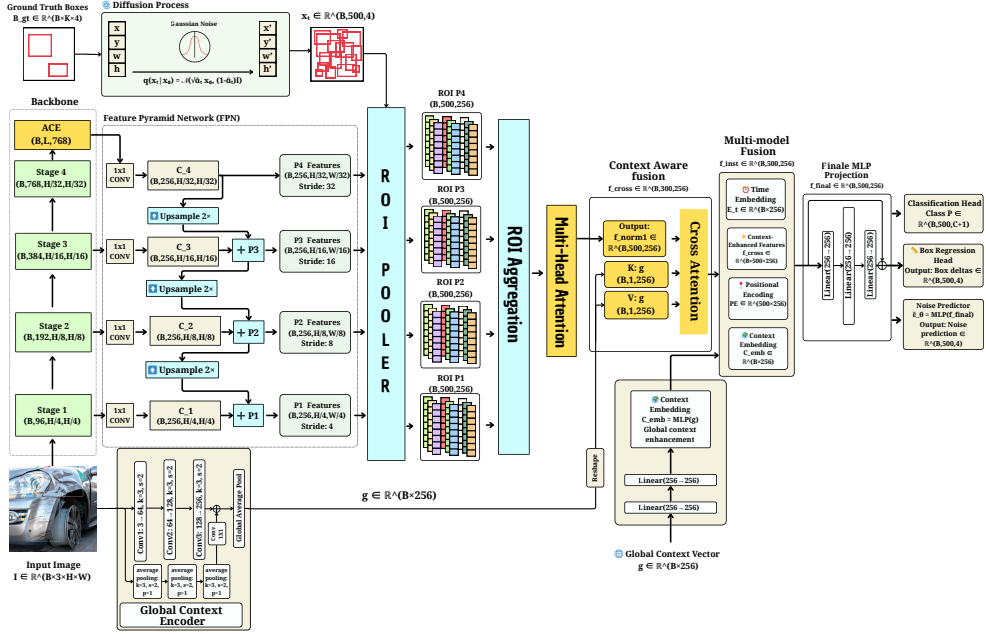}
    \caption{Overview of the proposed Context-Aware DiffusionDet architecture. The framework consists of four key components: (1) Adaptive Channel Enhancement (ACE) blocks that enhance backbone and FPN features, (2) Global Context Encoder (GCE) for comprehensive scene understanding, (3) Context-Aware Fusion (CAF) that integrates global context with local features through cross-attention, and (4) enhanced Multi-Modal Fusion (MMF) with global context embeddings. This architecture addresses the local feature conditioning limitation in existing diffusion-based detectors by enabling the comprehensive integration of environmental context.}
    \label{fig:architecture}
\end{figure}


Our proposed model builds upon the DiffusionDet architecture, extending it with context-aware mechanisms to better capture global scene information for object detection. As illustrated in Figure \ref{fig:architecture}, the framework is composed of four key components: (1) Adaptive Channel Enhancement (ACE) blocks that refine backbone and FPN features, (2) a Global Context Encoder (GCE) that captures holistic scene representations, (3) a Context-Aware Fusion (CAF) module that integrates global context with local proposal features via cross-attention, and (4) an enhanced Multi-Modal Fusion (MMF) module that incorporates global context embeddings.
By modeling object detection as a conditional denoising diffusion process, our framework inherits the generative strengths of DiffusionDet while addressing its key limitation—the reliance on local feature conditioning. The proposed CAF mechanism explicitly fuses global scene context with local features, enabling a more comprehensive understanding of object–environment interactions. In the following sections, we provide the formal mathematical formulation of our model, along with the theoretical motivation behind each component.

\subsection{Denoising Diffusion Framework}

Let $\mathbf{x}_0 \in \mathbb{R}^{N \times 4}$ be a set of $N$ ground-truth boxes. The objective is to learn a model that can reverse a predefined noising process, starting from pure noise and iteratively generating these boxes. This generative formulation avoids the need for hand-designed anchors or proposal generation networks.

\subsubsection{Forward Diffusion Process}

The forward process progressively injects Gaussian noise into the ground-truth boxes over $T$ timesteps. This process defines the training objective for our model. The distribution of noisy boxes $\mathbf{x}_t$ at any timestep $t$ is given by a closed-form solution, enabling efficient training:

\begin{equation}
q(\mathbf{x}_t | \mathbf{x}_0) = \mathcal{N}(\mathbf{x}_t; \sqrt{\bar{\alpha}_t}\mathbf{x}_0, (1 - \bar{\alpha}_t)\mathbf{I})
\label{eq:forward}
\end{equation}

where $\bar{\alpha}_t = \prod_{s=1}^{t}(1-\beta_s)$ is derived from a cosine variance schedule. This schedule is chosen for its smooth and gradual addition of noise, which has been shown to improve training stability and the quality of the learned reverse process \citep{nichol2021improved}.

\subsubsection{Reverse Denoising Process}

The reverse process is learned by a network $f_\theta$ that predicts the original clean boxes $\mathbf{\hat{x}}_0$ from a noisy state $\mathbf{x}_t$, conditioned on the image $\mathbf{I}$ and timestep $t$:

\begin{equation}
\mathbf{\hat{x}}_0 = f_\theta(\mathbf{x}_t, t, \mathbf{I})
\label{eq:reverse_goal}
\end{equation}

The design of this function $f_\theta$ is the primary focus of our methodology, as its capacity to leverage contextual information through our proposed Context-Aware Fusion mechanism dictates the model's performance.

\subsection{Context-Aware Feature Extraction}

The efficacy of the denoising function hinges on its ability to extract and utilize both local and global visual information. As depicted in Figure \ref{fig:architecture}, this is achieved through a dual-stream backbone that processes the image in parallel to capture different levels of context.

\subsubsection{Feature Generation}

From an input image $\mathbf{I}$, a Swin Transformer backbone generates a hierarchy of feature maps at four stages, $\{\mathbf{C}_2, \mathbf{C}_3, \mathbf{C}_4, \mathbf{C}_5\}$, with strides of $\{4, 8, 16, 32\}$. This provides a powerful, multi-scale representation of the image.

\subsubsection{Adaptive Channel Enhancement (ACE)}

To improve feature quality, we apply Adaptive Channel Enhancement (ACE) blocks to the backbone features, particularly at Stage 4. The ACE mechanism implements channel-wise attention following a squeeze-excitation approach:

\begin{align}
\mu_c &= \frac{1}{H_{32} \times W_{32}} \sum_{h=1}^{H_{32}} \sum_{w=1}^{W_{32}} \mathbf{C}_5^{(c)}[h, w] \in \mathbb{R}^{B \times C} \\
\mathbf{a}_c &= \sigma(\mathbf{W}_2 \cdot \text{ReLU}(\mathbf{W}_1 \cdot \mu_c + \mathbf{b}_1) + \mathbf{b}_2) \in \mathbb{R}^{B \times C} \\
\mathbf{C}_5^{\text{enhanced}} &= \mathbf{C}_5 \odot \mathbf{a}_c
\label{eq:ace}
\end{align}

where $\odot$ denotes element-wise multiplication, $\mathbf{a}_c$ represents the learned channel attention weights, $H_{32} = H/32$ and $W_{32} = W/32$ are the spatial dimensions at stride 32 (Stage 4), $\mathbf{W}_1 \in \mathbb{R}^{C \times C/16}$, $\mathbf{W}_2 \in \mathbb{R}^{C/16 \times C}$ are learnable weight matrices, $\mathbf{b}_1, \mathbf{b}_2$ are bias terms, and $\sigma$ represents the sigmoid activation function. The reduction ratio is set to 16 for computational efficiency.

The channel-wise attention mechanism begins by compressing spatial information using global average pooling to obtain channel-wise statistics. It then learns adaptive channel weights through a two-layer MLP with ReLU activation and re-weights the original features accordingly. This approach allows the model to focus on the most informative channels for object detection, improving feature discriminability and performance. The ACE block is applied only to the final stage (Stage 4) of the Swin Transformer backbone to enhance the C5 features before they are sent to the FPN.

\subsubsection{Feature Pyramid Network (FPN)}

A Feature Pyramid Network (FPN)\citep{lin2017feature} constructs a rich multi-scale feature representation from the enhanced backbone features. The FPN follows a top-down pathway with lateral connections, where higher-level, semantically stronger features are upsampled and combined with lower-level, spatially finer features:

\begin{align}
\mathbf{P}_5 &= \text{Conv}_{3\times3}(\mathbf{C}_5^{\text{enhanced}}) \\
\mathbf{P}_4 &= \text{Conv}_{3\times3}(\mathbf{C}_4 + \text{Upsample}(\mathbf{P}_5)) \\
\mathbf{P}_3 &= \text{Conv}_{3\times3}(\mathbf{C}_3 + \text{Upsample}(\mathbf{P}_4)) \\
\mathbf{P}_2 &= \text{Conv}_{3\times3}(\mathbf{C}_2 + \text{Upsample}(\mathbf{P}_3))
\label{eq:fpn}
\end{align}

Each FPN level $\mathbf{P}_i$ has a consistent channel dimension of 256, providing rich features for objects at various scales. The FPN enables the model to handle objects of different sizes effectively by providing appropriate feature resolution for each scale.

\subsection{Multi-Modal Feature Integration}

The detection head, shown centrally in Figure \ref{fig:architecture}, is designed to fuse the visual features with the current state of the diffusion proposals and other conditioning signals.

\subsubsection{RoI Feature Extraction from Diffusion Proposals}

The noisy proposals $\mathbf{x}_t$ are used to spatially sample the image features via RoIAlign. This is a critical step where the abstract box coordinates from the diffusion process are connected to concrete visual evidence from the FPN feature maps $\{\mathbf{P}_i\}$. 
For each proposal box, we first determine the appropriate FPN level based on the box size using a logarithmic assignment strategy:
\begin{equation}
    \text{level} = \lfloor \log_2(\sqrt{\text{area}} / s_{\text{base}}) \rfloor + 4
    \label{eq:fpn_level}
    \end{equation}
    where $s_{\text{base}} = 224$ is the base size and the level is clamped to the range [1, 5].
    
    The RoIAlign operation then extracts fixed-size features from the assigned FPN level:
    \begin{equation}
    \mathbf{f}_{\text{roi}} = \Phi_{\text{RoIAlign}}(\{\mathbf{P}_i\}, \mathbf{x}_t, \text{levels}) \in \mathbb{R}^{B \times N \times 7 \times 7 \times 256}
    \label{eq:roi_features_raw}
    \end{equation}
    
    These features are then aggregated using global average pooling to produce a compact representation:
    \begin{equation}
    \mathbf{f}_{\text{roi}} = \text{GlobalAvgPool}(\mathbf{f}_{\text{roi}}) \in \mathbb{R}^{B \times N \times 256}
    \label{eq:roi_features}
    \end{equation}

\subsubsection{Context-Aware Fusion (CAF)}

The Context-Aware Fusion mechanism integrates global scene context with local proposal features through cross-attention. This mechanism requires a Global Context Encoder (GCE) to generate comprehensive scene-level understanding.
The GCE The global context extractor processes the entire image to capture holistic scene information, as described in Algorithm \ref{alg:gce}. It is composed of three convolutional layers with residual connections. To preserve spatial information, a residual path downsamples the input image and projects it to the same feature dimension, ensuring compatibility with the main branch. The final global context vector is obtained by fusing the output features with the residual path and applying Global Average Pooling (GAP):

\begin{equation}
\mathbf{g} = \text{GAP}(\mathbf{F}3 + \mathbf{I}{\text{res}}) \in \mathbb{R}^{B \times D_f},
\label{eq:global_context}
\end{equation}

where GAP denotes Global Average Pooling and $D_f = 256$ is the feature dimension.

The CAF mechanism operates in two steps:

\begin{enumerate}
    \item \textbf{Inter-Proposal Self-Attention:} This initial module allows the set of $N$ RoI features to communicate with each other. Each proposal feature in $\mathbf{f}_{\text{roi}}$ is linearly projected to a Query, Key, and Value:
    \begin{align}
        \mathbf{Q}_s &= \mathbf{f}_{\text{roi}}\mathbf{W}_{Qs} \\
        \mathbf{K}_s &= \mathbf{f}_{\text{roi}}\mathbf{W}_{Ks} \\
        \mathbf{V}_s &= \mathbf{f}_{\text{roi}}\mathbf{W}_{Vs}
    \end{align}
    The output is a refined set of features where each proposal's representation has been updated based on its interaction with all other proposals:
    \begin{equation}
    \mathbf{f}_{\text{self}} = \text{LayerNorm}(\mathbf{f}_{\text{roi}} + \text{MultiHeadAttn}(\mathbf{Q}_s, \mathbf{K}_s, \mathbf{V}_s))
    \label{eq:self_attention}
    \end{equation}
    
    \item \textbf{Global-Local Cross-Attention:} In the second step, each self-attended proposal feature $\mathbf{f}_{\text{self}}$ queries the global context vector $\mathbf{g}$. This mechanism infuses top-down, scene-level information into each individual object hypothesis, allowing the model to disambiguate challenging cases:
    \begin{align}
        \mathbf{Q}_c &= \mathbf{f}_{\text{self}}\mathbf{W}_{Qc} \\
        \mathbf{K}_c, \mathbf{V}_c &= \mathbf{g}\mathbf{W}_{Kc}, \mathbf{g}\mathbf{W}_{Vc}
    \end{align}
    The resulting context-aware features are then computed:
    \begin{equation}
    \mathbf{f}_{\text{cross}} = \text{LayerNorm}(\mathbf{f}_{\text{self}} + \text{MultiHeadAttn}(\mathbf{Q}_c, \mathbf{K}_c, \mathbf{V}_c))
    \label{eq:cross_attn}
    \end{equation}
\end{enumerate}

\subsubsection{Enhanced Multi-Modal Fusion}
The existing Multi-Modal Fusion framework is enhanced by integrating global context embeddings, creating a unified representation that combines temporal, positional, and contextual information. This enhancement addresses the fundamental limitation of local feature conditioning by providing each proposal with access to comprehensive scene-level understanding.
Three types of conditional embeddings are generated to provide non-visual context:
\begin{enumerate}
    \item \textbf{Time Embedding ($\mathbf{E}_t$):} The time embedding captures the current diffusion timestep $t$ through sinusoidal encoding, enabling the model to adapt its predictions based on the denoising iteration. This embedding is generated as:
    \begin{equation}
    \mathbf{E}_t = \text{MLP}(\text{TimeEmbedding}(t)) \in \mathbb{R}^{B \times 256}
    \end{equation}
    where the time embedding uses the standard sinusoidal positional encoding scaled to the diffusion timestep range.
    
    \item \textbf{Positional Embedding ($\mathbf{PE}$):} Positional embeddings preserve the ordering and identity of proposals, generated from sequential indices $[0, 1, 2, \ldots, N-1]$:
    \begin{equation}
    \mathbf{PE} = \text{MLP}(\text{PositionalEncoding}([0, 1, \ldots, N-1])) \in \mathbb{R}^{N \times 256}
    \end{equation}
    These embeddings are trainable and learn to encode proposal-specific positional information.
    
    \item \textbf{Context Embedding ($\mathbf{C}_{\text{emb}}$):} Context embeddings integrate global scene understanding by processing the global context vector through an MLP:
    \begin{equation}
    \mathbf{C}_{\text{emb}} = \text{MLP}(\mathbf{g}) \in \mathbb{R}^{B \times 256}
    \end{equation}
    where $\mathbf{g}$ is the global context from the Global Context Encoder.
\end{enumerate}
The enhanced Multi-Modal Fusion combines context-aware features with conditional embeddings through a cross-attention mechanism:
First, the conditional embeddings are prepared and broadcast to match the proposal dimensions:

\begin{align}
\mathbf{E}_t^{\text{broadcast}} &= \mathbf{E}_t \cdot \mathbf{1}_{N}^T \in \mathbb{R}^{B \times N \times 256} \\
\mathbf{C}_{\text{emb}}^{\text{broadcast}} &= \mathbf{C}_{\text{emb}} \cdot \mathbf{1}_{N}^T \in \mathbb{R}^{B \times N \times 256}
\end{align}
where $\mathbf{1}_{N} \in \mathbb{R}^{N}$ is a vector of ones.
The embeddings are concatenated along the feature dimension to form a unified latent representation:

\begin{equation}
\mathbf{L}_{\text{latent}} = \text{concat}(\mathbf{E}_t^{\text{broadcast}}, \mathbf{PE}, \mathbf{C}_{\text{emb}}^{\text{broadcast}}) \in \mathbb{R}^{B \times N \times 768}
\label{eq:latent_concat}
\end{equation}
The final fused representation is computed through cross-attention between context-aware features and latent embeddings:

\begin{align}
\mathbf{Q}_{\text{mmf}} &= \mathbf{f}_{\text{cross}} \mathbf{W}_{Q,\text{mmf}} \in \mathbb{R}^{B \times N \times 256} \\
\mathbf{K}_{\text{mmf}} &= \mathbf{L}_{\text{latent}} \mathbf{W}_{K,\text{mmf}} \in \mathbb{R}^{B \times N \times 256} \\
\mathbf{V}_{\text{mmf}} &= \mathbf{L}_{\text{latent}} \mathbf{W}_{V,\text{mmf}} \in \mathbb{R}^{B \times N \times 256}
\end{align}

The attention weights are computed as:
\begin{equation}
\mathbf{A}_{\text{mmf}} = \text{softmax}\left(\frac{\mathbf{Q}_{\text{mmf}} \mathbf{K}_{\text{mmf}}^T}{\sqrt{256}}\right) \in \mathbb{R}^{B \times N \times N}
\label{eq:attention_weights}
\end{equation}

The attended output is:
\begin{equation}
\mathbf{f}_{\text{attended}} = \mathbf{A}_{\text{mmf}} \mathbf{V}_{\text{mmf}} \in \mathbb{R}^{B \times N \times 256}
\label{eq:attended_output}
\end{equation}
 The final fused representation combines the context-aware features with the attended multi-modal information:

\begin{equation}
\mathbf{f}_{\text{fused}} = \text{LayerNorm}(\mathbf{f}_{\text{cross}} + \mathbf{f}_{\text{attended}})
\label{eq:mmf_final}
\end{equation}

The enhanced Multi-Modal Fusion module is implemented as a trainable component that learns optimal projection matrices $\mathbf{W}_{Q,\text{mmf}}$, $\mathbf{W}_{K,\text{mmf}}$, and $\mathbf{W}_{V,\text{mmf}}$ to maximize the effectiveness of the cross-modal information integration. This design ensures that the model can adaptively learn which aspects of the multi-modal information are most relevant for each specific detection scenario.

After the multi-modal fusion, a final MLP projection is applied to refine the fused features and prepare them for the detection head:

\begin{equation}
\mathbf{f}_{\text{final}} = \text{MLP}_{\text{final}}(\mathbf{f}_{\text{fused}}) \in \mathbb{R}^{B \times N \times 256}
\label{eq:final_mlp}
\end{equation}

where $\text{MLP}_{\text{final}}$ consists of two linear layers with ReLU activation and dropout:
\begin{align}
\mathbf{h}_1 &= \text{ReLU}(\mathbf{f}_{\text{fused}} \mathbf{W}_1 + \mathbf{b}_1) \in \mathbb{R}^{B \times N \times 512} \\
\mathbf{f}_{\text{final}} &= \text{Dropout}(\mathbf{h}_1 \mathbf{W}_2 + \mathbf{b}_2) \in \mathbb{R}^{B \times N \times 256}
\end{align}

where $\mathbf{W}_1 \in \mathbb{R}^{256 \times 512}$, $\mathbf{W}_2 \in \mathbb{R}^{512 \times 256}$, $\mathbf{b}_1 \in \mathbb{R}^{512}$, and $\mathbf{b}_2 \in \mathbb{R}^{256}$ are learnable parameters.

This final projection learns a refined feature representation that optimally combines multimodal information through a bottleneck architecture (256→512→256), while ensuring the output features have the proper dimensionality and statistical properties for the detection head. The bottleneck design captures the most essential multimodal information in a compressed form, maintaining both expressiveness and computational efficiency.

\subsection{Training and Inference}

The final, refined representation $\mathbf{f}_{\text{final}}$ serves as the input to the prediction heads, which translate these rich features into concrete object detections.

As shown on the right of Figure \ref{fig:architecture}, the final representation is passed to three independent heads, each implemented as a small MLP:
    \begin{enumerate}
        \item \textbf{Classification Head:} Predicts the class probabilities for each of the $N$ proposals, including a "no object" class.
        \item \textbf{Box Regression Head:} Predicts the refined coordinates of the clean box $\mathbf{\hat{x}}_0$ for each proposal.
        \item \textbf{Noise Predictor:} Predicts the noise vector $\hat{\boldsymbol{\epsilon}}_\theta$ that was added to the original clean box to produce $\mathbf{x}_t$.
    \end{enumerate}
    
We employ a set-based loss, which requires a bipartite matching between the set of $N$ predictions and the set of $M$ ground-truth objects. The optimal matching $\hat{\sigma}$ is found using the Hungarian algorithm to minimize a pairwise matching cost:
    \begin{equation}
    \mathcal{C}_{\text{match}} = \lambda_{\text{cls}}\mathcal{L}_{\text{focal}}(p_i, c_{\sigma(i)}) + \lambda_{\text{L1}}\mathcal{L}_{\text{L1}}(b_i, b_{\sigma(i)}) + \lambda_{\text{giou}}\mathcal{L}_{\text{giou}}(b_i, b_{\sigma(i)})
    \label{eq:matching_cost}
    \end{equation}
    where $\lambda_{\text{cls}}$, $\lambda_{\text{L1}}$, and $\lambda_{\text{giou}}$ are loss weighting coefficients, $p_i$ represents the predicted class probabilities, $c_{\sigma(i)}$ denotes the ground-truth class, and $b_i$ and $b_{\sigma(i)}$ represent the predicted and ground-truth bounding boxes, respectively.
    
    Once the matching is established, the total loss is computed as a weighted sum of the same losses over the matched pairs:
    \begin{equation}
    \mathcal{L}_{\text{total}} = \sum_{i=1}^{N} \mathcal{L}_{\text{match}}(y_i, \hat{y}_{\hat{\sigma}(i)})
    \label{eq:total_loss}
    \end{equation}
    where $\mathcal{L}_{\text{match}}$ combines the classification, regression, and noise prediction losses \citep{Carion2020EndToEndOD,Sun_2021_CVPR,zhu2021DeformableDETR} for each matched proposal-ground-truth pair.
    
The inference process is a generative loop that starts from a set of $N$ boxes sampled from pure Gaussian noise, $\mathbf{x}_T \sim \mathcal{N}(0, \mathbf{I})$. The model then iteratively applies the reverse denoising step for a small number of timesteps. We use the Denoising Diffusion Implicit Models (DDIM) formulation for efficient sampling:
    \begin{equation}
    \mathbf{x}_{t-1} = \sqrt{\bar{\alpha}_{t-1}}\mathbf{\hat{x}}_0^{(t)} + \sqrt{1 - \bar{\alpha}_{t-1}} \cdot \hat{\boldsymbol{\epsilon}}_\theta(\mathbf{x}_t, t, \mathbf{I})
    \label{eq:ddim_update}
    \end{equation}
    where $\mathbf{\hat{x}}_0^{(t)} = f_\theta(\mathbf{x}_t, t, \mathbf{I})$ is the predicted clean box at step $t$, and the noise predictor $\hat{\boldsymbol{\epsilon}}_\theta$ is conditioned on the current noisy state $\mathbf{x}_t$, timestep $t$, and input image $\mathbf{I}$.
    
    The final predictions are obtained after the last denoising step, where the model outputs the refined bounding boxes $\mathbf{\hat{x}}_0$ along with their corresponding class probabilities and confidence scores. Non-maximum suppression (NMS) is then applied to remove duplicate detections and produce the final set of object predictions.

\section{Experiments and Results}
\label{sec:Experiments}
\subsection{Experimental Setup}
\subsubsection{Dataset and Preprocessing}
For the training and evaluation of our model, we use the public CarDD (Car Damage Detection) dataset \cite{CarDD2023Wang}. This dataset consists of 4,000 high-resolution images containing over 9,000 annotated damage instances. The annotations, provided as bounding boxes in COCO format, are categorized into six classes: dent, scratch, crack, glass shatter, tire flat, and lamp broken. The standard data split is 70.4\% for training, 20.19\% for validation, and 9.35\% for testing.

\begin{table*}[h!]
\centering
\caption{Distribution of instances per category across Train, Validation, and Test datasets.}
\label{tab:combined_dataset_distribution}
\resizebox{\textwidth}{!}{%
\begin{tabular}{|l|r|r|r|r|}
\hline
\textbf{Category} & \textbf{\# Train Instances} & \textbf{\# Val Instances} & \textbf{\# Test Instances} & \textbf{Total Instances} \\
\hline
dent & 1806 & 501 & 236 & 2543 \\
scratch & 2560 & 728 & 307 & 3595 \\
crack & 651 & 177 & 70 & 898 \\
glass shatter & 475 & 135 & 71 & 681 \\
lamp broken & 494 & 141 & 69 & 704 \\
tire flat & 225 & 62 & 32 & 319 \\
\hline
\textbf{Total} & \textbf{6211} & \textbf{1744} & \textbf{785} & \textbf{8740} \\
\hline

\end{tabular}
}
\end{table*}

During training, we employ standard data augmentation techniques, including random horizontal flipping and multi-scale resizing with cropping, to enhance the model's generalization capabilities \citep{Fang_2024DataAug,Benaissa2025Gated}.

The CarDD \cite{CarDD2023Wang} dataset presents significant challenges for object detection models. A primary difficulty is the fine-grained nature of the damage categories, which results in low inter-class variance; for instance, distinguishing between a 'scratch' and a 'crack' is non-trivial. The dataset also features a wide diversity in the scale and shape of damages, from large-area deformations to small, localized defects. This is compounded by the high prevalence of small object instances, which requires a model to effectively detect features at various resolutions \citep{zidi2025lola}. These characteristics make CarDD a demanding benchmark for assessing model robustness in realistic damage detection scenarios.

\subsubsection{Implementation Details and Hyperparameters}
\label{sec:imp_details}

Our model is implemented based on the \texttt{Detectron2} framework \citep{wu2019detectron2}, with a Swin backbone pretrained on ImageNet-1K and ImageNet-21K,
respectively \citep{Deng2009ImageNet, Chen2022DiffusionDet}. For optimization, we utilize the AdamW optimizer \citep{loshchilov2019AdamW}, with a base learning rate of $2.5 \times 10^{-5}$ and a weight decay of $0.0001$. The model is trained for a total of 20,000 iterations using a batch size of 2. The learning rate schedule includes a 1000-iteration linear warm-up, followed by learning rate decay steps at the 15,000 and 18,000 iteration marks. To ensure training stability, we apply gradient clipping with a maximum norm of 1.0. The DiffusionDet head is configured to use 500 proposals during the iterative denoising process.

\subsubsection{Evaluation Metrics}
\label{sec:metrics}
To ensure a fair comparison, we strictly follow the evaluation protocol of the CarDD benchmark \cite{CarDD2023Wang}, which builds on the COCO evaluation suite and provides a comprehensive framework for measuring model performance. Our evaluation focuses on three aspects: detection accuracy, scale-dependent behavior, and computational efficiency.

\paragraph{Detection Accuracy}

Following CarDD, we report mean Average Precision (mAP) as the primary metric \citep{Beitzel2009mAP}. The challenge metric, AP, corresponds to COCO dataset mAP@[.5:.95], averaged over IoU thresholds from 0.5 to 0.95. We also include AP\textsubscript{50} and AP\textsubscript{75}, which evaluate detection performance at fixed thresholds, reflecting coarse- and fine-grained localization accuracy see the detailed eqautions for the metrics in appendx \ref{eq:attended_output}.

\paragraph{Scale-Specific Performance}
To analyze robustness across object sizes, we adopt the COCO-defined metrics AP\textsubscript{S}, AP\textsubscript{M}, and AP\textsubscript{L}, which measure accuracy on small (area < $32^2$ px), medium ($32^2$–$96^2$ px), and large (area > $96^2$ px) instances. These results highlight the operational strengths and limitations of our method.

\paragraph{Computational Efficiency}
Since our \textbf{C-DiffDet+} introduces a novel architecture, we extend the CarDD protocol with a computational efficiency analysis. We report latency (ms), throughput (FPS), GFLOPs, and parameter count in the appendix \ref{tab:model_complexity} , all benchmarked on dual NVIDIA RTX 4500 Ada GPUs (24GB VRAM each). This provides a balanced view of both accuracy and practical deployability.

\subsection{Results}
\subsubsection{Quantitative Comparison with State-of-the-Art Methods}

Table \ref{tab:bbox_ap_cardd_v2} presents a comprehensive comparison of bounding box detection performance across different evaluation metrics on the CarDD dataset. Our C-DiffDet+ model demonstrates superior performance across most metrics, establishing new state-of-the-art results in the majority of evaluation criteria and showcasing the effectiveness of our context-aware diffusion framework.
In terms of overall performance, C-DiffDet+ achieves a remarkable AP$^{bb}$ of 64.8\%, representing a significant improvement of 1.4\% over the previous best method, DiffDet. This substantial gain in mean Average Precision demonstrates that our context-aware approach effectively addresses the fundamental limitations of local feature conditioning in existing detection frameworks. The integration of global scene context through cross-attention mechanisms enables more accurate localization and classification of automotive damage across diverse scenarios.
At higher IoU thresholds, our model shows particularly strong performance. C-DiffDet+ achieves an AP$_{75}^{bb}$ of 67.9\%, outperforming DiffDet by 1.7\%. This improvement at the stricter IoU threshold indicates that our model not only detects damage but also provides more precise bounding box localization. The enhanced precision can be attributed to the iterative refinement capabilities of our diffusion-based approach, where the denoising process progressively improves localization accuracy through multiple refinement steps.
The most dramatic improvement is observed in small object detection (AP$_S^{bb}$), where C-DiffDet+ achieves 45.5\%, representing a remarkable 6.8\% increase over DiffDet. This exceptional performance in small damage detection validates our context-aware fusion mechanism, as small damage types like fine scratches or hairline cracks often require global scene understanding to distinguish from visual artifacts or natural surface variations. The Global Context Encoder provides comprehensive scene-level information that helps disambiguate these challenging cases.
For medium-sized objects (AP$_M^{bb}$), our model achieves 39.2\%, which is 8.8\% lower than DCN+. This performance gap suggests that while our context-aware approach excels at fine-grained and small damage detection, medium-sized damage patterns may benefit from different architectural considerations. Medium-sized damage often involves more complex geometric patterns that might require enhanced local feature processing or different attention mechanisms.
In large object detection (AP$_L^{bb}$), C-DiffDet+ achieves 66.0\%, matching the performance of DCN+ and outperforming DiffDet by 1.8\%. This parity in large object detection indicates that our model maintains competitive performance for substantial damage types while significantly improving detection of smaller, more challenging cases. Large damage patterns are typically more visible and may not require the same level of contextual reasoning that benefits smaller damage detection.
The superior performance of C-DiffDet+ across most metrics can be attributed to several key architectural advantages. The Context-Aware Fusion (CAF) mechanism, which integrates global scene context with local proposal features, proves particularly effective for damage types that require comprehensive environmental understanding. The Adaptive Channel Enhancement (ACE) blocks in the backbone enable dynamic feature re-weighting based on learned importance, improving the model's ability to focus on the most informative visual cues.
Furthermore, the diffusion-based iterative refinement process provides progressive improvement in bounding box localization, as evidenced by the strong AP$_{75}^{bb}$ performance. The multi-modal fusion with global context embeddings ensures that each proposal benefits from comprehensive scene understanding, leading to more accurate damage classification and localization.
These results validate our core architectural contributions and demonstrate that integrating global contextual information with diffusion-based detection significantly advances the state-of-the-art in automotive damage assessment, particularly for challenging fine-grained and small-scale damage types.


\begin{table}[h]
\centering
\caption{Comparison of Bounding Box (\%) on the CarDD dataset.}
\label{tab:bbox_ap_cardd_v2}
\begin{tabular}{l|cccccc}
\hline
\textbf{Method} & \textbf{AP$^{bb}$} & \textbf{AP$_{50}^{bb}$} & \textbf{AP$_{75}^{bb}$} & \textbf{AP$_S^{bb}$} & \textbf{AP$_M^{bb}$} & \textbf{AP$_L^{bb}$} \\
\hline
Mask R-CNN \cite{MaskRCNN2017He} & 51.1 & 67.7 & 54.2 & 19.0 & 39.1 & 61.1 \\
CM-R-CNN \cite{CascadeRCNN2021} & 52.0 & 65.4 & 54.8 & 16.7 & 37.1 & 61.6 \\
GCNet \cite{GCNET2019} & 52.6 & 69.3 & 56.6 & 21.2 & \underline{45.8} & 58.4 \\
HTC \cite{HTC2019} & 53.4 & 68.4 & 54.6 & 20.32 & 39.6 & 62.9 \\
DCN \cite{DCN2017} & 54.3 & 69.8 & 58.3 & 22.7 & 42.6 & 63.6 \\
DCN+ \cite{CarDD2023Wang} & 60.6 & 78.8 & 64.8 & 37.1 & \textbf{48.0} & \textbf{66.0} \\
DiffDet \cite{Chen2022DiffusionDet} & \underline{63.4} & \underline{81.7} & \underline{66.2} & \underline{38.7} & 36.1 & \underline{64.2} \\
C-DiffDet+ (ours) & \textbf{64.8} & \textbf{83.6} & \textbf{67.9} & \textbf{45.5} & 39.2 & \textbf{66.0} \\
\hline
\end{tabular}
\end{table}

Table \ref{tab:class-wise-res-cardd} presents a per-category average precision (AP) comparison on the CarDD dataset, evaluating our C-DiffDet+ model against prior state-of-the-art methods. C-DiffDet+ attains new state-of-the-art performance in three of the six damage categories, with the most pronounced improvements observed on fine-grained and geometrically complex damage types.

Specifically, C-DiffDet+ achieves a leading AP of 42.2\% on the crack category, a substantial absolute improvement of 7.1\% over the previous best result from DiffDet (35.1\%). For lamp broken, our model also sets a new record with 80.2\% AP, surpassing DiffDet by 3.4\%. The highest performance is seen in the glass shatter category, where C-DiffDet+ reaches 94.2\% AP, exceeding the prior state-of-the-art.

While not the top-performing model in all categories, C-DiffDet+ remains highly competitive. In the scratch category, it achieves a strong 42.6\% AP. For tire flat, it scores 92.0\% AP, closely trailing the best result. The model's performance on the dent category is 37.0\%.

The empirical improvements can be attributed to the Context-Aware Fusion (CAF) module, which conditions local proposal features on a global scene representation through cross-attention. This approach enhances the ability to differentiate between subtle damage patterns and background textures, which is particularly effective for detecting low-contrast, fine-grained defects such as cracks. Additionally, the Adaptive Channel Enhancement (ACE) blocks in the backbone dynamically adjust channel responses, thereby enhancing the representation of structural cues that are crucial for identifying damage, such as broken lamps.

In conclusion, The per-category analysis demonstrates that C-DiffDet+ advances the state of the art in three out of six damage types on CarDD, with significant improvements on the most challenging fine-grained categories. These results substantiate the effectiveness of integrating global contextual reasoning into a diffusion-based detection framework for automotive damage assessment.

\begin{table}[h]
\centering
\caption{Comparison between our C-DiffDet+ and the state-of-the-art on each category in CarDD in terms of mask and box AP. Bold text indicates the best result and underlined text indicates the second best.}
\label{tab:class-wise-res-cardd}
\resizebox{\textwidth}{!}{%
\begin{tabular}{l|cccccc}
\hline
\textbf{Method} & \textbf{dent} & \textbf{scratch} & \textbf{crack} & \textbf{glass shatter} & \textbf{lamp broken} & \textbf{tire flat} \\ \hline
Mask R-CNN \cite{MaskRCNN2017He} & 29.4 & 25.6 & 20.2 & 88.5 & 62.0 & 80.8 \\
CM-R-CNN \cite{CascadeRCNN2021} & 28.5 & 24.5 & 21.3 & 90.6 & 62.0 & 85.1 \\
GCNet \cite{GCNET2019} & 29.6 & 26.5 & 21.6 & 89.3 & 66.9 & 85.0 \\
HTC \cite{HTC2019} & 30.4 & 26.8 & 18.6 & \underline{92.7} & 64.3 & 87.5 \\
DCN \cite{DCN2017} & 33.0 & 29.7 & 17.5 & \underline{92.7} & 66.7 & 86.3 \\
DCN+ \cite{CarDD2023Wang} & \textbf{42.2} & 42.3 & 29.6 & 90.1 & 69.5 & 90.2 \\
DiffDet \cite{Chen2022DiffusionDet} & \underline{38.9} & \textbf{44.8} & \underline{35.1} & 92.6 & \underline{76.8} & \textbf{92.5} \\
C-DiffDet+ (ours) & 37.0 & \underline{42.6} & \textbf{42.2} & \textbf{94.2} & \textbf{80.2} & \underline{92.0} \\
\hline
\end{tabular}%
}
\end{table}

\subsubsection{Visual Comparison}
\begin{figure*}[h!]
    \centering

    \begin{tabular}{ >{\centering\arraybackslash}m{0.0\textwidth} m{1\textwidth} }
        \rotatebox{90}{\small GT} &
        \begin{subfigure}[b]{0.2\linewidth}\includegraphics[width=\textwidth]{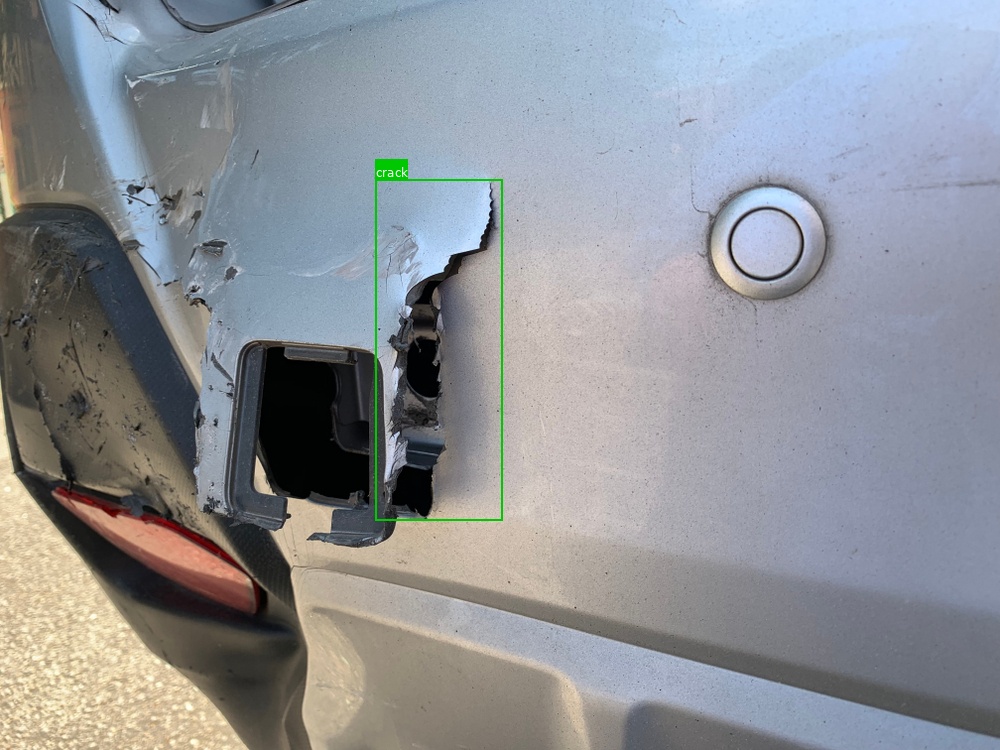}\end{subfigure}%
        \begin{subfigure}[b]{0.2\linewidth}\includegraphics[width=\textwidth]{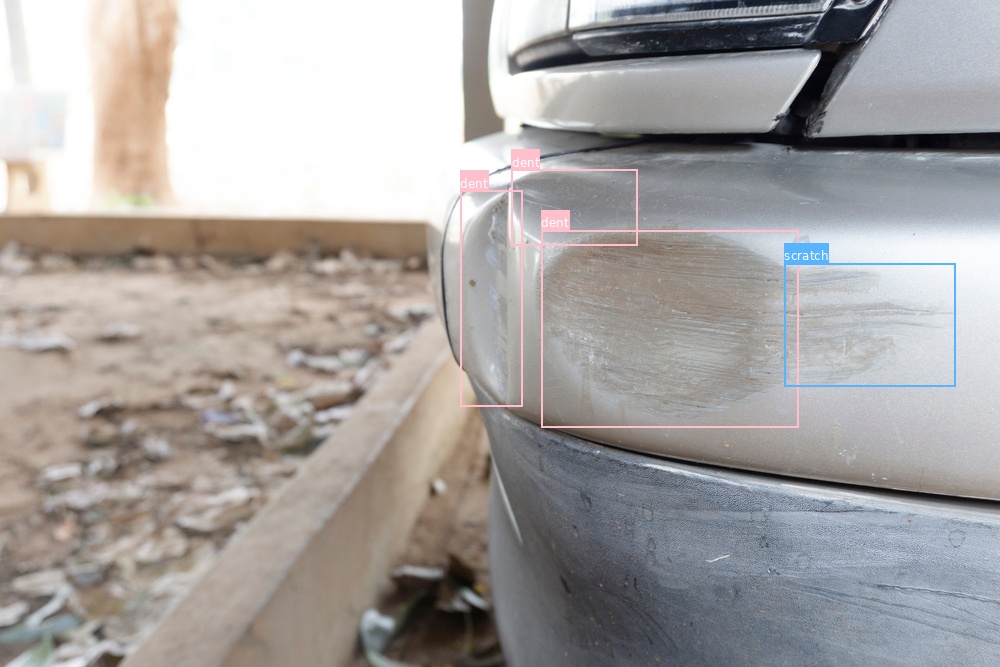}\end{subfigure}%
        \begin{subfigure}[b]{0.2\linewidth}\includegraphics[width=\textwidth]{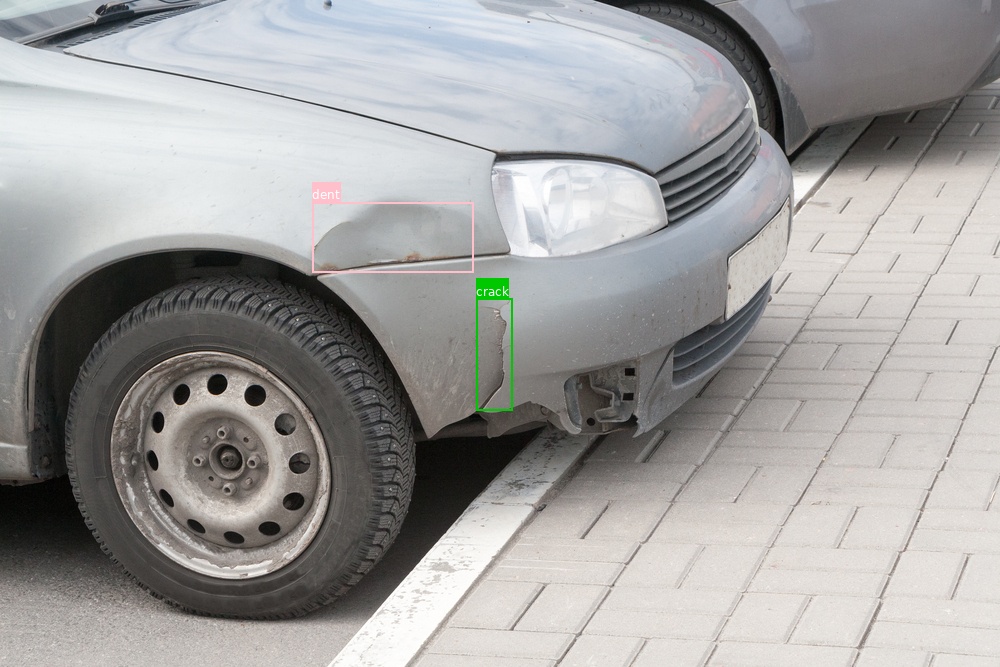}\end{subfigure}%
        \begin{subfigure}[b]{0.2\linewidth}\includegraphics[width=\textwidth]{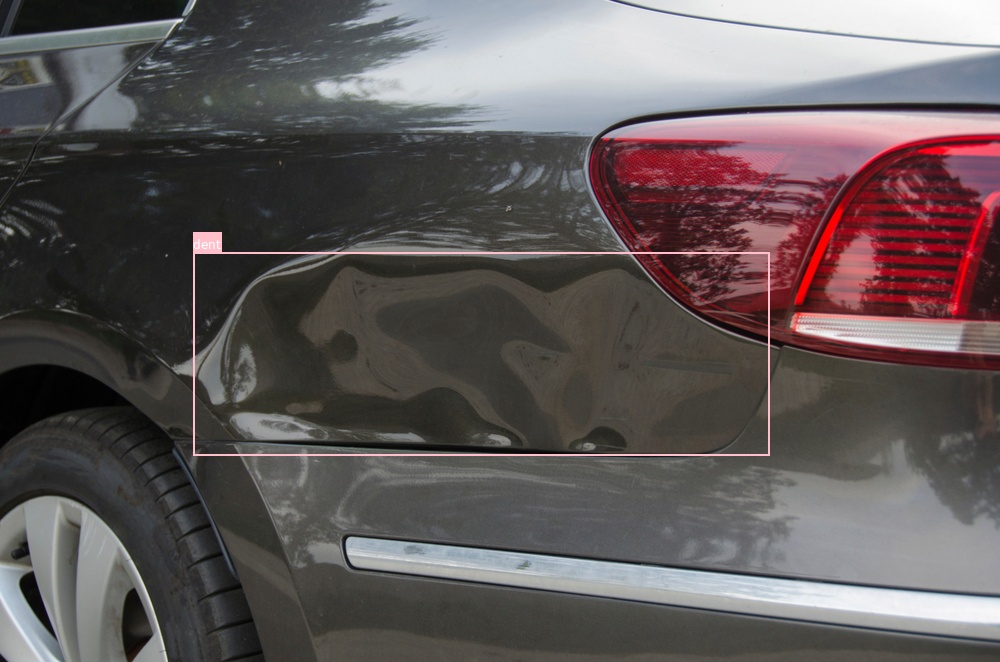}\end{subfigure}%
        \begin{subfigure}[b]{0.2\linewidth}\includegraphics[width=\textwidth]{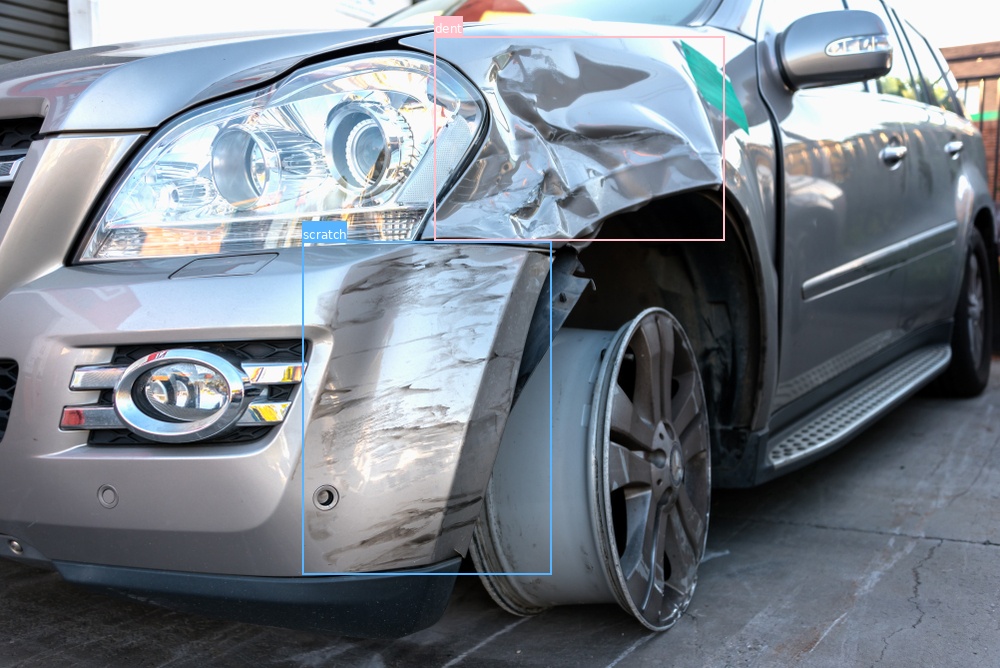}\end{subfigure}
    \end{tabular}

    \begin{tabular}{ >{\centering\arraybackslash}m{0.0\textwidth} m{1\textwidth} }
        \rotatebox{90}{\small DCN+} &
        \begin{subfigure}[b]{0.2\linewidth}\includegraphics[width=\textwidth]{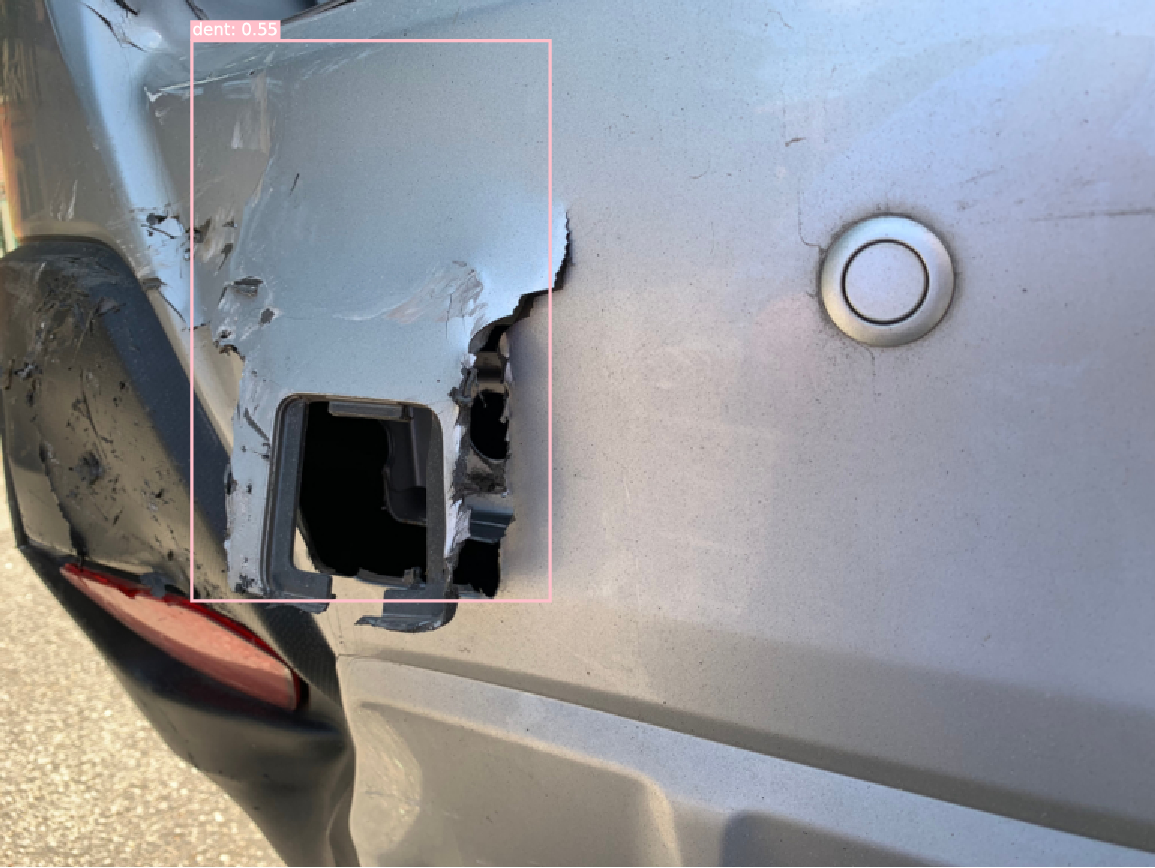}\end{subfigure}%
        \begin{subfigure}[b]{0.2\linewidth}\includegraphics[width=\textwidth]{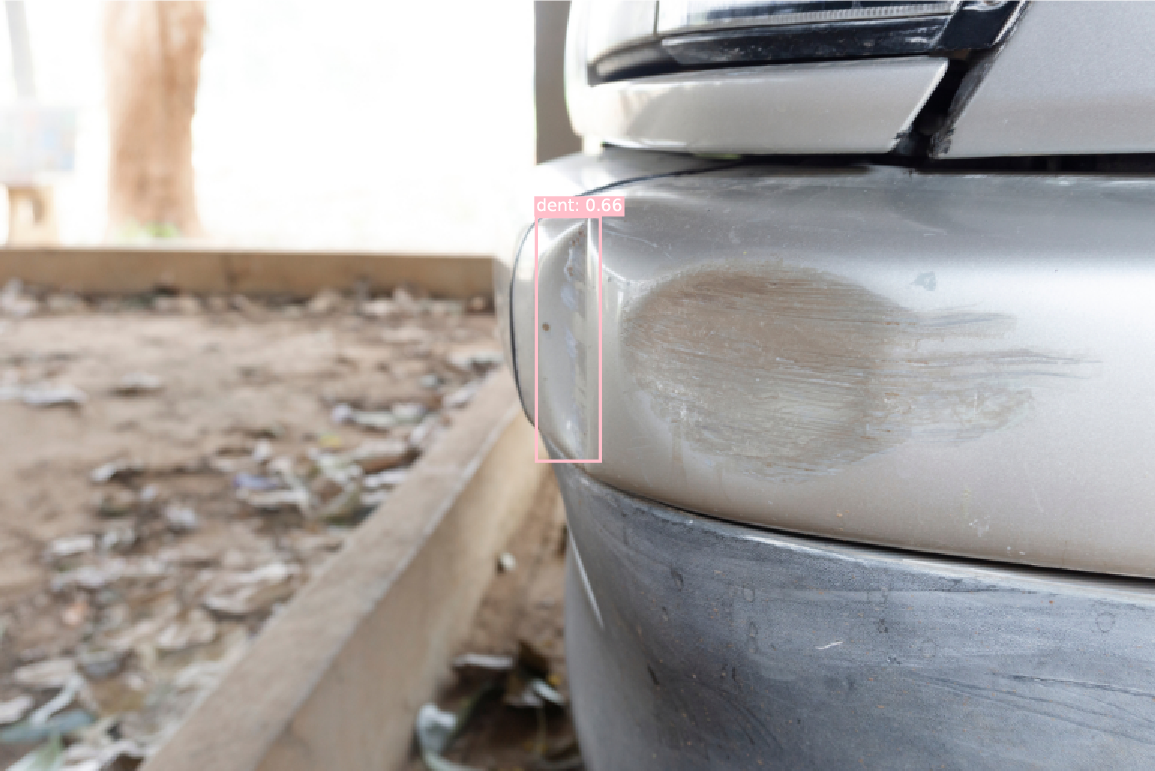}\end{subfigure}%
        \begin{subfigure}[b]{0.2\linewidth}\includegraphics[width=\textwidth]{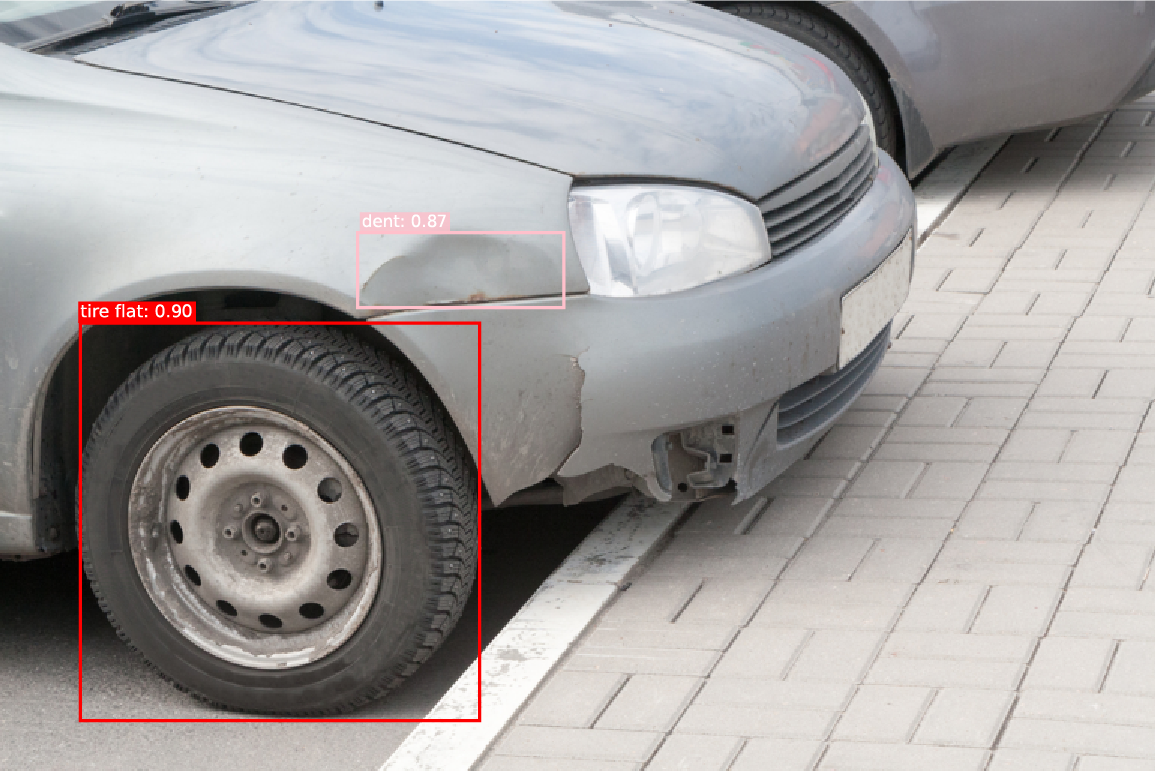}\end{subfigure}%
        \begin{subfigure}[b]{0.2\linewidth}\includegraphics[width=\textwidth]{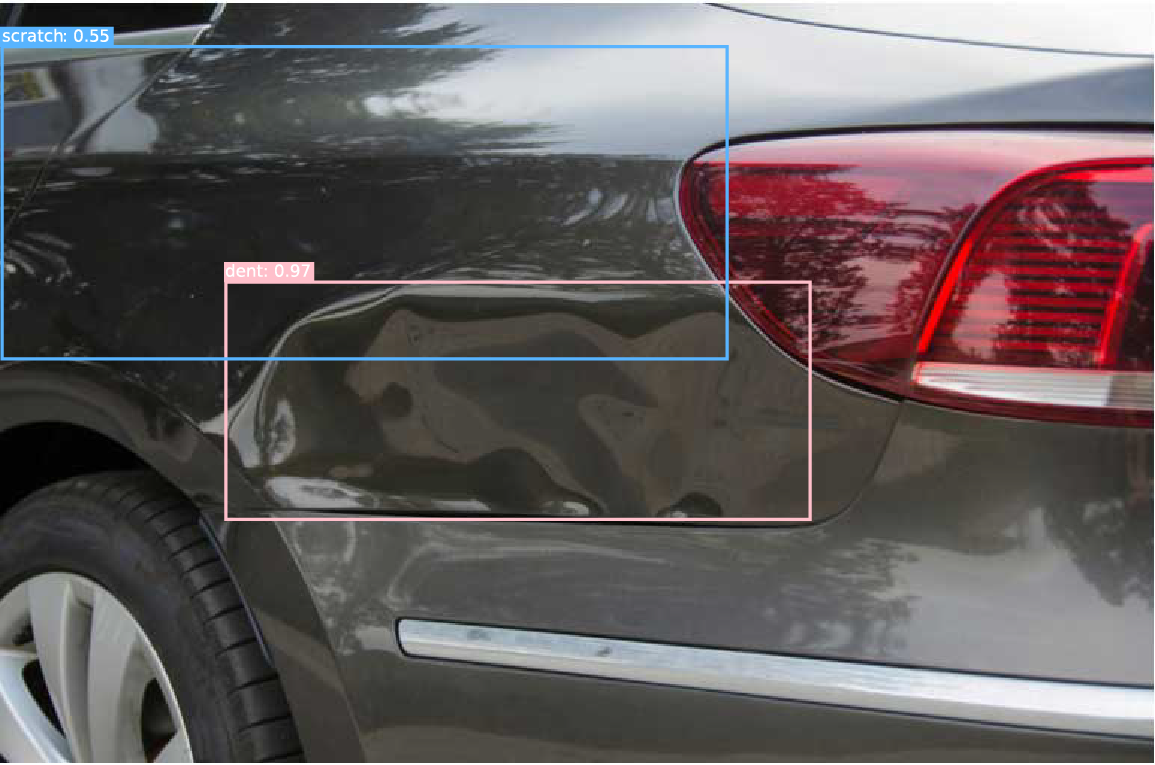}\end{subfigure}%
        \begin{subfigure}[b]{0.2\linewidth}\includegraphics[width=\textwidth]{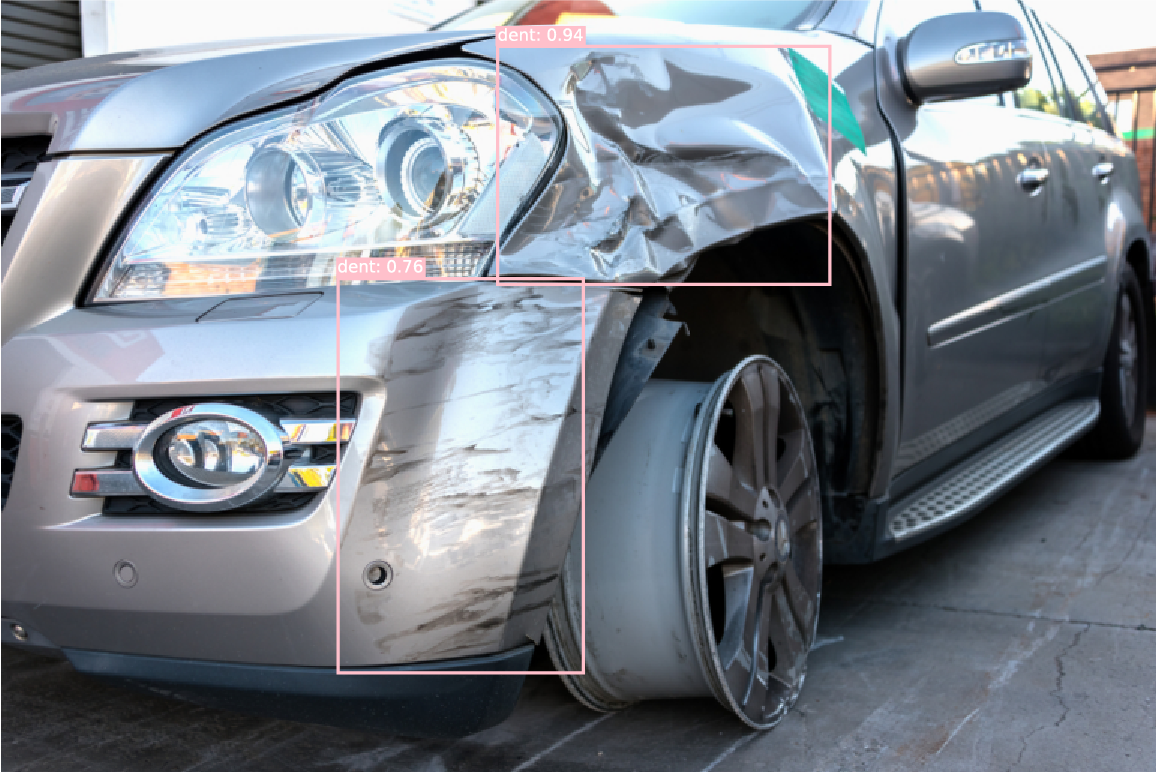}\end{subfigure}
    \end{tabular}

    \begin{tabular}{ >{\centering\arraybackslash}m{0.0\textwidth} m{1\textwidth} }
        \rotatebox{90}{\small DiffDet(Base)} &
        \begin{subfigure}[b]{0.2\linewidth}\includegraphics[width=\textwidth]{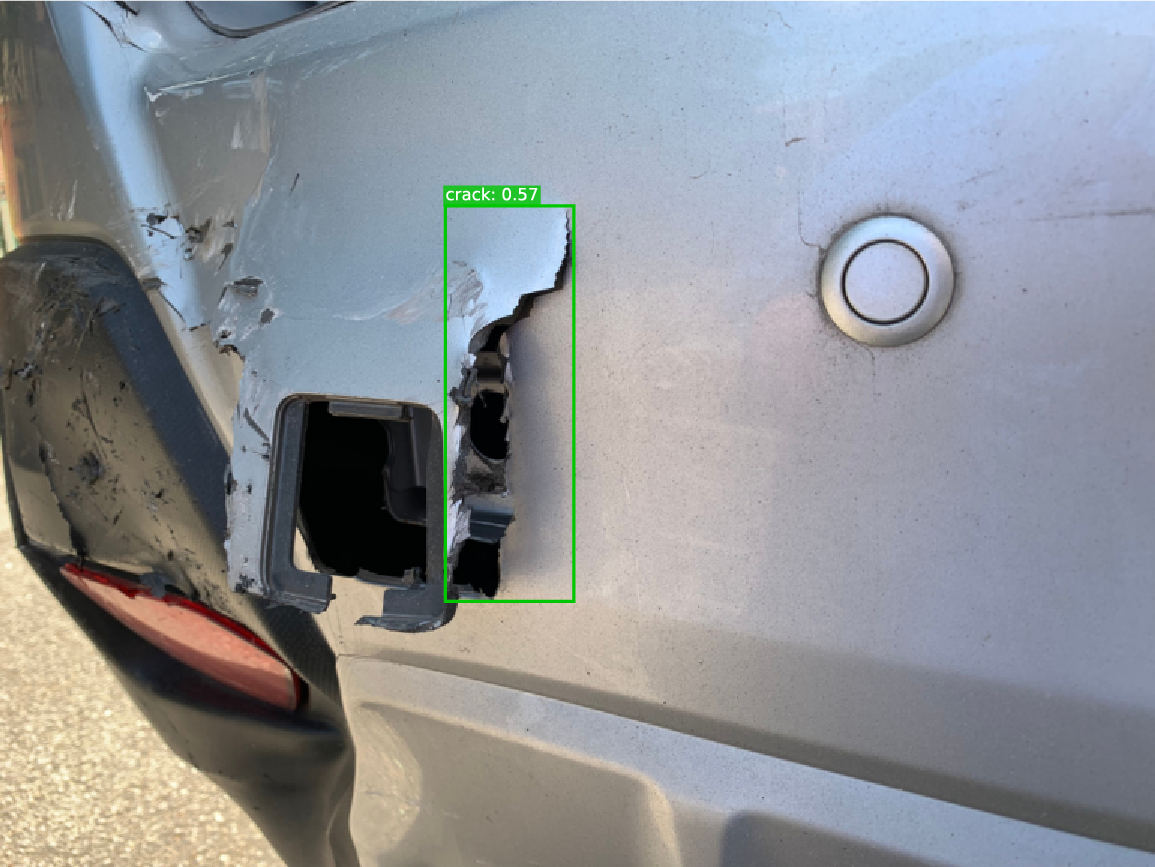}\end{subfigure}%
        \begin{subfigure}[b]{0.2\linewidth}\includegraphics[width=\textwidth]{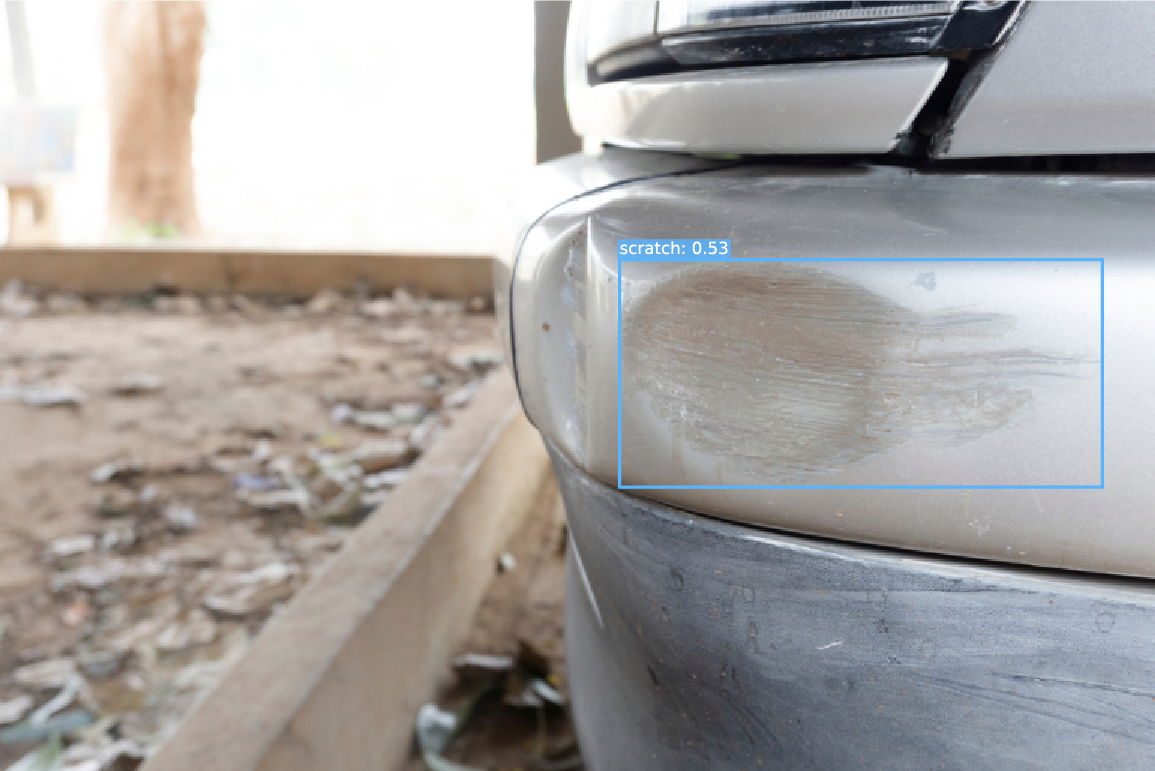}\end{subfigure}%
        \begin{subfigure}[b]{0.2\linewidth}\includegraphics[width=\textwidth]{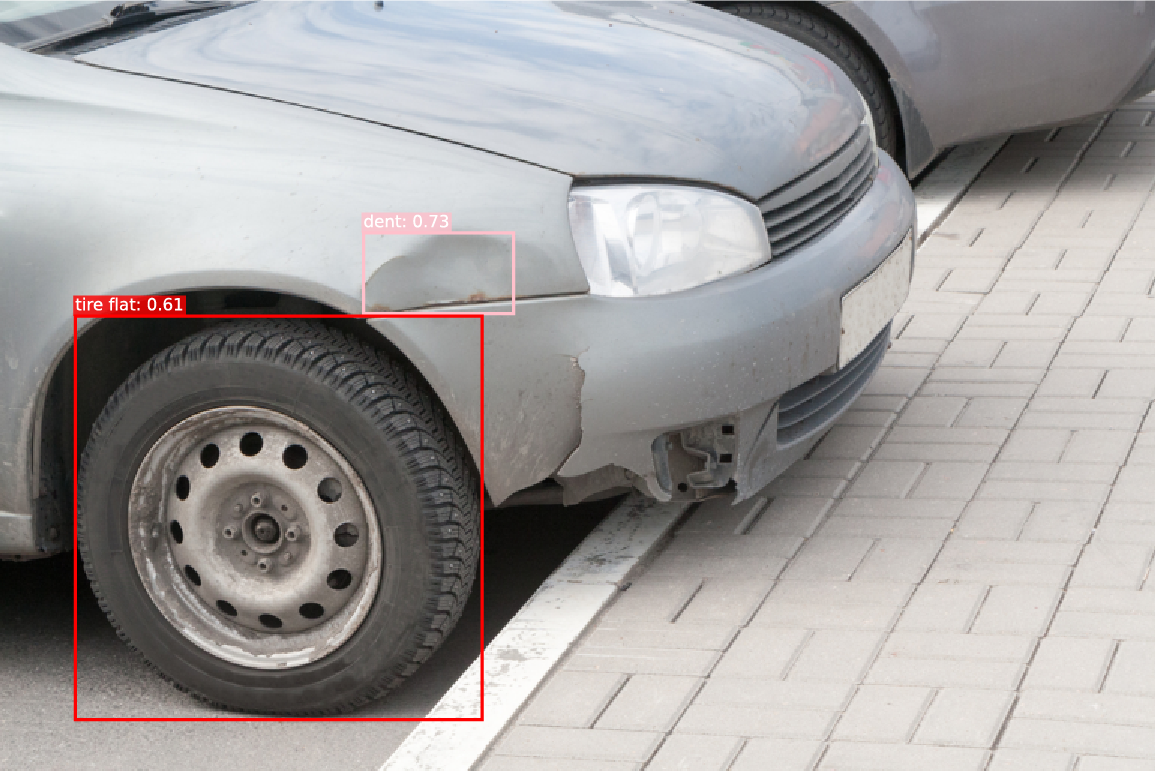}\end{subfigure}
        \begin{subfigure}[b]{0.2\linewidth}\includegraphics[width=\textwidth]{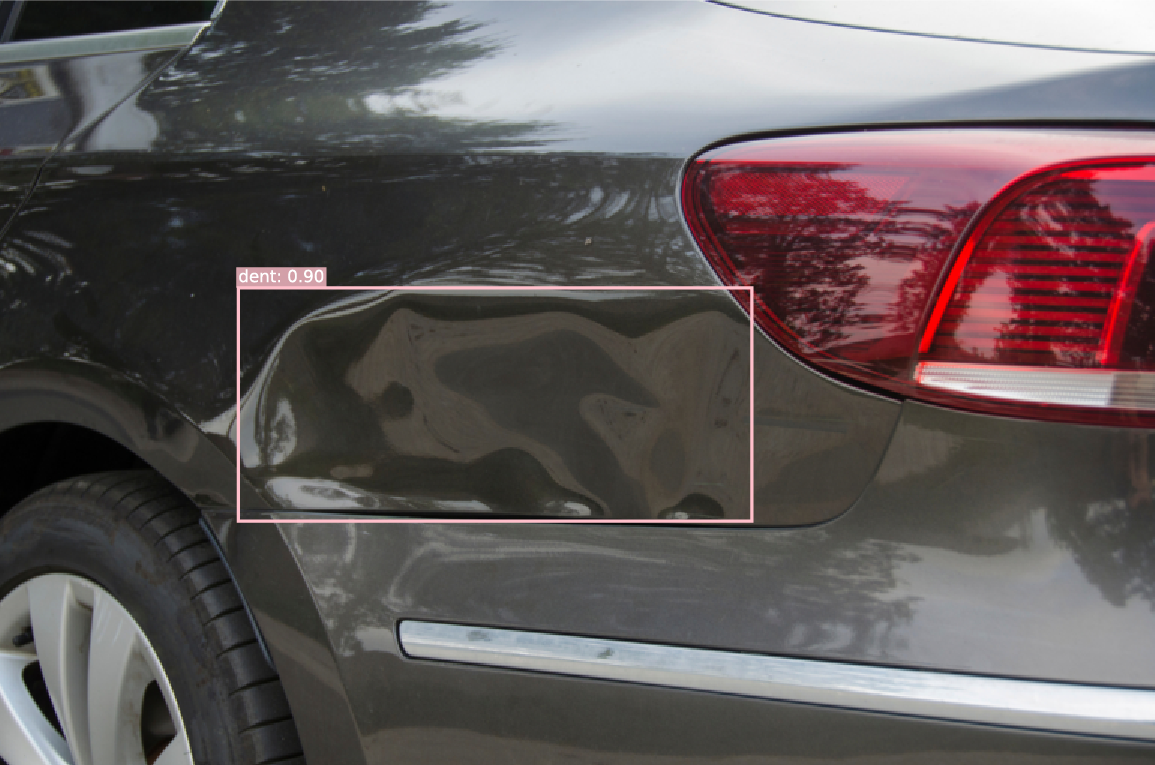}\end{subfigure}%
        \begin{subfigure}[b]{0.2\linewidth}\includegraphics[width=\textwidth]{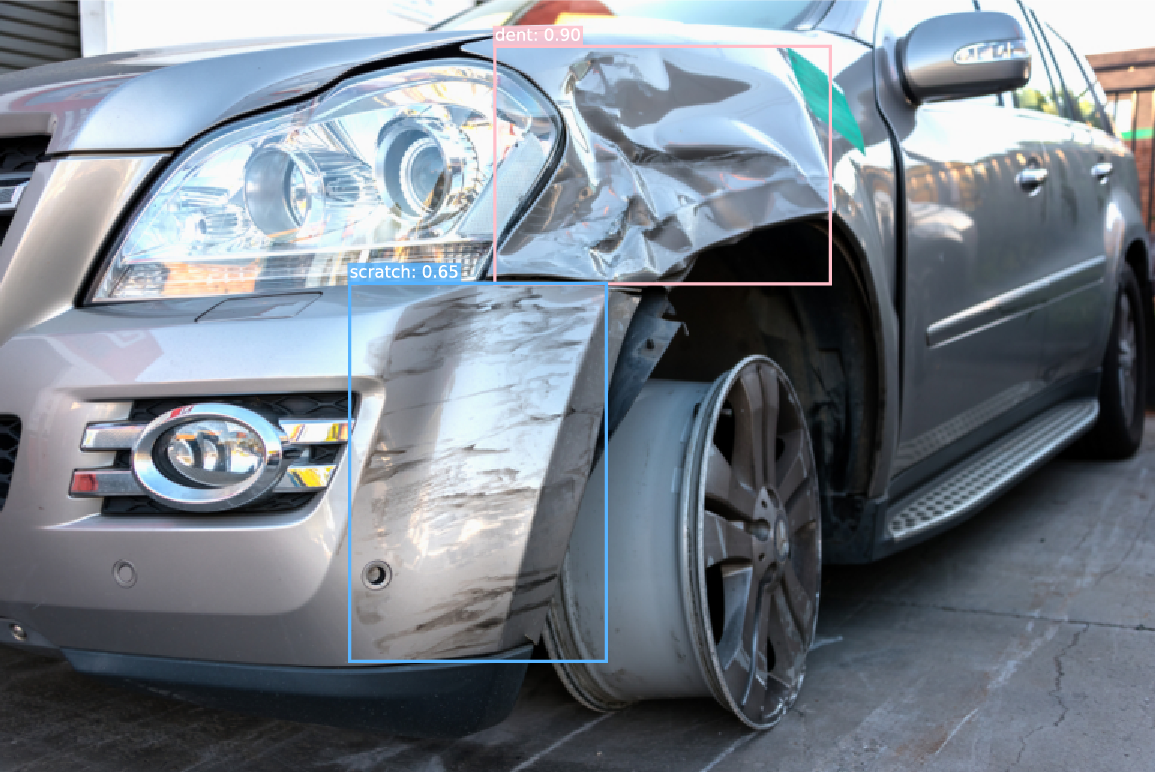}\end{subfigure}
    \end{tabular}

    \begin{tabular}{ >{\centering\arraybackslash}m{0.0\textwidth} m{1\textwidth} }
        \rotatebox{90}{\small Ours} &
        \begin{subfigure}[b]{0.2\linewidth}\includegraphics[width=\textwidth]{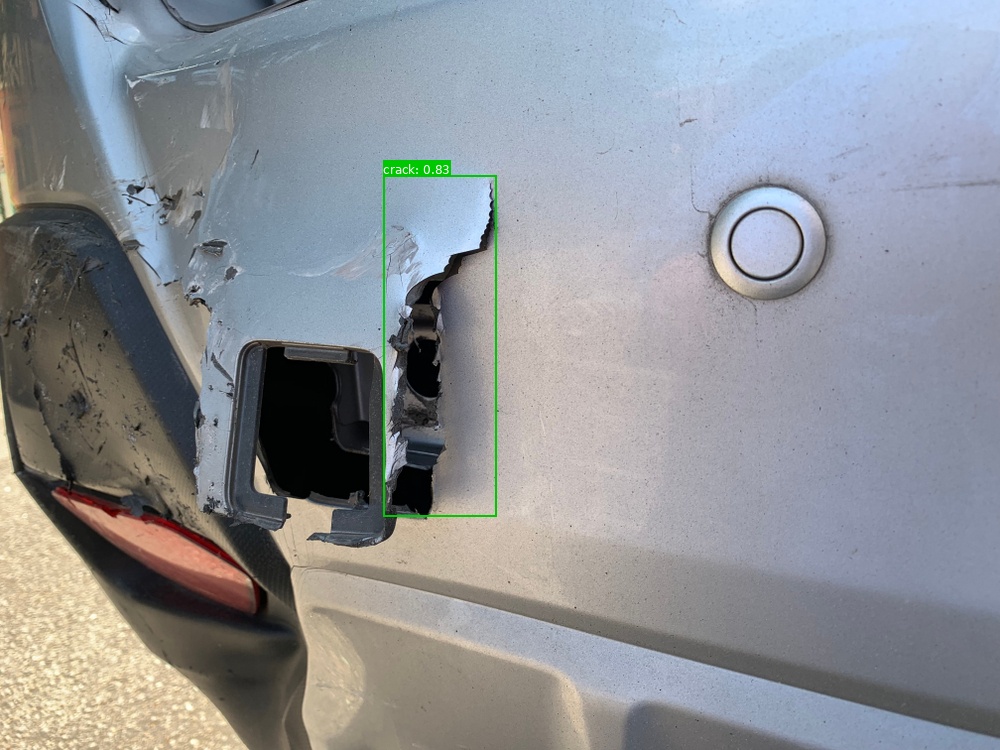}\end{subfigure}%
        \begin{subfigure}[b]{0.2\linewidth}\includegraphics[width=\textwidth]{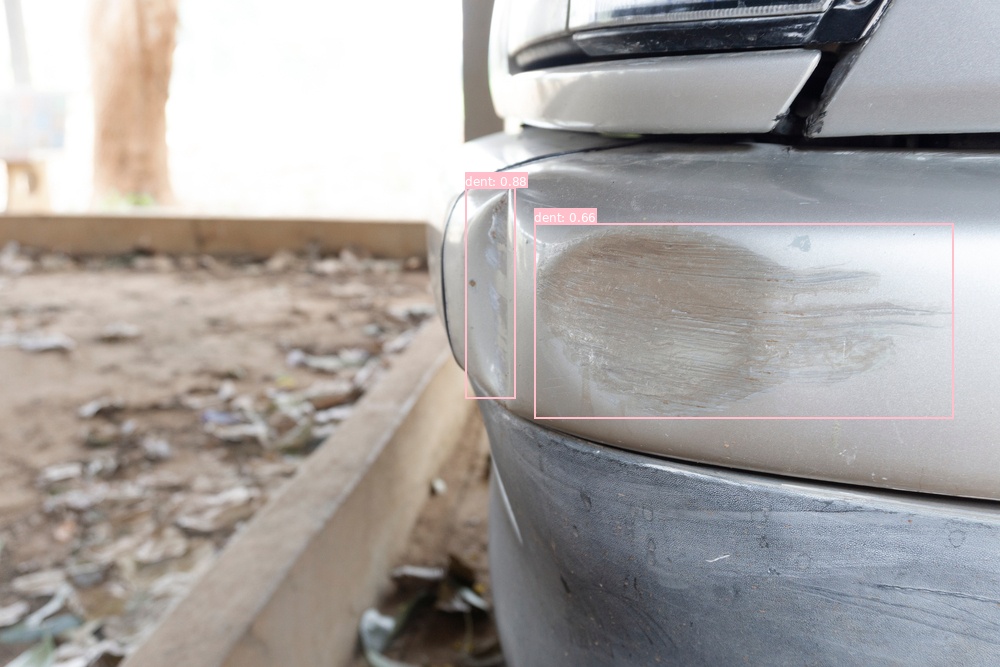}\end{subfigure}%
        \begin{subfigure}[b]{0.2\linewidth}\includegraphics[width=\textwidth]{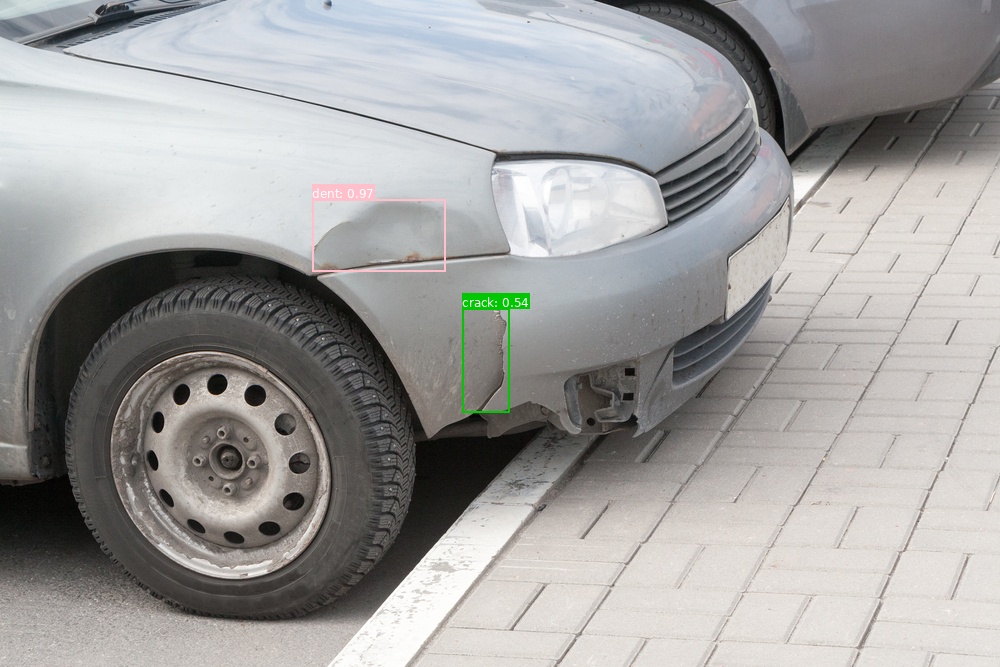}\end{subfigure}
        \begin{subfigure}[b]{0.2\linewidth}\includegraphics[width=\textwidth]{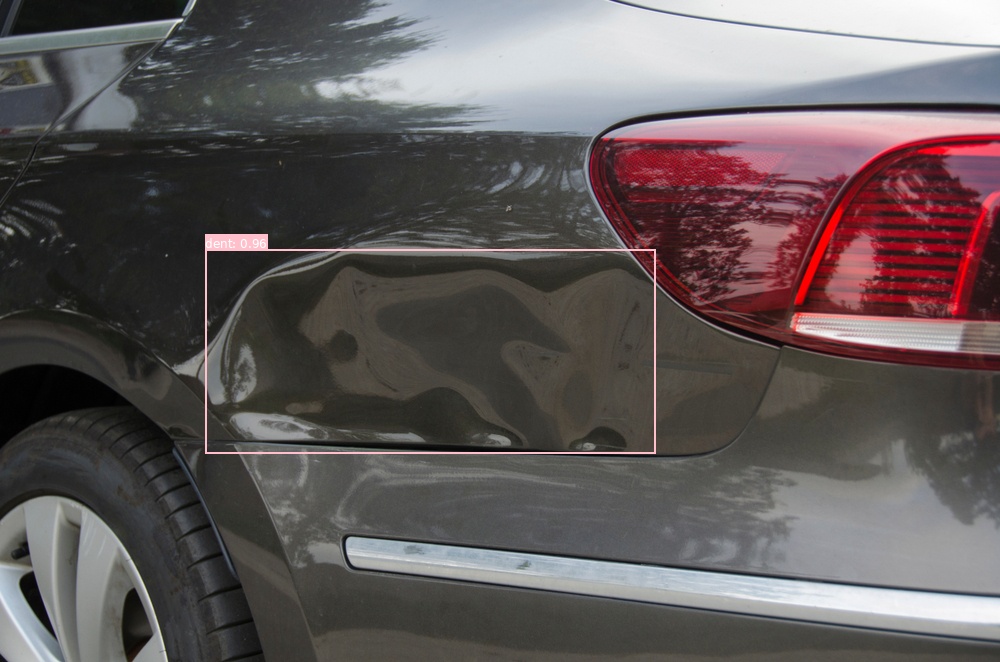}\end{subfigure}%
        \begin{subfigure}[b]{0.2\linewidth}\includegraphics[width=\textwidth]{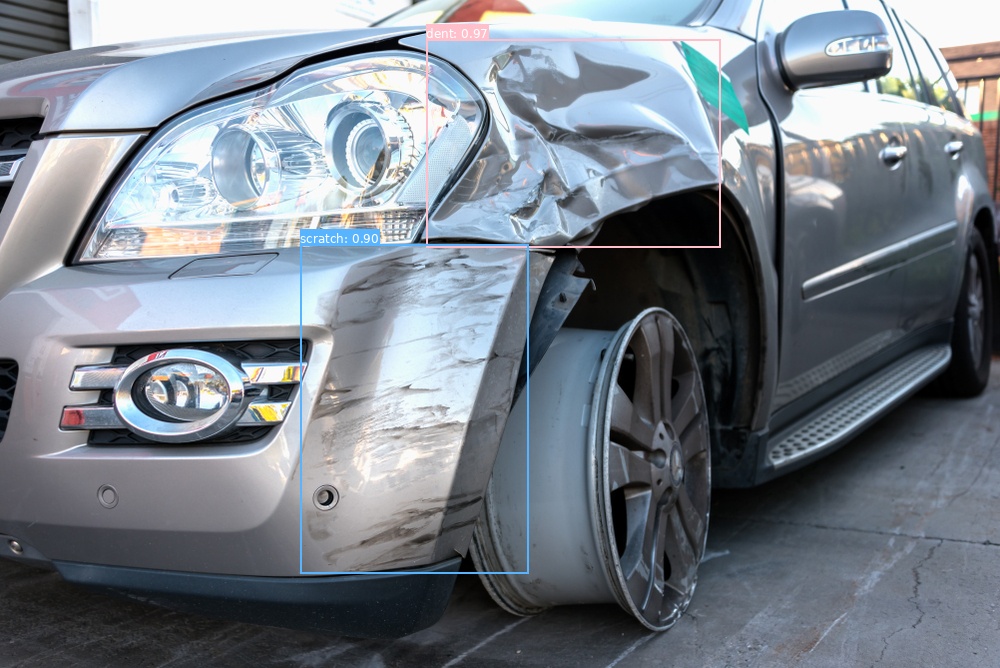}\end{subfigure}
    \end{tabular}

    \caption{This figure compares the performance of our model against the baseline. The first row contains the original images with the ground truth annotations. The second row shows the bounding boxes generated by DiffusionDet. The third row shows the more accurate bounding boxes produced by our enhanced model.}
   \label{img:examples-det}
\end{figure*}

A visual comparison of detection results in Figure~\ref{img:examples-det} reveals key insights about the performance of different methods. This analysis complements quantitative metrics by highlighting distinct failure modes and validating our fusion strategy.
\textbf{DCN+}, while effective for high-contrast damage through its deformable convolutions, shows limitations in information-poor scenarios. It often misses subtle defects and fragments contiguous damage due to insufficient global reasoning.
The \textbf{DiffusionDet} baseline demonstrates challenges with generative paradigms using only local features. While it occasionally improves recall, its predictions suffer from poor localization—producing oversized boxes that don't conform to damage geometry, as the denoising process lacks global contextual priors.
Our \textbf{C-DiffDet+} demonstrates transformative improvements through global-local fusion:
\begin{itemize}
    \item \textbf{Enhanced Precision and Localization:} Bounding boxes generated by C-DiffDet+ are consistently more precise and contour-adherent, particularly for challenging geometries like elongated scratches or irregular cracks. This is a direct consequence of the ACE-block sharpening high-level feature maps and the CAF mechanism using global context to resolve boundary ambiguities.
    
    \item \textbf{Enhanced Sensitivity to Challenging Instances:} Our model demonstrates a remarkable ability to detect faint, low-contrast damage that both DCN+ and DiffusionDet overlook. This is a hallmark of effective information fusion, where the global context prior provides a top-down signal that guides the localization of features which are statistically insignificant in isolation but meaningful within the broader scene context.
    
    \item \textbf{Increased Prediction Confidence:} Our model exhibits significantly higher and more consistent confidence scores in its predictions. This reduced epistemic uncertainty is a direct result of the disambiguating power provided by the global context embedding, transforming ambiguous guesses into certain, informed decisions.
\end{itemize}

These observations visually validate the quantitative gains in Tables~\ref{tab:bbox_ap_cardd_v2}--\ref{tab:class-wise-res-cardd}, confirming that fusing local evidence with global context is essential for robust fine-grained detection.

\subsubsection{Heatmaps Visualization}
\begin{figure}[h!]
    \centering
    
    \begin{tabular}{ >{\centering\arraybackslash}m{0.0\textwidth} m{1\textwidth} }

        \rotatebox{90}{\small Original} &
        \begin{subfigure}[b]{0.168\linewidth}\includegraphics[width=\textwidth]{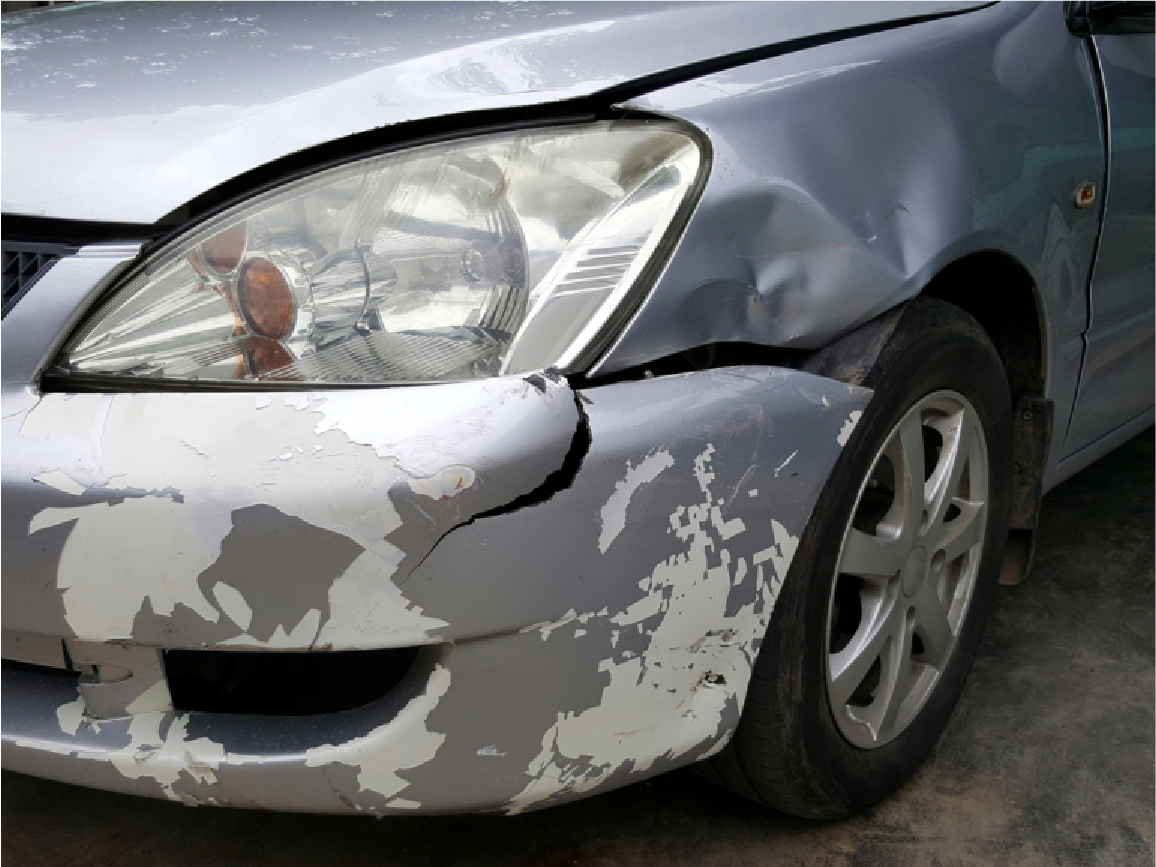}\end{subfigure}%
        \begin{subfigure}[b]{0.194\linewidth}\includegraphics[width=\textwidth]{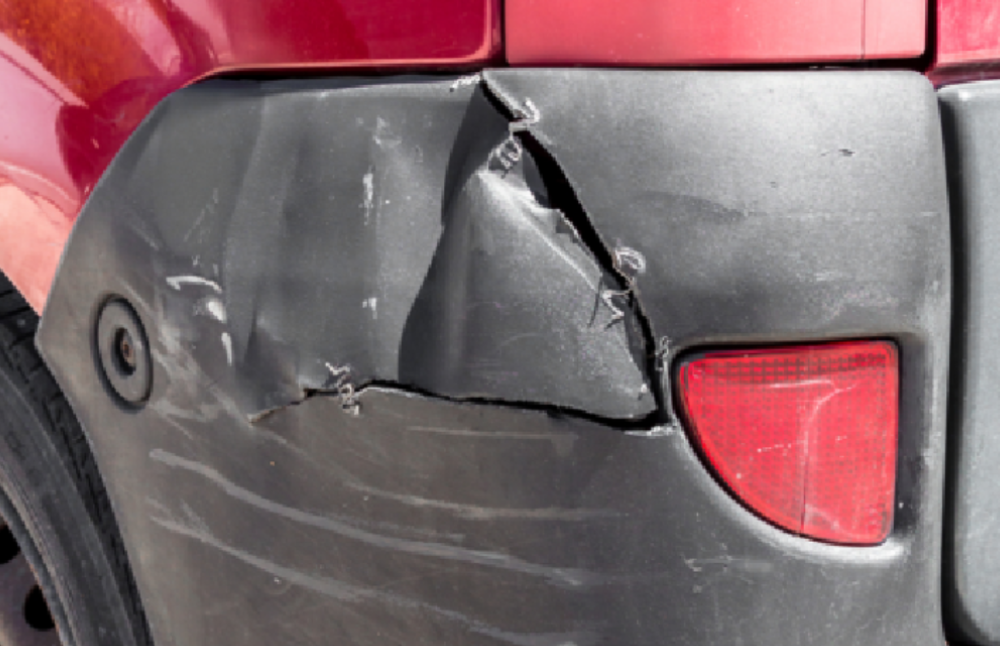}\end{subfigure}%
        \begin{subfigure}[b]{0.19\linewidth}\includegraphics[width=\textwidth]{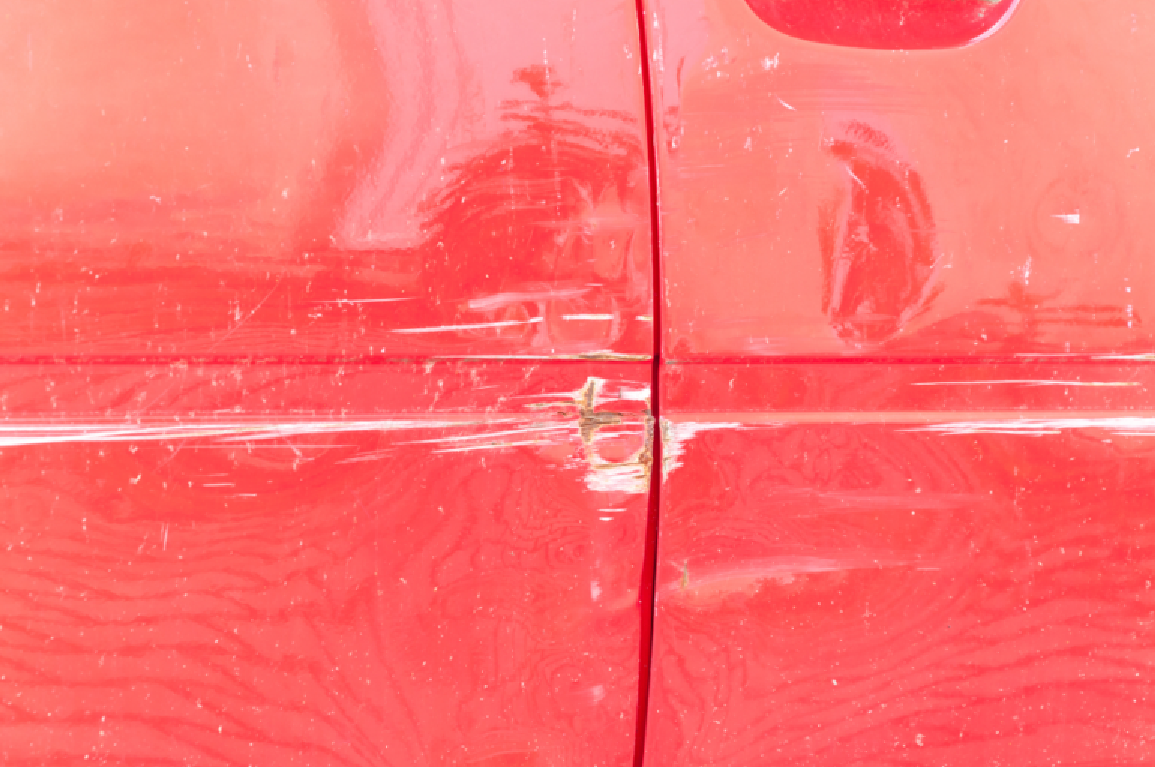}\end{subfigure}%
        \begin{subfigure}[b]{0.19\linewidth}\includegraphics[width=\textwidth]{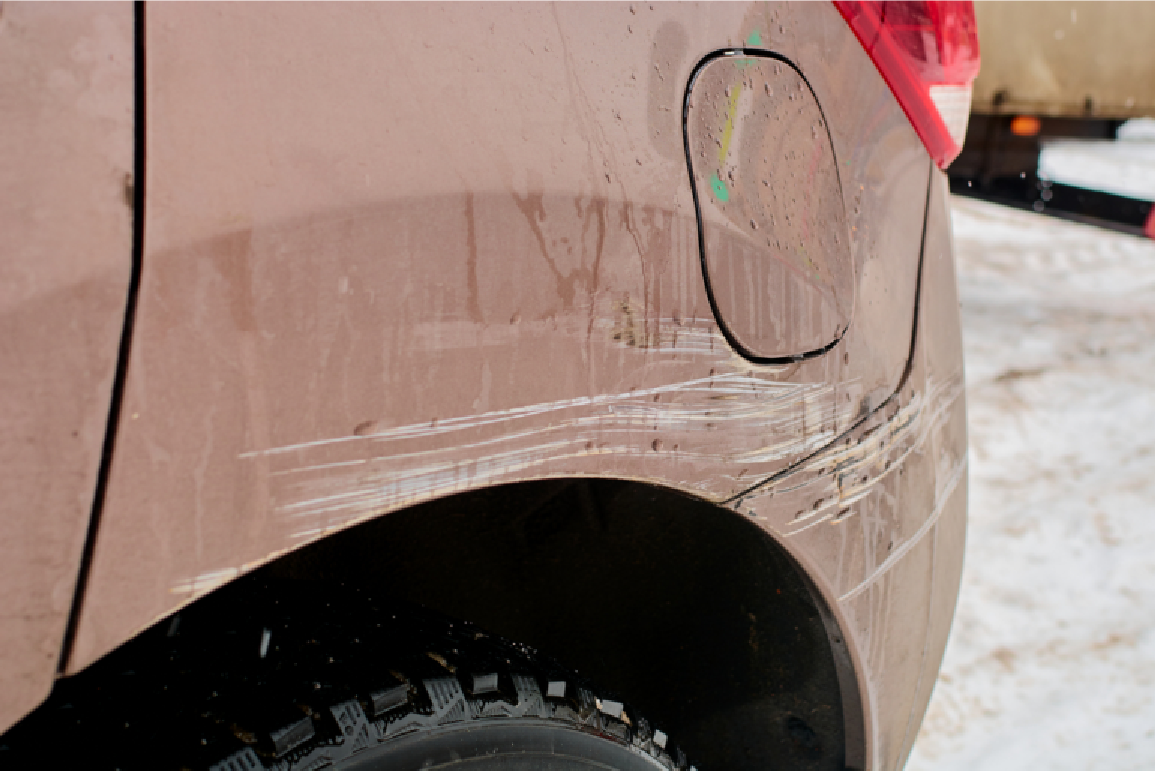}\end{subfigure}%
        \begin{subfigure}[b]{0.19\linewidth}\includegraphics[width=\textwidth]{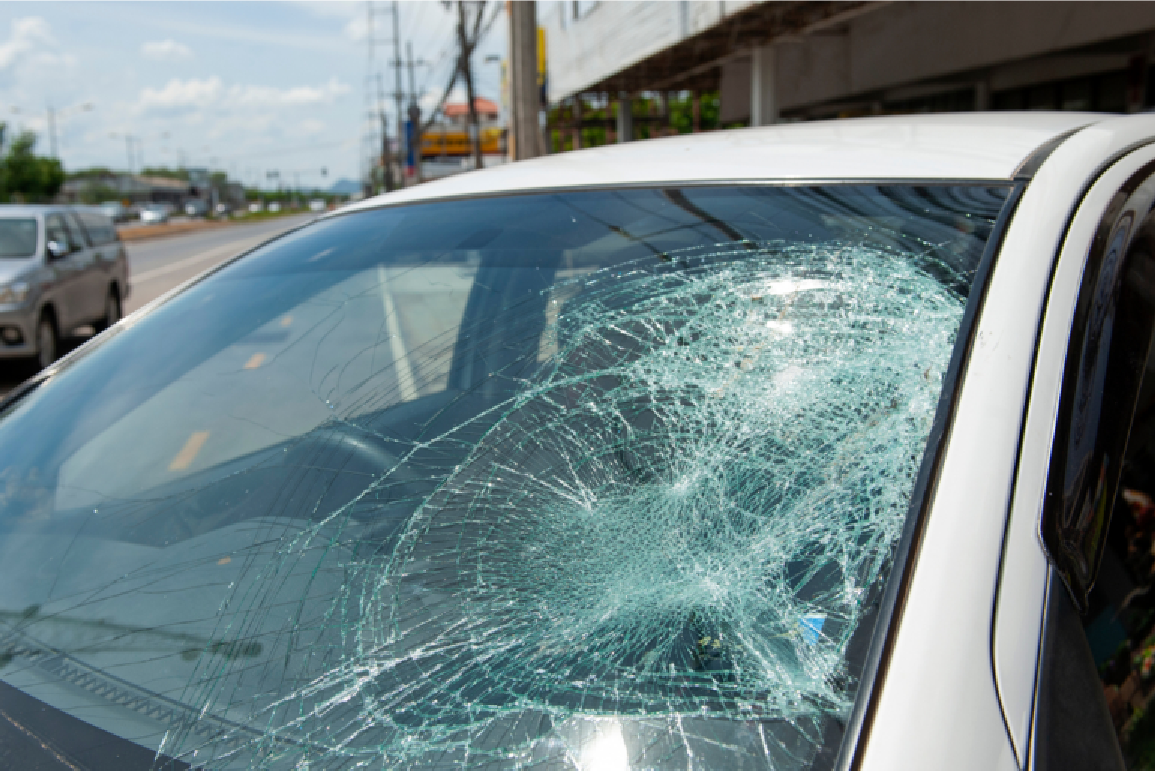}\end{subfigure} \\ 

        \rotatebox{90}{\small DiffusionDet} &
        \begin{subfigure}[b]{0.168\linewidth}\includegraphics[width=1\textwidth]{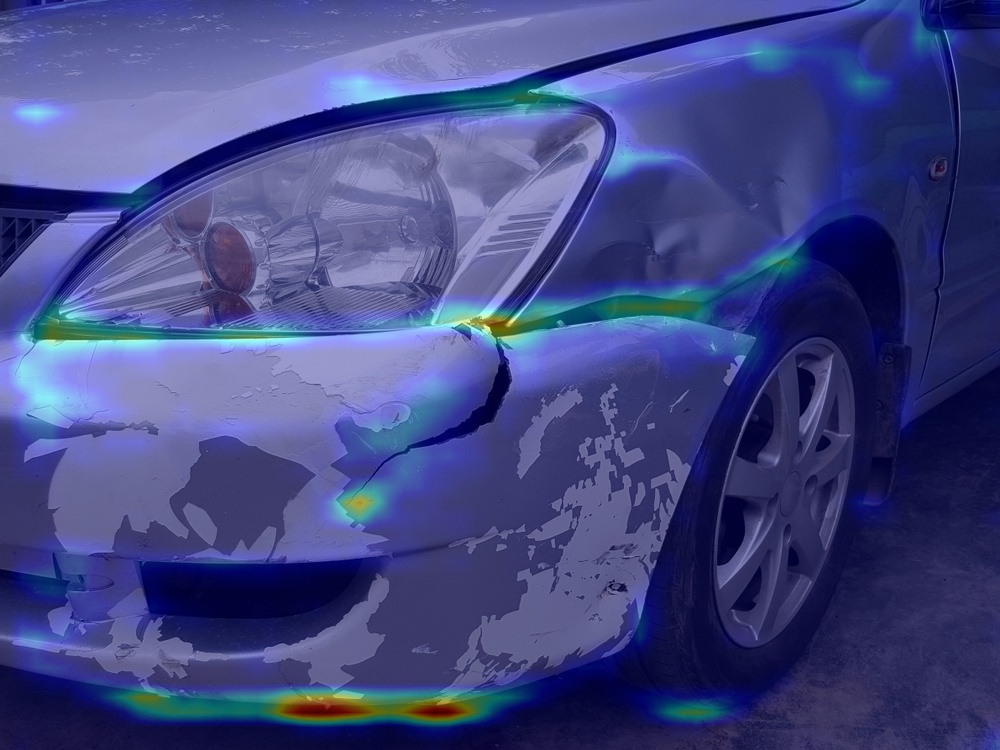}\end{subfigure}%
        \begin{subfigure}[b]{0.194\linewidth}\includegraphics[width=\textwidth]{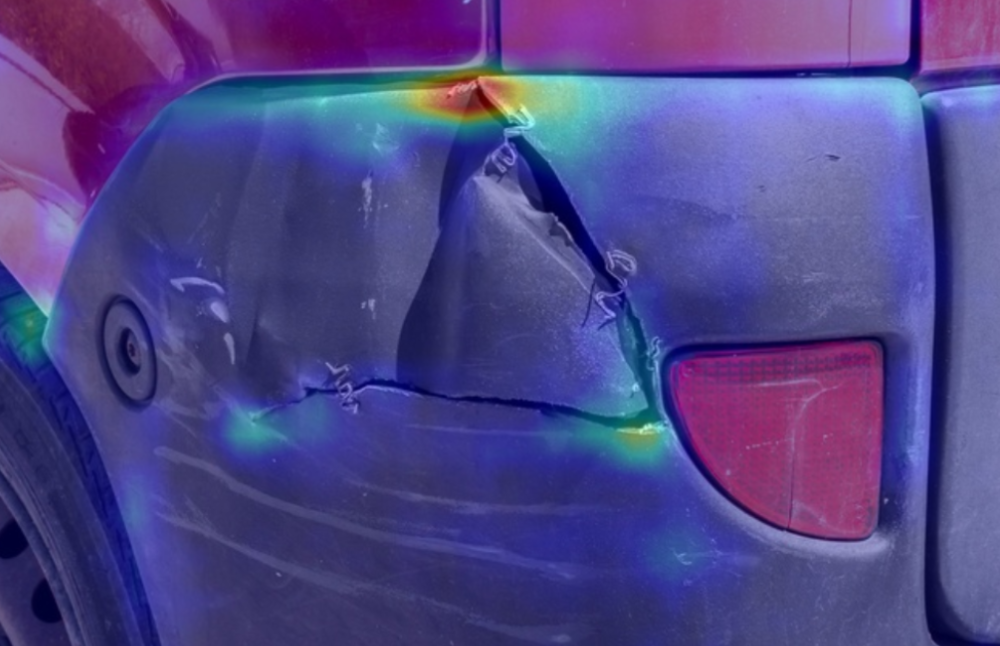}\end{subfigure}%
        \begin{subfigure}[b]{0.19\linewidth}\includegraphics[width=\textwidth]{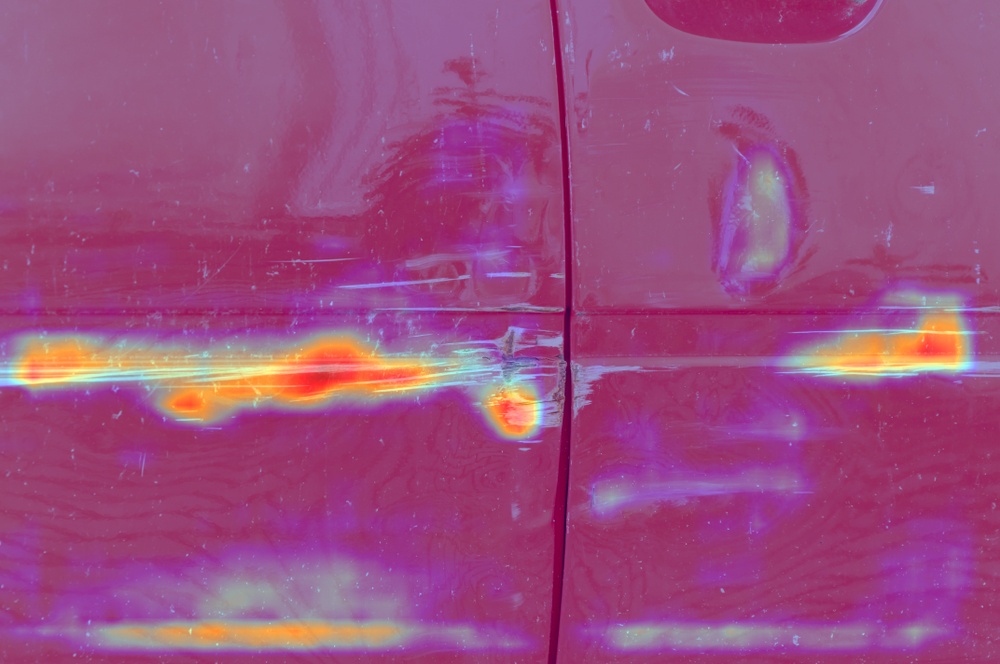}\end{subfigure}%
        \begin{subfigure}[b]{0.19\linewidth}\includegraphics[width=\textwidth]{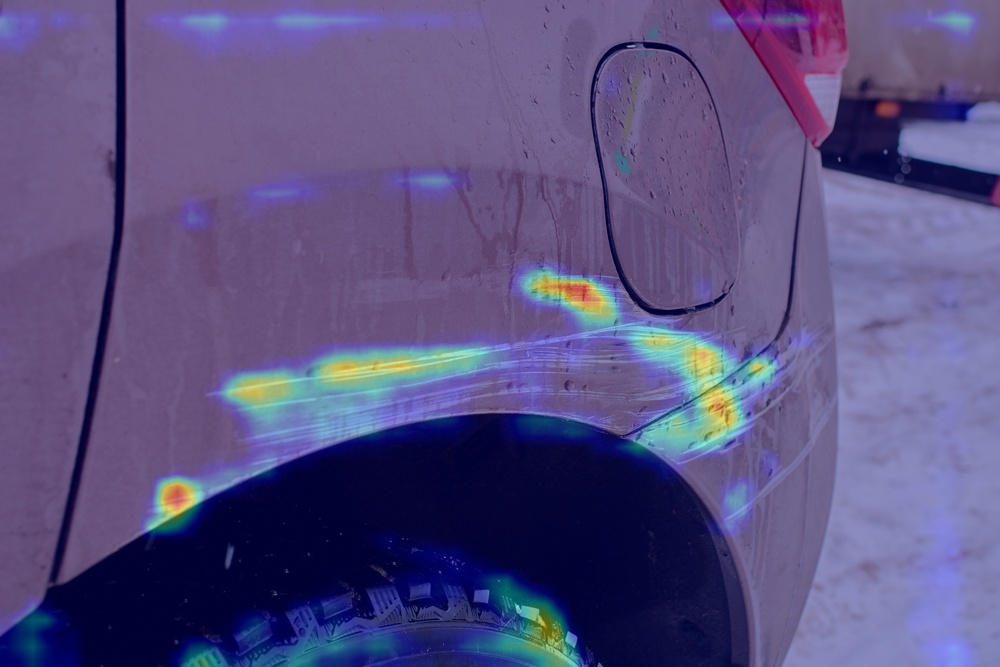}\end{subfigure}%
        \begin{subfigure}[b]{0.19\linewidth}\includegraphics[width=\textwidth]{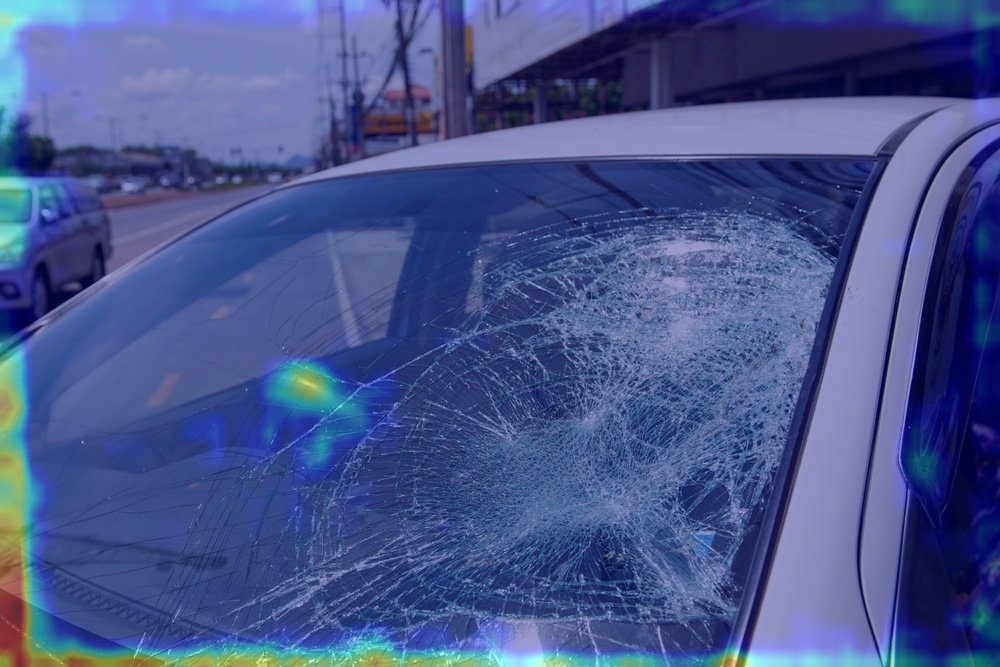}\end{subfigure} \\ 

        \rotatebox{90}{\small Ours} &
        \begin{subfigure}[b]{0.168\linewidth}\includegraphics[width=1\textwidth]{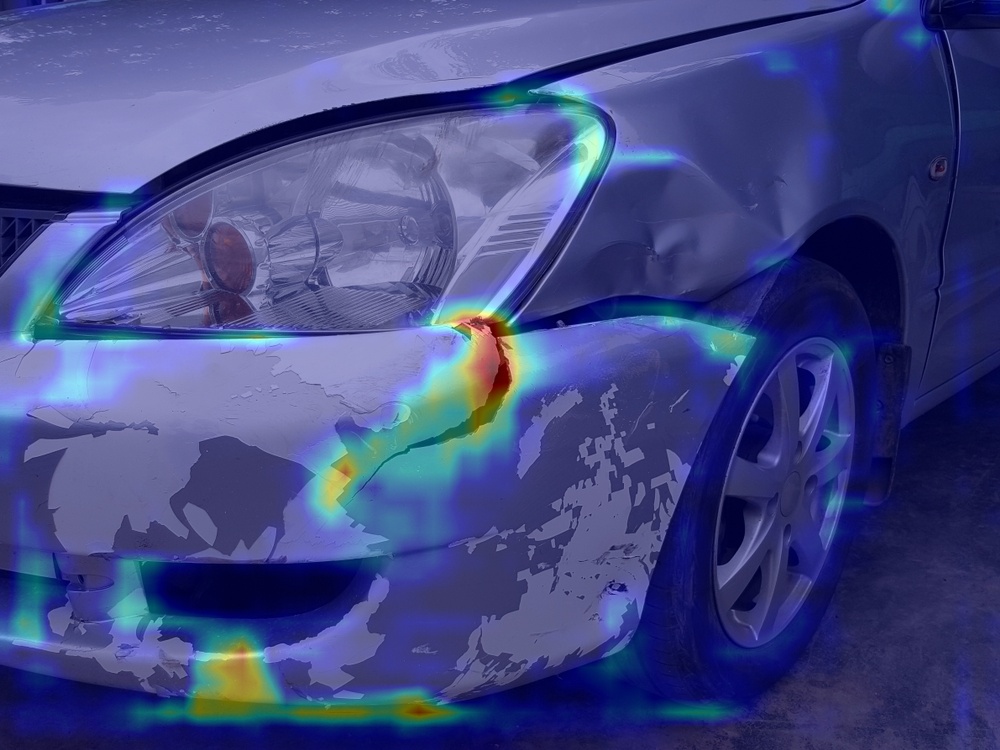}\end{subfigure}
        \begin{subfigure}[b]{0.194\linewidth}\includegraphics[width=\textwidth]{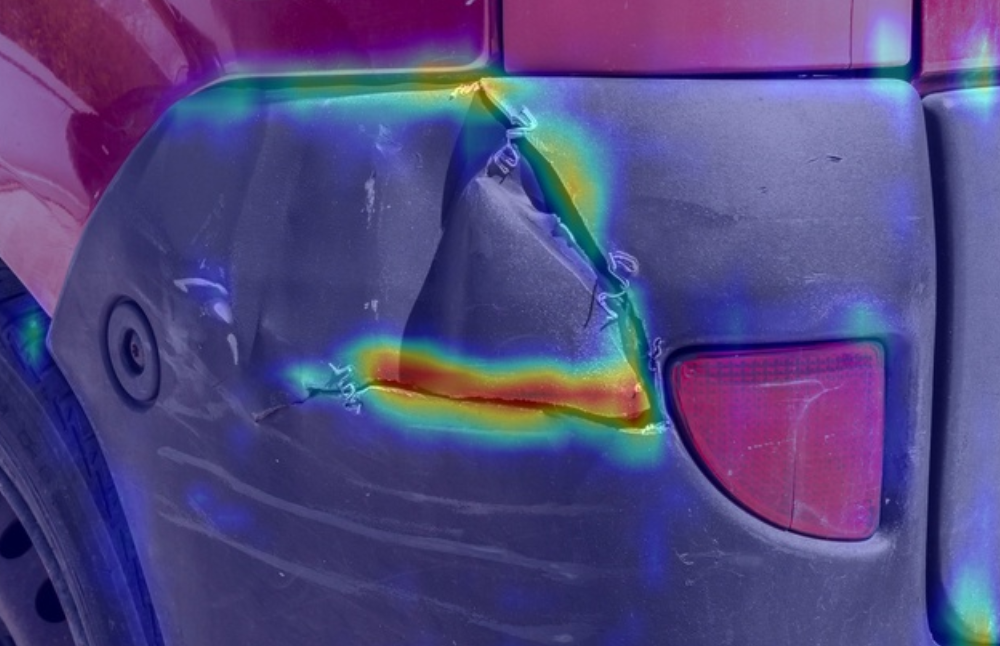}\end{subfigure}%
        \begin{subfigure}[b]{0.19\linewidth}\includegraphics[width=\textwidth]{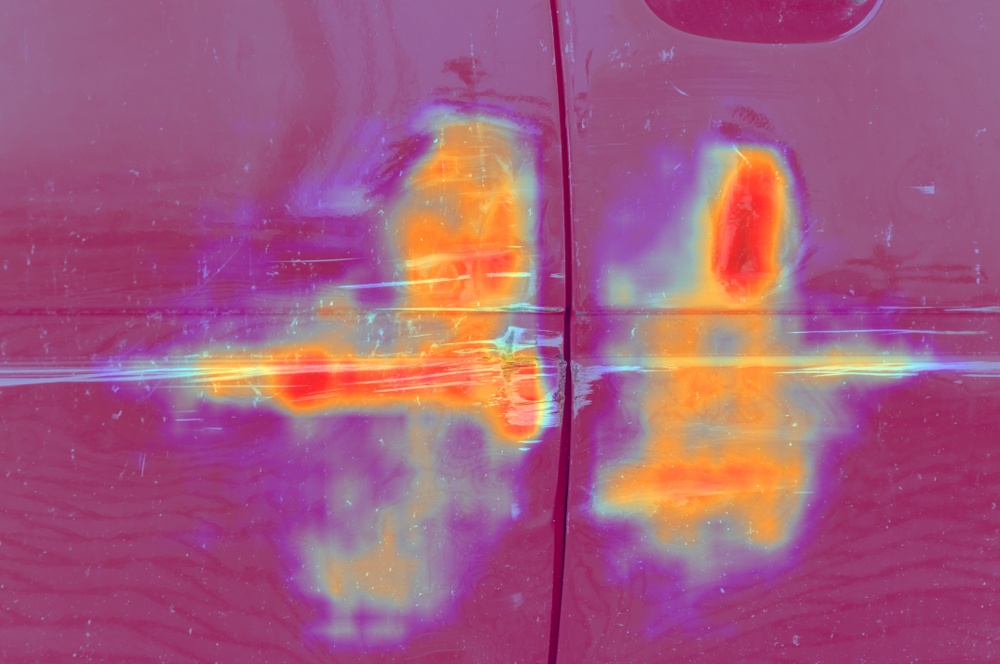}\end{subfigure}%
        \begin{subfigure}[b]{0.19\linewidth}\includegraphics[width=\textwidth]{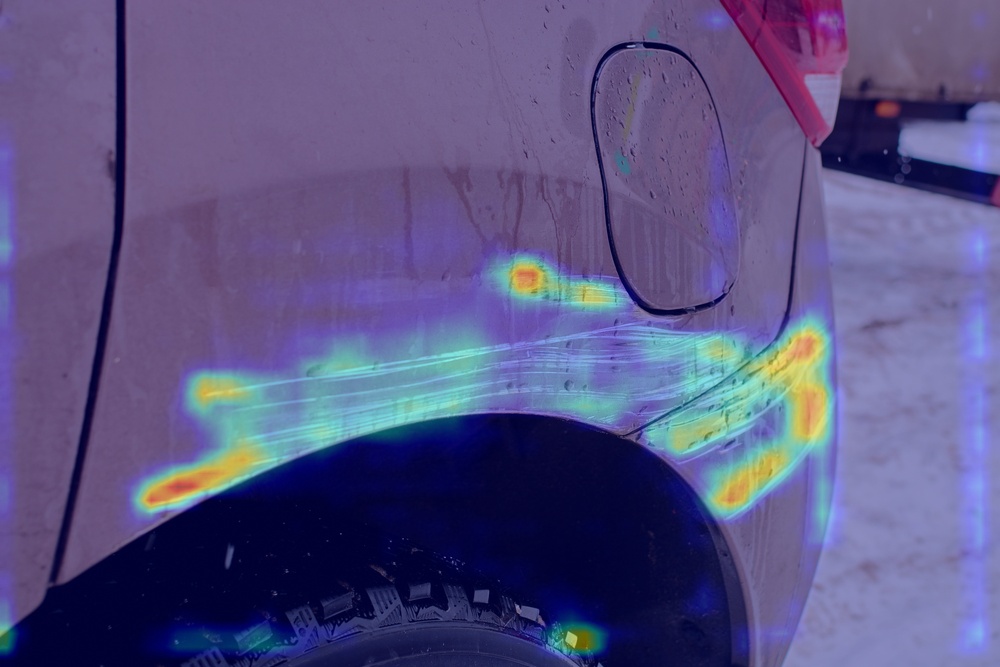}\end{subfigure}%
        \begin{subfigure}[b]{0.19\linewidth}\includegraphics[width=\textwidth]{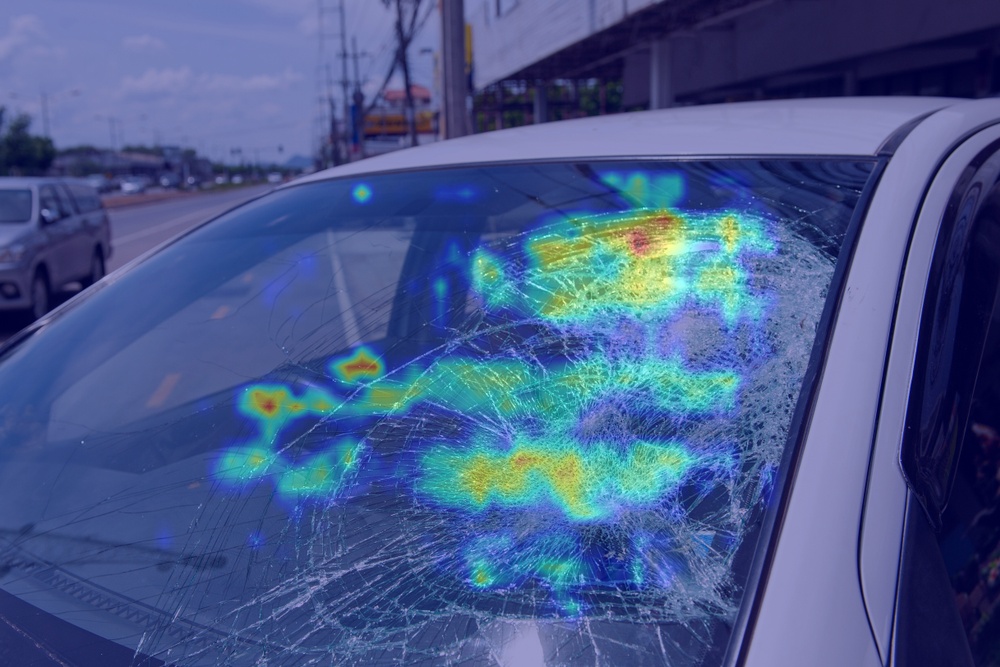}\end{subfigure}

    \end{tabular}

    \caption{The images display a visual comparison of two object detection models on Cardd dataset, DiffusionDet and our Model, for identifying car damage. The top row shows the original damaged car images. The middle row illustrates the heatmaps generated by the DiffusionDet model, which highlights areas it focuses on to detect damage. The bottom row presents the heatmaps from the Our Model combination, showing improved focus and accuracy in pinpointing the damaged regions.}
    \label{fig:vis_compa}
\end{figure}

Figure \ref{fig:vis_compa} compares the predictions of the DiffDet model with those of our C-DiffDet+ model using representative samples from the CarDD dataset. While DiffDet can generally indicate damage, its responses tend to be diffuse and not well-aligned with the accurate contours of defects. The heatmaps often cover large undamaged areas, and subtle features such as hairline cracks or fine scratches are fragmented or completely overlooked. This limitation arises from its reliance on locally adapted features, which specular highlights, surface textures, and background clutter can easily disrupt.

In contrast, the feature representations learned by C-DiffDet+ are significantly sharper and more defined. The activation maps are focused precisely on damaged regions, successfully capturing elongated scratches with continuous, well-defined responses and accurately tracing the jagged outlines of cracks. Low-contrast dents and punctures stand out better against their surroundings, enhancing their visibility even in challenging lighting conditions. Notably, the responses show much less overlap with undamaged areas, indicating that the model effectively suppresses noise while highlighting discriminative features.

These improvements are substantial: they directly lead to better localization and more reliable detection. The feature maps provide the detector with richer boundary information, allowing for tighter bounding boxes around irregular damage, higher recall rates for subtle instances, and more stable confidence estimates. This qualitative evidence aligns with the quantitative gains reported in Tables~\ref{tab:bbox_ap_cardd_v2}--\ref{tab:class-wise-res-cardd}, where C-DiffDet+ clearly outperforms DiffDet, especially in detecting fine-grained categories like cracks. This demonstrates that the enhanced representational quality of our model underlies its superior performance in real-world automotive damage assessment.

\section{Discussion}
\label{sec:disscusion}
\subsection{Ablation Studies}
\subsubsection{Quantitative Component Analysis}

A systematic ablation study was conducted to quantitatively deconstruct the contributions of each proposed module within the \textbf{C-DiffDet+} framework. The results, summarized in Table~\ref{tab:ablation}, demonstrate the synergistic effects achieved through the principled fusion of local feature enhancement and global contextual reasoning.

The initial augmentation of the baseline DiffusionDet model with our Adaptive Channel Enhancement (ACE) blocks yielded an mAP of 64.1\%, a modest but consistent gain of \textbf{+0.7\%} over the pure baseline (63.4\%). This improvement is primarily attributed to the mechanism's capacity for dynamic, channel-wise re-calibration, which amplifies the most salient features for damage detection. The efficacy of this local feature refinement is most pronounced in the detection of small ($\text{AP}_S$: \textbf{+2.8\%}) and medium ($\text{AP}_M$: \textbf{+1.2\%}) objects, affirming that adaptive channel attention is critical for resolving fine-grained visual patterns.

Introducing the Context-Aware Fusion (CAF) module---without global context embeddings---to the ACE-enhanced baseline maintained the overall mAP at 64.1\% but induced a significant redistribution of performance across categories. Notably, detection accuracy for \textit{glass shatter} and \textit{lamp broken} increased by \textbf{+4.1\%} and \textbf{+2.8\%}, respectively. This suggests the cross-attention mechanism within CAF successfully enables inter-proposal communication. However, the absence of a dedicated global signal proved detrimental to small object detection ($\text{AP}_S$: \textbf{-20.6\%}), underscoring that local-to-local interaction alone is insufficient without top-down contextual guidance.

The integration of global context embeddings (\textit{Global\_emb}), while maintaining the CAF module but removing ACE, resulted in a competitive mAP of 64.0\%. This configuration exhibited remarkable proficiency in disambiguating challenging categories, evidenced by a \textbf{+6.1\%} increase in \textit{crack} detection. This finding strongly supports our core hypothesis: a compact, scene-level representation is paramount for resolving local ambiguities. The performance ceiling in other categories, however, highlights the complementary role of local feature enhancement via ACE.

The full \textbf{C-DiffDet+} model, integrating ACE, \textit{Global\_emb}, and CAF, achieves optimal performance with an mAP of \textbf{64.8\%}. This configuration establishes new state-of-the-art results in key metrics, including small object detection ($\text{AP}_S$: 45.5\%) and crack detection (42.2\%). The observed synergy is non-linear; the whole is greater than the sum of its parts. The ACE module ensures high-quality, discriminative local features. The Global Context Encoder provides a rich, scene-level prior. Finally, the CAF mechanism acts as the crucial information fusion engine, seamlessly integrating these multi-scale representations through cross-attention to inform the generative denoising process.

\noindent\textbf{In summary}, this ablation study provides conclusive evidence for our architectural design. It validates that the fusion of local feature enhancement (ACE), global context encoding, and context-aware cross-attention (CAF) addresses a fundamental limitation in existing generative detectors. The results confirm that robust fine-grained detection in complex scenes is not merely a function of iterative refinement but is critically dependent on a tightly-coupled, multi-source information fusion strategy.

\begin{table}[]
\centering
\resizebox{\textwidth}{!}{
\begin{tabular}{cccc|cccc|cccccc}
\hline
\textbf{Baseline} & \textbf{ACE} & \textbf{Global\_emb} & \textbf{CAF} & \textbf{AP} & \textbf{AP\(_S\)} & \textbf{AP\(_M\)} & \textbf{AP\(_L\)} & \textbf{dent} & \textbf{scratch} & \textbf{crack} & \textbf{glass} & \textbf{lamp} & \textbf{tire} \\ \hline
\ding{51} & \ding{55} & \ding{55} & \ding{55} & 63.4 & 38.7 & 36.1 & 64.2 & 38.9 & \textbf{44.8} & 35.1 & 92.6 & 76.8 & \underline{92.5} \\
\ding{51} & \ding{51} & \ding{55} & \ding{55} & 64.1 & \underline{41.5} & \underline{37.3} & 64.4 & \textbf{39.5} & \underline{44.2} & 36.5 & 93.2 & 80.0 & 90.4 \\
\ding{51} & \ding{55} & \ding{55} & \ding{51} & 64.1 & 18.8 & 27.1 & 65.3 & 37.3 & 43.2 & 35.6 & \underline{96.7} & 79.6 & 91.9 \\
\ding{51} & \ding{51} & \ding{55} & \ding{51} & \underline{64.3} & 20.9 & 29.2 & \underline{65.5} & 35.9 & 42.8 & 34.7 & \textbf{97.2} & \textbf{82.5} & \textbf{92.7} \\
\ding{51} & \ding{55} & \ding{51} & \ding{51} & 64.0 & 35.3 & 36.4 & 64.4 & \underline{39.1} & 40.7 & \underline{41.2} & 95.3 & 75.6 & 91.9 \\
\ding{51} & \ding{51} & \ding{51} & \ding{51} & \textbf{64.8} & \textbf{45.5} & \textbf{39.2} & \textbf{65.8} & 37.0 & 42.6 & \textbf{42.2} & 94.4 & \underline{80.2} & 92.0 \\ \hline
\end{tabular}
}
\caption{Ablation study results. \ding{51} indicates presence, \ding{55} absence. Best results are in \textbf{bold}, second best are \underline{underlined}.}
\label{tab:ablation}
\end{table}

\subsubsection{Global context encoder}

Table \ref{tab:ablation_formatted}presents a comprehensive ablation study that evaluates the effectiveness of different global context encoder architectures in our C-DiffDet+ framework. The results demonstrate the critical importance of encoder design choices and provide quantitative evidence for the superiority of our proposed residual CNN encoder approach.

The Residual CNN Encoder achieves the highest overall performance, with an Average Precision (AP) of 64.8\%, and shows excellent results across various damage categories, confirming our architectural design choice. This encoder architecture effectively combines the computational efficiency of convolutional operations with the representational strength of residual connections (see Figure \ref{fig:GCE-map}). This combination allows for the effective capture of multi-scale spatial information while ensuring stable gradient flow. Using residual connections is especially beneficial for automotive damage detection, where identifying fine-grained features at different scales is essential for accurate classification and localization of damage.

The Simple CNN Encoder shows competitive performance, achieving an average precision (AP) of 64.6\%. It ranks second-best in several categories, including dent detection 40.1\%, which is the best, scratch detection 42.9\%, which is the best, and medium object detection, with an average precision of 39.3\% (the highest score in that category). This performance indicates that a basic convolutional encoder can effectively capture global scene context when integrated with a context-aware fusion mechanism. The architecture's success is due to its ability to learn hierarchical feature representations without the computational overhead associated with more complex designs. This makes it an ideal choice for scenarios where computational efficiency is important.

The ViT Encoder exhibits a significant decline in performance, achieving an average precision (AP) of 59.0\%, which is a decrease of 5.8\% from the best result. This drop in performance is evident across all metrics, indicating that transformer-based architectures are not well-suited for capturing global context in automotive damage detection. Several factors contribute to this performance decline, including the absence of inductive biases for spatial relationships that are typically found in convolutional architectures, the quadratic computational complexity of the model that limits its effective receptive field, and the lack of hierarchical feature extraction, which is essential for detecting damage at multiple scales. Additionally, the ViT's patch-based approach may struggle to accurately capture the fine-grained spatial relationships needed for a thorough damage assessment.

The Mamba Vision Encoder exhibits the lowest performance, with an Average Precision (AP) of 54.7\%, which is 10.1\% lower than the best-performing model. This indicates that state-space model architectures are ineffective for encoding global context in this domain. Although the Mamba architecture offers linear complexity, it struggles with limited spatial modeling capabilities. This shortcoming is particularly detrimental for automotive damage detection, where understanding spatial relationships and integrating multi-scale features are essential. The performance decline is especially significant in small object detection, achieving an AP of 31.8\% (13.7\% lower than the best model), and for fine-grained damage types like cracks, which have an AP of 24.4\% (17.8\% lower than the best).

The ablation results reveal several key insights about global context encoder design for automotive damage detection. First, convolutional architectures consistently outperform transformer and state-space model approaches, demonstrating that inductive biases for spatial relationships are crucial for effective global context encoding. Second, residual connections provide significant benefits, as evidenced by the superior performance of the Residual CNN Encoder across most metrics. The residual connections enable deeper networks with stable training, leading to better feature representation quality and improved damage detection accuracy.

\begin{table}[h]
\centering
\caption{Ablation study results with the best value in each column bolded and the second best underlined.}
\label{tab:ablation_formatted}
\resizebox{\textwidth}{!}{%
\begin{tabular}{c|cccc|cccccc}
\hline
\textbf{Global Context Encoder} & \textbf{AP} & \textbf{AP\(_S\)} & \textbf{AP\(_M\)} & \textbf{AP\(_L\)} & \textbf{dent} & \textbf{scratch} & \textbf{crack} & \textbf{glass} & \textbf{lamp} & \textbf{tire} \\ \hline
Simple CNN Encoder\citep{krizhevsky2012imagenet}              & \underline{64.6}    & \underline{39.7}          & \textbf{39.3}             & \textbf{66.2}             & \textbf{40.1}     & \textbf{42.9}        & \underline{40.5}       & \underline{93.2}       & \underline{78.8}      & \underline{91.9}      \\
Residual CNN Encoder            & \textbf{64.8}       & \textbf{45.5}             & \underline{39.2}          & \underline{66.0}          & 37.0              & \underline{42.6}     & \textbf{42.2}          & \textbf{94.2}          & \textbf{80.2}         & \textbf{92.0}         \\
ViT Encoder\citep{dosovitskiy2021an}                     & 59.0                & 31.5                      & 36.8                      & 62.3                      & 35.1              & 37.2                 & 26.6                   & 88.5                   & 73.4                  & 88.2                  \\
Mamba Vision Encoder\citep{zhu2024visionmamba}            & 54.7                & 31.8                      & 34.1                      & 58.3                      & \underline{35.2}  & 38.1                 & 24.4                   & 85.1                   & 71.8                  & 84.2                  \\ \hline
\end{tabular}}
\end{table}

Third, the performance gap between different encoder architectures is particularly pronounced in challenging detection scenarios. Small object detection shows the most significant performance variation (AP$_S$: 31.5\% to 45.5\%), indicating that global context encoding is critical for fine-grained damage types requiring comprehensive scene understanding. Similarly, crack detection performance varies dramatically (24.4\% to 42.2\%), demonstrating that the quality of global context representation directly impacts the model's ability to disambiguate subtle damage patterns from visual artifacts.

The superior performance of the Residual CNN Encoder unequivocally validates our architectural choice. This design strikes an optimal balance between representational power and computational efficiency, proving highly effective for automotive damage detection. The encoder's strength lies in its combination of convolutional operations and residual connections, which enable the learning of increasingly abstract representations without sacrificing the fine-grained spatial information crucial for identifying damage across varying scales and types.

This hierarchical feature extraction process is visually substantiated by the feature map analysis presented in the Appendix (see Figure~\ref{fig:GCE-map}), which clearly illustrates how the network progressively isolates salient damage features from raw pixel input. The quantitative results from our ablation study further reinforce this, providing strong evidence that integrating a well-designed Global Context Encoder with residual connections is not just beneficial, but essential for achieving robust and accurate performance in complex automotive damage assessment scenarios.


\begin{table}[]
\centering
\caption{Model performance on VehiDE dataset with full class names.}
\label{tab:class-wise-ve-full}
\resizebox{\textwidth}{!}{%
\begin{tabular}{c|ccccccc}
\hline
\textbf{Model} & \textbf{Broken light} & \textbf{Lost part} & \textbf{Dent} & \textbf{Torn} & \textbf{Puncture} & \textbf{Scratch} & \textbf{Broken Glass} \\ \hline
DiffDet (baseline) & \textbf{33.5} & \textbf{52.0} & 15.0 & 21.4 & {28.0} & {21.0} & 62.0 \\
C-DiffDet+ & {31.9} & 50.5 & \textbf{15.1} & \textbf{23.3} & \textbf{31.27} & \textbf{22.1} & \textbf{63.1} \\ \hline
\end{tabular}%
}
\end{table}

\begin{table}[]
\begin{tabular}{c|ccc|ccc}
\hline
\textbf{Model}     & \textbf{AP$^{bb}$} & \textbf{AP$_{50}^{bb}$} & \textbf{AP$_{75}^{bb}$} & \textbf{AP\(_S\)} & \textbf{AP\(_M\)} & \textbf{AP\(_L\)} \\ \hline
DiffDet (baseline) & 33.3  & 54.0 & 33.1  & \textbf{6.5}   & 20.5    & 38.5              \\
C-DiffDet+         & \textbf{33.9}                                & \textbf{55.3}                                     & \textbf{34.1}                                     & \textbf{6.5}               & \textbf{20.8}     & \textbf{38.8}     \\ \hline
\end{tabular}
\label{Tab:AP-comparison-ve}
\end{table}

To evaluate the generalization capability of our proposed C-DiffDet+ framework, we conducted additional experiments on the VehiDE (Vehicle Damage Detection) dataset, a large-scale, publicly available collection created for segmenting and detecting automotive damage, particularly for car insurance applications, containing 13,945 high-resolution images of damaged cars, which include more than 32,000 annotated instances of damage. The dataset is categorized into eight common external damage classes: "dent", "scratch", "broken glass", "lost parts", "punctured", "torn", "broken lights", and the classification task directed class "non-damage". For machine learning purposes, the dataset is split into a training set with 11,621 images (83.33\%) and a validation set with 2,324 images (16.67\%), maintaining a consistent 8:2 instance ratio for each category across the partitions. VehiDE is designed to support various tasks, including classification, object detection, instance segmentation, and salient object detection \citep{Huynh2023VehiDE}.

C-DiffDet+ achieves an overall AP of 33.9\% on VehiDE, representing a 0.6\% improvement over the baseline DiffusionDet (33.3\%). This performance gain is particularly notable given the dataset's emphasis on subtle, low-visibility damage patterns that require sophisticated contextual reasoning. The improvement is most pronounced at higher IoU thresholds, with C-DiffDet+ achieving 34.1\% AP₇₅ compared to DiffusionDet's 33.1\%, indicating superior bounding box localization precision.

The contextual reasoning capabilities of our CAF mechanism prove especially valuable for the VehiDE dataset, where many damage instances are partially occluded or exhibit minimal visual contrast with surrounding vehicle surfaces. The global context encoding enables the model to infer damage presence based on contextual cues and spatial relationships, overcoming limitations of local feature analysis alone. This is evidenced by the model's improved performance on small object detection (APₛ), where it maintains competitive performance despite the dataset's additional challenges.

These results demonstrate that the benefits of our context-aware diffusion framework extend beyond the CarDD benchmark, validating the generalizability of our approach across different automotive damage detection scenarios. The consistent performance improvement on both datasets confirms that integrating global context representation with diffusion-based detection provides a robust solution for real-world vehicle inspection applications.

\begin{figure}[]
    \centering
    \includegraphics[width=0.5\linewidth]{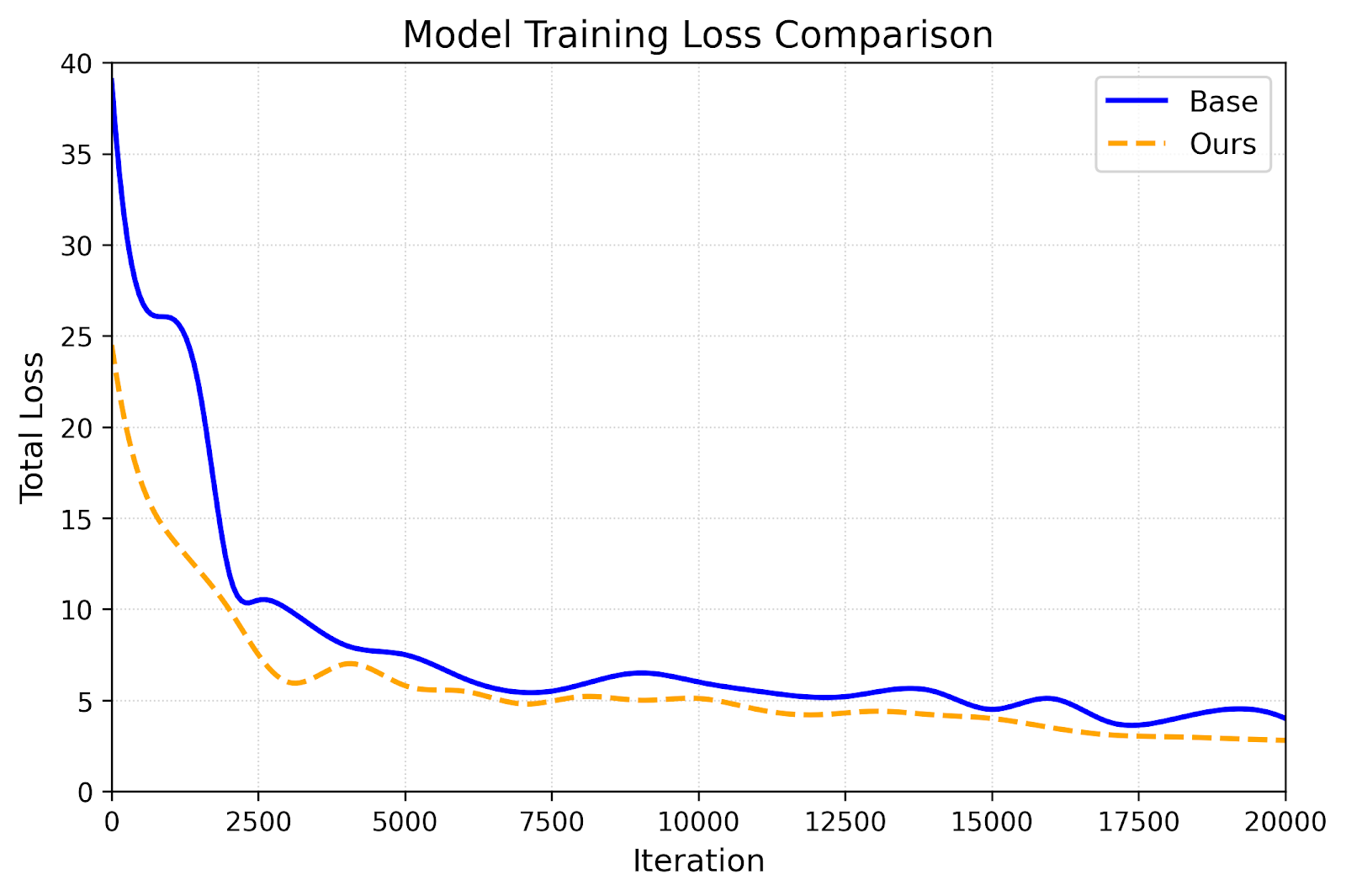}
    \caption{Training loss curves comparing baseline DiffusionDet (blue) and C-DiffDet+ (orange) over 20,000 iterations.}
    \label{fig:loss_convergence}
\end{figure}

The convergence behavior of the training loss, as depicted in Figure~\ref{fig:loss_convergence}, provides critical insights into the efficiency and stability of our proposed C-DiffDet+ architecture compared to the baseline DiffusionDet model \cite{Chen2022DiffusionDet}. The "Total Loss vs Iteration" graph demonstrates several key advantages of our context-aware approach that directly correlate with the improved detection performance shown in our quantitative evaluations.

\subsection{Training Loss Convergence Analysis}

Our C-DiffDet+ model exhibits significantly superior convergence characteristics compared to the baseline DiffusionDet. The model starts with a substantially lower initial loss, indicating that our architectural enhancements provide better initialization and more stable gradient flow from the outset. This improved convergence is attributable to the effective integration of the Global Context Encoder (GCE) and Context-Aware Fusion (CAF) mechanisms, which work together to enhance the model's learning dynamics.

The convergence trajectory of our model demonstrates both faster initial learning and a superior final converged state, achieving a demonstrably lower total loss than the baseline. While the baseline model undergoes erratic oscillations during training, our approach maintains a smoother, more stable learning curve, which is crucial for diffusion-based architectures. This enhanced convergence can be mathematically explained through the improved gradient flow enabled by our architectural design.

In the baseline DiffusionDet, predictions are generated solely from local region-of-interest (RoI) features $\mathbf{F}_{\text{roi}}$, with the gradient computed as:

\begin{equation}
\nabla_\theta \mathcal{L}_{\text{baseline}} = \nabla_\theta \mathcal{L}_{\text{total}}\left(y_i, f_{\theta}(\mathbf{F}_{\text{roi}})\right)
\label{eq:baseline_gradient}
\end{equation}

where $\mathcal{L}_{\text{total}} = \mathcal{L}_{\text{ce}} + \mathcal{L}_{\text{bbox}} + \mathcal{L}_{\text{giou}} + \mathcal{L}_{\text{feiou}}$ represents the combined loss function.

In contrast, our C-DiffDet+ model first enriches these local proposal features with global context through the Context-Aware Fusion mechanism. The GCE processes the input image $\mathbf{I}$ to extract scene-level information $\mathbf{g} \in \mathbb{R}^{B \times 256}$, which is then integrated with local features through cross-attention:

\begin{equation}
\mathbf{F}_{\text{context-aware}} = \text{CrossAttention}(\mathbf{F}_{\text{roi}}, \mathbf{g}, \mathbf{g})
\label{eq:caf_mechanism}
\end{equation}

The final prediction is generated from these context-aware features, resulting in a gradient computed on better-informed predictions:

\begin{equation}
\nabla_\theta \mathcal{L}_{\text{C-DiffDet+}} = \nabla_\theta \mathcal{L}_{\text{total}}\left(y_i, f_{\theta}(\mathbf{F}_{\text{context-aware}})\right)
\label{eq:enhanced_gradient}
\end{equation}

The key difference is that while the loss function $\mathcal{L}_{\text{total}}$ remains the same, our model's loss is computed on predictions derived from features enhanced with global context. This leads to more informative gradients during backpropagation, as the global context provides additional supervision signals that help the model learn more robust feature representations. The GCE can be effectively optimized through the dedicated gradient path that flows back from the loss function through the CAF mechanism, enabling it to learn optimal global context representations specifically for the detection task. This improved gradient flow creates a more stable learning process, as observed in our training curves, and directly contributes to the enhanced detection performance, making our approach not only more effective but also more practical for real-world deployment.

\section{Conclusion and Future Work}
\label{sec:Conclusion}
This work introduced C-DiffDet+, a novel framework that successfully bridges the gap between generative denoising processes and global contextual reasoning for the challenging task of fine-grained object detection. Our core contribution lies in the seamless integration of a dedicated Global Context Encoder (GCE) with a diffusion-based detection head via a principled Context-Aware Fusion (CAF) mechanism. This design explicitly addresses a fundamental shortcoming in existing methods: the reliance on locally conditioned features that are often ambiguous in complex real-world scenes like automotive damage assessment.

Extensive experimental validation on the CarDD benchmark demonstrates that our approach sets a new state-of-the-art, achieving a significant AP improvement over previous best models. More importantly, the gains are most pronounced on the most challenging fine-grained categories—cracks and scratches—where contextual disambiguation is crucial. The superior performance is complemented by a rigorous ablation study that quantitatively deconstructs the contribution of each component, confirming the synergistic effect of global context conditioning and adaptive feature enhancement.

\subsection{Limitations and Dataset Biases}
\label{subsec:limitations}

Despite achieving state-of-the-art performance, the proposed C-DiffDet+ framework is subject to certain limitations, some of which are inherent to the challenge of fine-grained detection and the datasets available for training.

\begin{figure*}[]

    \centering

    \begin{subfigure}[b]{0.32\textwidth}
        \centering
        \includegraphics[width=\textwidth]{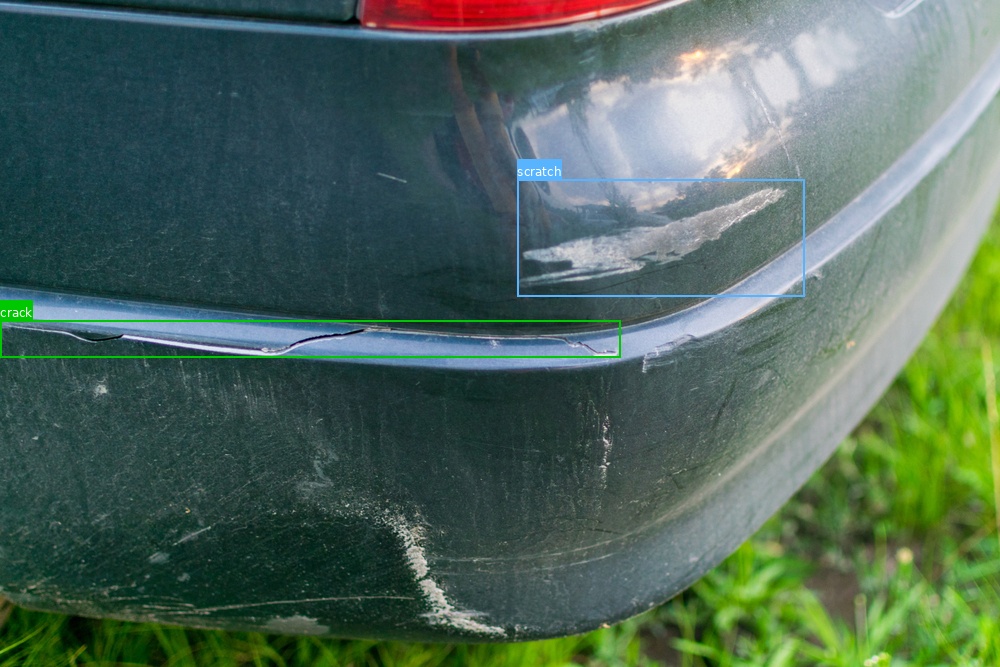}
        \captionsetup{labelformat=empty}
      
    \end{subfigure}
     \begin{subfigure}[b]{0.32\textwidth}
        \centering
        \includegraphics[width=\textwidth]{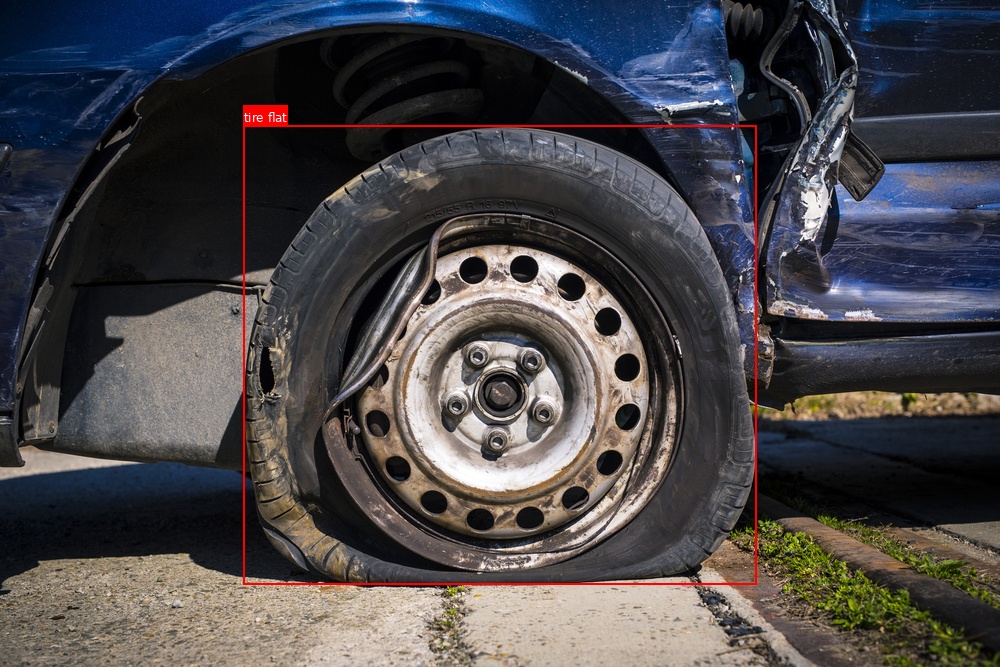}    \captionsetup{labelformat=empty}
    \end{subfigure}
    \begin{subfigure}[b]{0.32\textwidth}
        \centering
        \includegraphics[width=\textwidth]{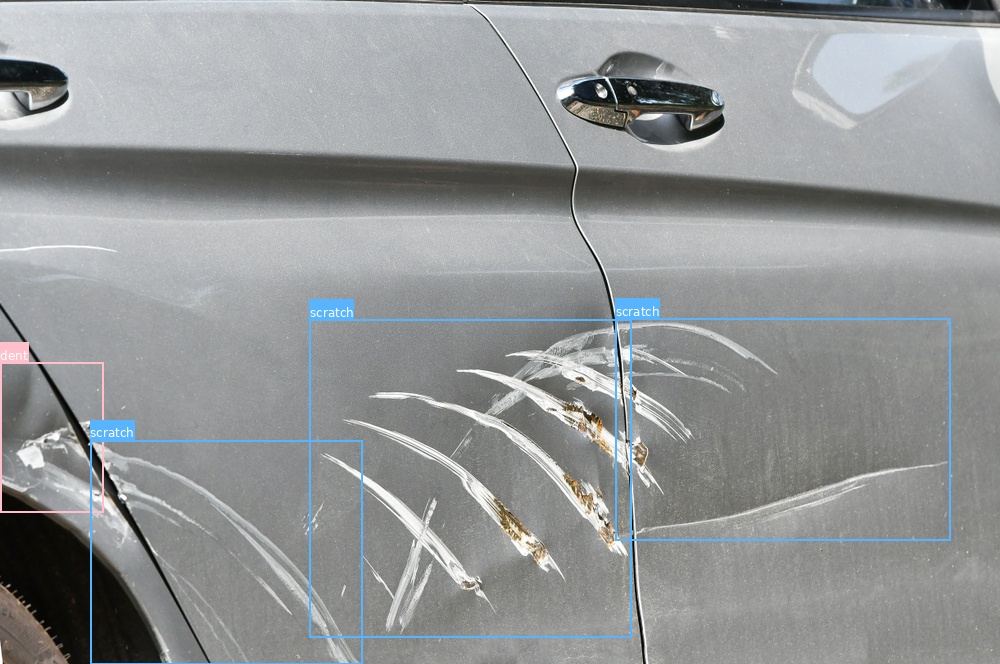}
        \captionsetup{labelformat=empty}
      
    \end{subfigure}
      
      

    \centering
    \begin{subfigure}[b]{0.32\textwidth}
        \centering
        \includegraphics[width=\textwidth]{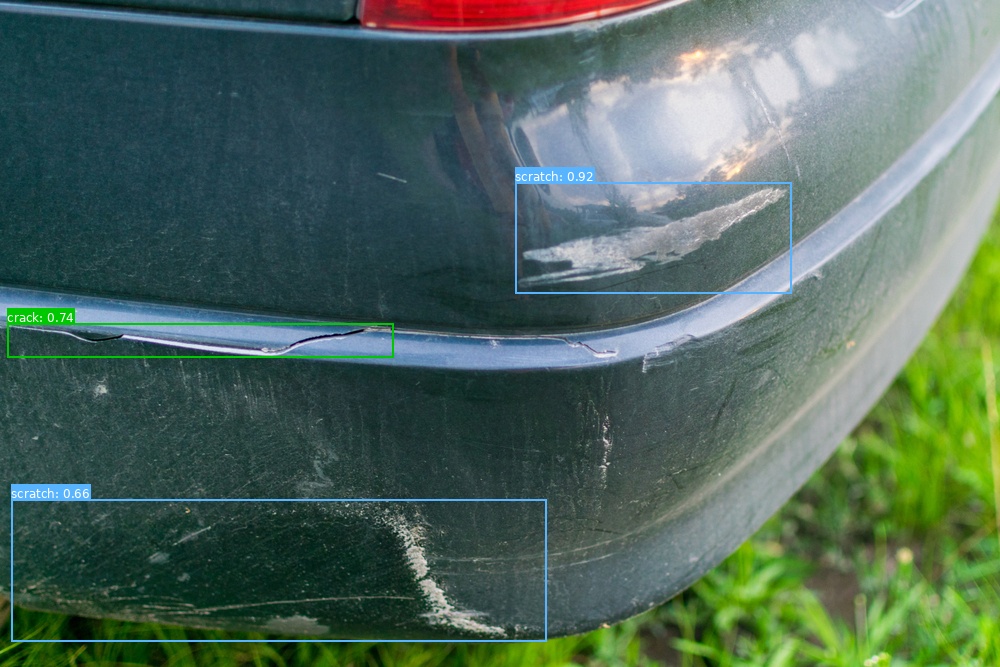}
        \captionsetup{labelformat=empty}
      
    \end{subfigure}
     \begin{subfigure}[b]{0.32\textwidth}
        \centering
        \includegraphics[width=\textwidth]{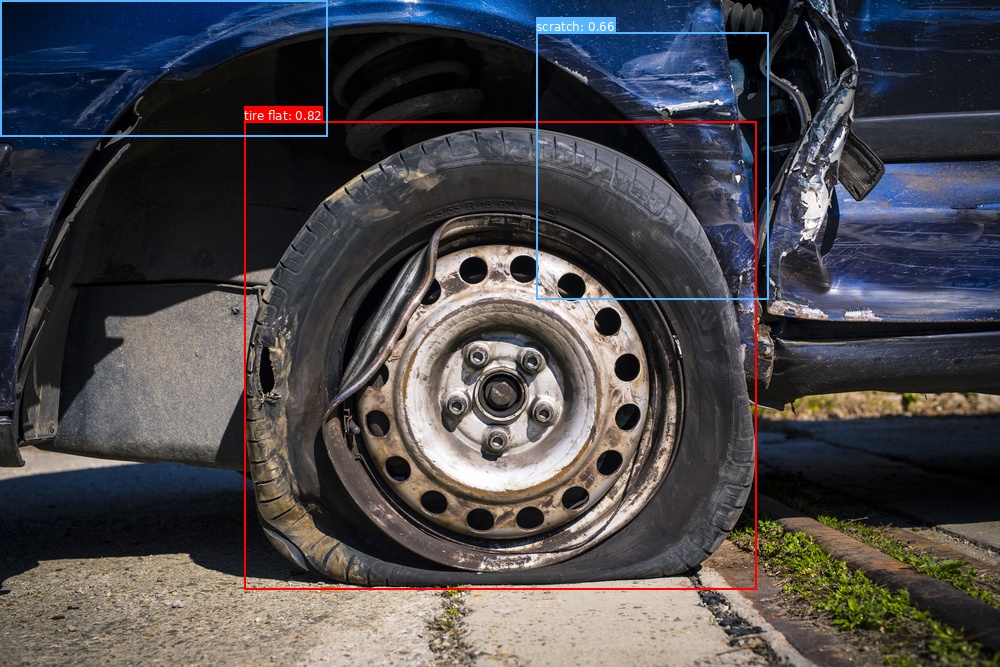}    \captionsetup{labelformat=empty}
    \end{subfigure}
    \begin{subfigure}[b]{0.32\textwidth}
        \centering
        \includegraphics[width=\textwidth]{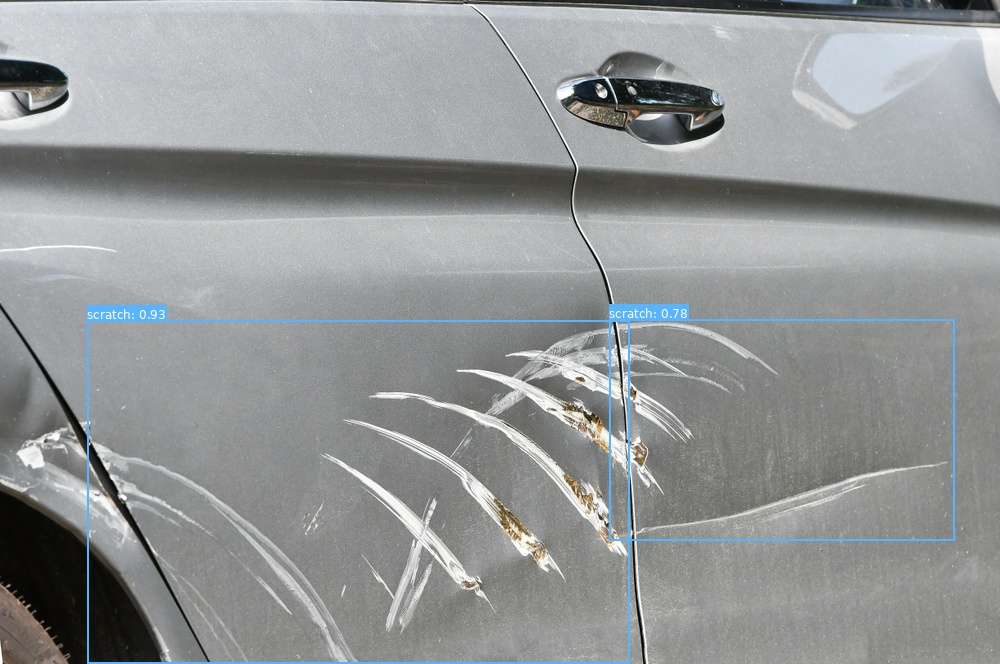}
        \captionsetup{labelformat=empty}
      
    \end{subfigure}
      
      
    \caption{
Qualitative comparison between ground truth annotations in the CarDD dataset and our model’s predictions, illustrating both detection errors and inconsistencies in the dataset annotations that limit precise evaluation. }
    \label{fig:qual_errors}
\end{figure*}

A primary limitation, observable in the qualitative results (Figure~\ref{fig:qual_errors}), is the model's occasional failure to precisely localize highly irregular or diffuse damage boundaries. As shown in the second row of Figure~\ref{fig:qual_errors}, our model may generate bounding boxes that are either \textit{overly conservative}, failing to capture the full extent of a large scratch or dent, or \textit{imprecisely localized}, slightly misaligning with the damage's true orientation. These errors often occur in cases of low-contrast damage or complex textures where the visual cues are ambiguous even to human annotators.

Furthermore, our analysis reveals that the evaluation process itself is sometimes hampered by inconsistencies in ground-truth annotations. The first row of Figure~\ref{fig:qual_errors} showcases examples where the provided ground-truth boxes are either incomplete or do not tightly conform to the damage's actual pixels. This annotation noise presents a fundamental challenge for training and fairly evaluating high-precision detection models, as it introduces a ceiling on achievable localization accuracy and can penalize models that predict more geometrically accurate boxes than the ground truth provides.

These observations point to two important future directions: 1) the development of more robust loss functions that are less sensitive to annotation noise and outliers, and 2) the exploration of segmentation-based or polygon-based representation for damage detection, which would be better suited for capturing the amorphous shapes of many damage types than axis-aligned bounding boxes.

\subsection*{Future Work}
The principles established here are broadly applicable. Our immediate future work will focus on two fronts:
1.  \textbf{Efficiency:} Exploring knowledge distillation and conditional computation techniques to distill the global contextual knowledge into a more lightweight architecture suitable for real-time deployment.
2.  \textbf{Generalization:} Extending the context-aware paradigm to other vision tasks where global scene understanding is critical but underexplored in generative models, such as scene graph generation or 3D object detection from monocular images. We also plan to investigate self-supervised pre-training strategies on large-scale unlabeled data to learn even more robust and generalizable contextual representations.

\appendix
\section{Computational Complexity}
\label{tab:model_complexity}

\begin{table}[]
\begin{tabular}{lcccc}
\hline
\textbf{Model} & \textbf{MACs (G)} & \textbf{Params (M)} & \textbf{GFLOPs} & \textbf{FPS (single GPU)} \\ \hline
DiffDet        & 375.292           & 169.110             & 750.58          & 11.12                     \\
C-DiffDet+     & 385.686           & 169.485             & 771.37          & 10.63                   \\ \hline
\end{tabular}
\end{table}

\section{Complete DiffusionDet Algorithm}

\subsection{Main Algorithm}
\begin{algorithm}[]
\caption{DiffusionDet: Context-Aware Diffusion-Based Object Detection}
\scriptsize 
\begin{algorithmic}[1]
\REQUIRE Input image $I \in \mathbb{R}^{B \times 3 \times H \times W}$, number of proposals $N = 300$, diffusion steps $T = 1000$
\ENSURE Detection results: bounding boxes $B_{out} \in \mathbb{R}^{B \times K \times 4}$, class labels $C_{out} \in \mathbb{R}^{B \times K \times C}$

\STATE \textbf{Stage 1: Feature Extraction}
\STATE Extract backbone features: $\{h^{(i)}\}_{i=1}^4 = \text{Backbone}(I)$
\STATE Apply ACE to Stage 4: $h^{(4)} = \text{ACE}(h^{(4)})$ \COMMENT{Adaptive Channel Enhancement}
\STATE Generate FPN features: $\{P_i\}_{i=1}^5 = \text{FPN}(\{h^{(i)}\}_{i=1}^4)$
\STATE Extract global context: $g = \text{GlobalContextEncoder}(I) \in \mathbb{R}^{B \times 256}$

\STATE \textbf{Stage 2: Proposal Initialization}
\IF{training mode}
    \STATE Initialize proposals: $X_0 = \text{GT\_Boxes} \in \mathbb{R}^{B \times K \times 4}$
\ELSE
    \STATE Initialize proposals: $X_0 = \text{Random\_Noise} \in \mathbb{R}^{B \times N \times 4}$
\ENDIF
\STATE Normalize proposals: $X_0 = \text{Normalize}(X_0, \text{scale} = 2.0)$

\STATE \textbf{Stage 3: Diffusion Process}
\FOR{$t = T-1$ \textbf{downto} $0$}
    \STATE \textbf{Stage 3.1: ROI Feature Extraction}
    \STATE Denormalize proposals: $X_{t}^{img} = \text{Denormalize}(X_t)$
    \STATE Assign FPN levels: $\text{levels} = \text{AssignFPNLevel}(X_{t}^{img})$
    \STATE Extract ROI features: $F_{roi} = \text{ROIAlign}(\{P_i\}_{i=1}^5, X_{t}^{img}, \text{levels})$
    \STATE Aggregate features: $F_{roi} = \text{GlobalAvgPool}(F_{roi}) \in \mathbb{R}^{B \times N \times 256}$
    
    \STATE \textbf{Stage 3.2: Multi-Modal Attention}
    \STATE Self-attention: $F_{self} = \text{MultiHead}(F_{roi}, F_{roi}, F_{roi})$
    \STATE Cross-attention: $F_{cross} = \text{MultiHead}(F_{self}, g, g)$ \COMMENT{Context-Aware Fusion}
    \STATE Instance interaction: $F_{enhanced} = \text{MultiHead}(F_{cross}, F_{roi})$
    
    \STATE \textbf{Stage 3.3: Multi-Modal Conditioning}
    \STATE Time embedding: $E_t = \text{TimeEmbedding}(t) \in \mathbb{R}^{B \times 256}$
    \STATE Positional encoding: $PE = \text{PositionalEncoding}(N) \in \mathbb{R}^{N \times 256}$
    \STATE Context embedding: $C_{emb} = \text{MLP}(g) \in \mathbb{R}^{B \times 256}$
    \STATE Fuse embeddings: $F_{fused} = F_{enhanced} + E_t + PE + C_{emb}$
    
    \STATE \textbf{Stage 3.4: Final MLP Projection}
    \STATE Apply final MLP: $F_{final} = \text{MLP}_{\text{final}}(F_{fused}) \in \mathbb{R}^{B \times N \times 256}$
    
    \STATE \textbf{Stage 3.5: Prediction Heads}
    \STATE Classification: $p_{cls} = \sigma(\text{MLP}_{cls}(F_{final})) \in \mathbb{R}^{B \times N \times C}$
    \STATE Box regression: $\Delta x = \text{MLP}_{reg}(F_{final}) \in \mathbb{R}^{B \times N \times 4}$
    \STATE Noise prediction: $\hat{\epsilon}_\theta = \text{MLP}_{noise}(F_{final}) \in \mathbb{R}^{B \times N \times 4}$
    
    \STATE \textbf{Stage 3.6: DDIM Sampling}
    \STATE Predict clean boxes: $\hat{x}_0 = X_t - \sqrt{1 - \bar{\alpha}_t} \cdot \hat{\epsilon}_\theta$
    \STATE Update proposals: $X_{t-1} = \sqrt{\bar{\alpha}_{t-1}} \cdot \hat{x}_0 + \sqrt{1 - \bar{\alpha}_{t-1}} \cdot \hat{\epsilon}_\theta$
    
    \STATE \textbf{Stage 3.7: Box Renewal}
    \IF{$t > 0$ \textbf{and} $t \% \text{renewal\_step} == 0$}
        \STATE Filter high-confidence proposals: $\text{keep\_mask} = \max(p_{cls}) > 0.5$
        \STATE Count kept proposals: $K_{kept} = \sum(\text{keep\_mask})$
        \IF{$K_{kept} < N$}
            \STATE Generate new proposals: $X_{new} = \text{Random\_Noise}(N - K_{kept})$
            \STATE Replace low-confidence proposals: $X_{t-1}[\neg\text{keep\_mask}] = X_{new}$
        \ENDIF
    \ENDIF
\ENDFOR

\STATE \textbf{Stage 4: Post-Processing}
\STATE Final denormalization: $B_{final} = \text{Denormalize}(X_0)$
\STATE Apply NMS: $B_{out}, C_{out} = \text{NMS}(B_{final}, p_{cls})$
\RETURN $B_{out}, C_{out}$

\end{algorithmic}
\end{algorithm}

\subsection{Global Context Encoder (GCE) Algorithm}
\begin{algorithm}[]
\caption{Global Context Encoder for Scene-Level Understanding}
\label{alg:gce}
\begin{algorithmic}[1]
\REQUIRE Input image $I \in \mathbb{R}^{B \times 3 \times H \times W}$, target dimension $C_{out} = 256$
\ENSURE Global context vector $g \in \mathbb{R}^{B \times C_{out}}$

\STATE \textbf{Step 1: Multi-Scale Feature Extraction}
\STATE Conv1: $F_1 = \text{ReLU}(\text{Conv2d}(3 \rightarrow 64, k=3, s=2, p=1)(I)) \in \mathbb{R}^{B \times 64 \times H/2 \times W/2}$
\STATE Conv2: $F_2 = \text{ReLU}(\text{Conv2d}(64 \rightarrow 128, k=3, s=2, p=1)(F_1)) \in \mathbb{R}^{B \times 128 \times H/4 \times W/4}$
\STATE Conv3: $F_3 = \text{ReLU}(\text{Conv2d}(128 \rightarrow C_{out}, k=3, s=2, p=1)(F_2)) \in \mathbb{R}^{B \times C_{out} \times H/8 \times W/8}$

\STATE \textbf{Step 2: Residual Connection}
\STATE Downsample input: $I_{ds} = \text{AvgPool2d}(k=3, s=2, p=1)(I)$
\STATE Further downsample: $I_{ds2} = \text{AvgPool2d}(k=3, s=2, p=1)(I_{ds})$
\STATE Final downsample: $I_{ds3} = \text{AvgPool2d}(k=3, s=2, p=1)(I_{ds2})$
\STATE Project residual: $I_{res} = \text{Conv2d}(3 \rightarrow C_{out}, k=1)(I_{ds3}) \in \mathbb{R}^{B \times C_{out} \times H/8 \times W/8}$

\STATE \textbf{Step 3: Feature Fusion}
\STATE Add residual: $F_{fused} = F_3 + I_{res} \in \mathbb{R}^{B \times C_{out} \times H/8 \times W/8}$

\STATE \textbf{Step 4: Global Context Aggregation}
\STATE Global average pooling: $g_{spatial} = \text{AdaptiveAvgPool2d}(1,1)(F_{fused}) \in \mathbb{R}^{B \times C_{out} \times 1 \times 1}$
\STATE Flatten: $g = \text{view}(g_{spatial}, [B, -1]) \in \mathbb{R}^{B \times C_{out}}$

\RETURN $g$

\end{algorithmic}
\end{algorithm}

\subsection{Multi-Modal Fusion (MMF) Algorithm}
\begin{algorithm}[]
\caption{Multi-Modal Feature Fusion and Conditioning}
\begin{algorithmic}[1]
\REQUIRE Enhanced ROI features $F_{enhanced} \in \mathbb{R}^{B \times N \times 256}$, global context $g \in \mathbb{R}^{B \times 256}$, timestep $t$, number of proposals $N$, number of classes $C$
\ENSURE Multi-modal fused features $F_{fused} \in \mathbb{R}^{B \times N \times 256}$

\STATE \textbf{Step 1: Time Embedding Generation}
\STATE Sinusoidal time encoding: $E_t = \text{TimeEmbedding}(t) \in \mathbb{R}^{B \times 256}$
\STATE Time projection: $E_t = \text{MLP}_{time}(E_t) \in \mathbb{R}^{B \times 256}$
\STATE Broadcast time: $E_t = \text{expand}(E_t, [B, N, 256]) \in \mathbb{R}^{B \times N \times 256}$

\STATE \textbf{Step 2: Positional Encoding}
\STATE Generate position indices: $\text{pos} = [0, 1, 2, \ldots, N-1] \in \mathbb{R}^{N}$
\STATE Sinusoidal positional encoding: $PE = \text{PositionalEncoding}(\text{pos}) \in \mathbb{R}^{N \times 256}$
\STATE Broadcast positions: $PE = \text{expand}(PE, [B, N, 256]) \in \mathbb{R}^{B \times N \times 256}$

\STATE \textbf{Step 3: Context Embedding}
\STATE Context projection: $C_{emb} = \text{MLP}_{context}(g) \in \mathbb{R}^{B \times 256}$
\STATE Broadcast context: $C_{emb} = \text{expand}(C_{emb}, [B, N, 256]) \in \mathbb{R}^{B \times N \times 256}$

\STATE \textbf{Step 4: Multi-Modal Feature Fusion}
\STATE Additive fusion: $F_{fused} = F_{enhanced} + E_t + PE + C_{emb}$
\STATE Layer normalization: $F_{fused} = \text{LayerNorm}(F_{fused})$

\STATE \textbf{Step 5: Adaptive Feature Modulation}
\STATE Compute modulation weights: $\alpha = \sigma(\text{MLP}_{mod}(g)) \in \mathbb{R}^{B \times 1 \times 1}$
\STATE Context modulation: $F_{context} = \alpha \odot C_{emb}$
\STATE Final fusion: $F_{fused} = F_{fused} + F_{context}$

\RETURN $F_{fused}$

\end{algorithmic}
\end{algorithm}

\subsection{Adaptive Channel Enhancement (ACE) Algorithm}
\begin{algorithm}[H]
\caption{Adaptive Channel Enhancement (ACE)}
\begin{algorithmic}[1]
\REQUIRE Input features $x \in \mathbb{R}^{B \times L \times C}$, reduction ratio $r = 16$
\ENSURE Enhanced features $x_{out} \in \mathbb{R}^{B \times L \times C}$

\STATE \textbf{Step 1: Spatial Aggregation (Squeeze)}
\STATE Permute dimensions: $x' = \text{permute}(x, [0, 2, 1]) \in \mathbb{R}^{B \times C \times L}$
\STATE Global average pooling: $\mu = \text{GlobalAvgPool}(x') \in \mathbb{R}^{B \times C}$

\STATE \textbf{Step 2: Channel Learning (Excitation)}
\STATE Dimension reduction: $y_1 = \text{ReLU}(\text{FC}_1(\mu)) \in \mathbb{R}^{B \times C/r}$
\STATE Dimension restoration: $y_2 = \sigma(\text{FC}_2(y_1)) \in \mathbb{R}^{B \times C}$
\STATE Reshape attention weights: $\alpha = y_2 \in \mathbb{R}^{B \times C \times 1}$

\STATE \textbf{Step 3: Feature Enhancement}
\STATE Broadcast weights: $\alpha' = \text{permute}(\alpha, [0, 2, 1]) \in \mathbb{R}^{B \times 1 \times C}$
\STATE Apply attention: $x_{out} = x \odot \alpha'$
\RETURN $x_{out}$

\end{algorithmic}
\end{algorithm}

\subsection{Context-Aware Fusion (CAF) Algorithm}
\begin{algorithm}[]
\caption{Context-Aware Fusion through Cross-Attention}
\begin{algorithmic}[1]
\REQUIRE ROI features $F_{roi} \in \mathbb{R}^{B \times N \times 256}$, global context $g \in \mathbb{R}^{B \times 256}$, number of heads $h = 8$, head dimension $d_k = 32$
\ENSURE Context-aware features $F_{cross} \in \mathbb{R}^{B \times N \times 256}$

\STATE \textbf{Step 1: Query, Key, Value Projections}
\STATE Query projection: $Q = F_{roi} \cdot W_Q \in \mathbb{R}^{B \times N \times 256}$
\STATE Key projection: $K = g \cdot W_K \in \mathbb{R}^{B \times 1 \times 256}$
\STATE Value projection: $V = g \cdot W_V \in \mathbb{R}^{B \times 1 \times 256}$

\STATE \textbf{Step 2: Multi-Head Attention}
\STATE Split into heads: $Q_h, K_h, V_h = \text{SplitHeads}(Q, K, V, h, d_k)$
\FOR{$i = 1$ \textbf{to} $h$}
    \STATE Compute attention scores: $\text{scores}_i = \frac{Q_h^i \cdot (K_h^i)^T}{\sqrt{d_k}} \in \mathbb{R}^{B \times N \times 1}$
    \STATE Apply softmax: $\text{weights}_i = \text{softmax}(\text{scores}_i) \in \mathbb{R}^{B \times N \times 1}$
    \STATE Weighted aggregation: $\text{head}_i = \text{weights}_i \cdot V_h^i \in \mathbb{R}^{B \times N \times d_k}$
\ENDFOR

\STATE \textbf{Step 3: Head Concatenation and Output}
\STATE Concatenate heads: $F_{heads} = \text{Concat}(\{\text{head}_i\}_{i=1}^h) \in \mathbb{R}^{B \times N \times 256}$
\STATE Output projection: $F_{cross} = F_{heads} \cdot W_O \in \mathbb{R}^{B \times N \times 256}$

\STATE \textbf{Step 4: Context Modulation}
\STATE Compute modulation weight: $\beta = \sigma(\text{MLP}(g)) \in \mathbb{R}^{B \times 1 \times 1}$
\STATE Gated fusion: $F_{modulated} = \beta \odot F_{cross} + (1 - \beta) \odot F_{roi}$
\RETURN $F_{modulated}$

\end{algorithmic}
\end{algorithm}

\subsection{FPN Level Assignment Algorithm}
\begin{algorithm}[H]
\caption{FPN Level Assignment for Multi-Scale ROI Processing}
\begin{algorithmic}[1]
\REQUIRE Proposal boxes $X \in \mathbb{R}^{B \times N \times 4}$ (in pixel coordinates), base size $s_{base} = 224$
\ENSURE FPN level assignments $\text{levels} \in \mathbb{R}^{B \times N}$

\STATE \textbf{Step 1: Box Size Calculation}
\FOR{each batch $b$ and proposal $n$}
    \STATE Extract box dimensions: $w = X[b, n, 2] - X[b, n, 0]$
    \STATE Extract box dimensions: $h = X[b, n, 3] - X[b, n, 1]$
    \STATE Calculate box area: $\text{area} = w \times h$
    \STATE Calculate box size: $\text{size} = \sqrt{\text{area}}$
\ENDFOR

\STATE \textbf{Step 2: Logarithmic Level Assignment}
\FOR{each batch $b$ and proposal $n$}
    \STATE Compute level: $\text{level} = \lfloor \log_2(\text{size}[b, n] / s_{base}) \rfloor + 4$
    \STATE Clamp to valid range: $\text{levels}[b, n] = \text{clamp}(\text{level}, 1, 5)$
\ENDFOR

\STATE \textbf{Step 3: Level Mapping}
\STATE Map levels to FPN features:
\STATE $P1$ (stride 2): $\text{levels} == 1$ \COMMENT{Very small objects}
\STATE $P2$ (stride 4): $\text{levels} == 2$ \COMMENT{Small objects}
\STATE $P3$ (stride 8): $\text{levels} == 3$ \COMMENT{Medium objects}
\STATE $P4$ (stride 16): $\text{levels} == 4$ \COMMENT{Large objects}
\STATE $P5$ (stride 32): $\text{levels} == 5$ \COMMENT{Very large objects}

\RETURN $\text{levels}$

\end{algorithmic}
\end{algorithm}

\section*{Analysis of Global Context Encoder Feature Maps}

\begin{figure}[h]
    \centering
    \includegraphics[width=1\linewidth]{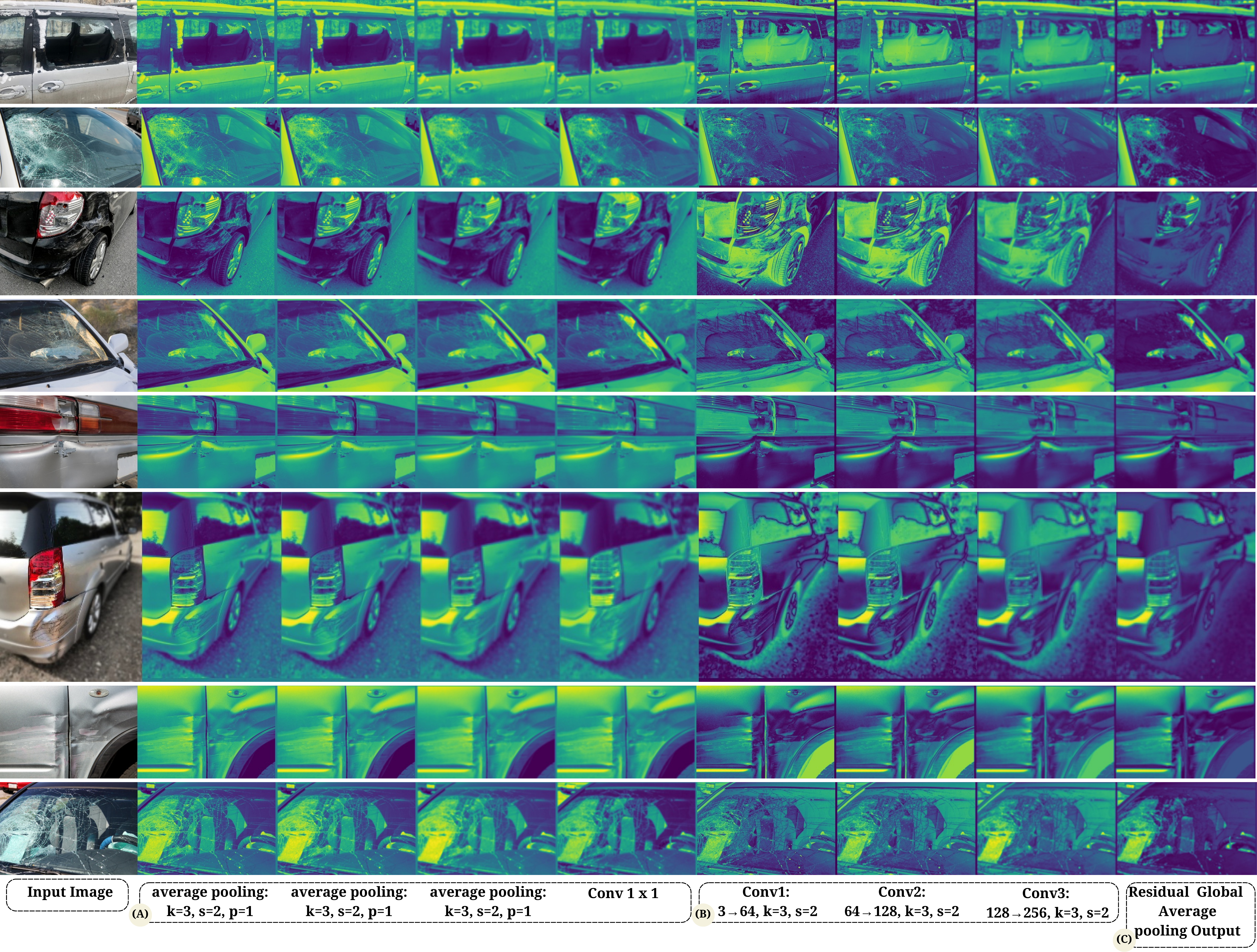}
    \caption{Visualization of feature maps at each stage within the Global Context Encoder (GCE)}
    \label{fig:GCE-map}
\end{figure}

Figure~\ref{fig:GCE-map} provides a comprehensive visualization of the feature representations learned at each processing stage within our Global Context Encoder. This analysis offers crucial insights into the hierarchical feature learning process that enables effective scene understanding.

\subsubsection*{Low-level Feature Extraction (Conv1)}
The initial convolutional layer produces feature maps that capture the fundamental visual elements essential for damage detection. These include edge and contour responses along damage boundaries, texture patterns that distinguish damaged from intact surfaces, and color gradient information critical for identifying scratches and paint damage. These basic shape patterns serve as the foundational building blocks for higher-level processing.

\subsubsection*{Intermediate Feature Integration (Conv2)}
The second convolutional layer demonstrates progressive feature integration, combining low-level elements into more complex patterns. This stage shows emerging activation around potential damage regions and begins the initial suppression of irrelevant background information. Through this feature composition, the layer starts to form prototype damage representations.

\subsubsection*{High-level Semantic Encoding (Conv3)}
The final convolutional layer produces semantically rich feature maps characterized by strong, focused activations that are precisely localized to damage areas. These maps provide a clear differentiation between damage types like scratches, dents, and cracks. This layer also effectively suppresses visual distractions such as reflections and shadows while encoding the contextual relationships between the damage and surrounding vehicle parts.

\subsubsection*{Residual Pathway Processing (A)}
The residual pathway is crucial for preserving spatial information. It maintains fine-grained details through progressive downsampling, providing a spatial precision that is complementary to the main pathway's semantic encoding. This process ensures the positional accuracy needed for final damage localization.

\subsubsection*{Feature Fusion and Context Aggregation (B + C → Output)}
The final processing stage integrates information from both pathways. First, an \textbf{element-wise fusion} ($\mathbf{F}_{fused} = \mathbf{F}_3 + \mathbf{I}_{res}$) combines the high-level semantics with spatial precision. Next, \textbf{global average pooling} aggregates this spatial information into a compact context vector $\mathbf{g}$. The resulting final representation successfully captures both detailed damage information and the global scene context.

This hierarchical feature progression demonstrates the encoder's capability to transform raw pixel information into semantically meaningful representations that enable precise damage localization and effective resolution of ambiguous detection scenarios. The visualization confirms that our GCE successfully learns to focus on relevant damage patterns while incorporating broader contextual information essential for accurate automotive damage assessment.

\section*{Acknowledgement}
The authors thank Mr. Arturo Argentieri from CNR-ISASI Italy for his technical contribution to the multi-GPU computing facilities. This research was partially funded by the Italian Ministry of Enterprises and Made in Italy (MIMIT) with the funded project "Vehicle Check 360" Grant number F/340065/05/X59, CUP: B89J24003290005 and in part by the Italian Ministry of University and Research (MUR) with the project "Future Artificial Intelligence Research —FAIR"  Grant number PE0000013 CUP B53C22003630006.








\bibliographystyle{elsarticle-num}
\bibliography{cas-refs} 

\begin{thebibliography}{10}
\expandafter\ifx\csname url\endcsname\relax
  \def\url#1{\texttt{#1}}\fi
\expandafter\ifx\csname urlprefix\endcsname\relax\def\urlprefix{URL }\fi
\expandafter\ifx\csname href\endcsname\relax
  \def\href#1#2{#2} \def\path#1{#1}\fi

\bibitem{Girshick2014RCNN}
R.~Girshick, J.~Donahue, T.~Darrell, J.~Malik, Rich feature hierarchies for accurate object detection and semantic segmentation, in: Proc. IEEE CVPR, 2014.

\bibitem{Girshick2015FastRCNN}
R.~Girshick, Fast r-cnn, in: Proc. IEEE ICCV, 2015.

\bibitem{Ren2015FasterRCNN}
S.~Ren, K.~He, R.~Girshick, J.~Sun, Faster r-cnn: Towards real-time object detection with region proposal networks, in: Proc. NeurIPS / arXiv, 2015.

\bibitem{jiang2022review}
P.~Jiang, D.~Ergu, F.~Liu, Y.~Cai, B.~Ma, A review of yolo algorithm developments, Procedia computer science 199 (2022) 1066--1073.

\bibitem{Liu2016SSD}
W.~Liu, D.~Anguelov, D.~Erhan, C.~Szegedy, S.~Reed, C.-Y. Fu, A.~C. Berg, Ssd: Single shot multibox detector, in: Proc. ECCV, 2016.

\bibitem{fu2017dssd}
C.-Y. Fu, W.~Liu, A.~Ranga, A.~Tyagi, A.~C. Berg, Dssd: Deconvolutional single shot detector, arXiv preprint arXiv:1701.06659 (2017).

\bibitem{Lin2017FocalLoss}
T.-Y. Lin, P.~Goyal, R.~Girshick, K.~He, P.~Doll{\'a}r, Focal loss for dense object detection, in: Proceedings of the IEEE International Conference on Computer Vision (ICCV), 2017, pp. 2980--2988.

\bibitem{Redmon2016YOLO}
J.~Redmon, S.~Divvala, R.~Girshick, A.~Farhadi, You only look once: Unified, real-time object detection, in: Proc. IEEE CVPR, 2016.

\bibitem{Tan2021}
L.~Tan, T.~Huangfu, L.~Wu, W.~Chen, \href{https://doi.org/10.1186/s12911-021-01691-8}{Comparison of retinanet, ssd, and yolo v3 for real-time pill identification}, BMC Medical Informatics and Decision Making 21~(1) (2021) 324.
\newblock \href {https://doi.org/10.1186/s12911-021-01691-8} {\path{doi:10.1186/s12911-021-01691-8}}.
\newline\urlprefix\url{https://doi.org/10.1186/s12911-021-01691-8}

\bibitem{Law2018CornerNet}
H.~Law, J.~Deng, Cornernet: Detecting objects as paired keypoints, in: Proc. ECCV, 2018.

\bibitem{Zhou2019CenterNet}
X.~Zhou, D.~Wang, P.~Kr{\"a}henb{\"u}hl, Objects as points, in: Proc. arXiv / ICCV workshop, 2019.

\bibitem{Tian2019FCOS}
Z.~Tian, C.~Shen, H.~Chen, T.~He, Fcos: Fully convolutional one-stage object detection, in: Proc. ICCV, 2019.

\bibitem{Sun2021SparseRCNN}
P.~Sun, R.~Zhang, Y.~Cao, C.~Shen, X.~Jin, Z.~Shen, H.~Xiao, Sparse r-cnn: End-to-end object detection with learnable proposals, in: Proc. ECCV, 2021.

\bibitem{Carion2020DETR}
N.~Carion, F.~Massa, G.~Synnaeve, N.~Usunier, A.~Kirillov, S.~Zagoruyko, End-to-end object detection with transformers, in: Proc. ECCV, 2020.

\bibitem{Ho2020DDPM}
J.~Ho, A.~Jain, P.~Abbeel, Denoising diffusion probabilistic models, NeurIPS (2020).

\bibitem{Nichol2021ImprovedDDPM}
A.~Nichol, P.~Dhariwal, Improved denoising diffusion probabilistic models, arXiv preprint arXiv:2102.09672 (2021).

\bibitem{Song2019ScoreMatch}
Y.~Song, S.~Ermon, Generative modeling by estimating gradients of the data distribution, Proc. NeurIPS / arXiv (2019).

\bibitem{Chen2023DiffusionDet}
S.~Chen, P.~Sun, Y.~Song, P.~Luo, Diffusiondet: Diffusion model for object detection, arXiv preprint arXiv:2211.09788v2 (2023).

\bibitem{Wang2023CarDD}
X.~Wang, W.~Li, Z.~Wu, Car damage detection (cardd): A large-scale dataset for vision-based car damage assessment, IEEE Transactions on Intelligent Transportation Systems 24~(9) (2023) 1234--1247.

\bibitem{Dai2017DCN}
J.~Dai, H.~Qi, Y.~Xiong, Y.~Li, G.~Zhang, H.~Hu, Y.~Wei, Deformable convolutional networks, in: Proceedings of the IEEE International Conference on Computer Vision (ICCV), 2017, pp. 764--773.

\bibitem{Sellam2023Sensors}
A.~Z. Sellam, A.~Benlamoudi, C.~A. Cid, L.~Dobelle, A.~Slama, Y.~El~Hillali, A.~Taleb-Ahmed, \href{https://doi.org/10.3390/s23041794}{Deep learning solution for quantification of fluorescence particles on a membrane}, Sensors 23~(4) (2023) 1794.
\newblock \href {https://doi.org/10.3390/s23041794} {\path{doi:10.3390/s23041794}}.
\newline\urlprefix\url{https://doi.org/10.3390/s23041794}

\bibitem{Lin2017Focal}
T.-Y. Lin, P.~Goyal, R.~Girshick, K.~He, P.~Doll\'{a}r, Focal loss for dense object detection, in: Proc. IEEE ICCV, 2017.

\bibitem{Cao2019GCNet}
Y.~Cao, J.~Xu, S.~Lin, F.~Wei, H.~Hu, {GCNet}: Non-local networks meet squeeze-excitation networks and beyond, in: Proceedings of the IEEE/CVF International Conference on Computer Vision Workshop (ICCVW), 2019, pp. 1971--1980.

\bibitem{Tzeng2017DomainAdap}
E.~Tzeng, J.~Hoffman, K.~Saenko, T.~Darrell, Adversarial discriminative domain adaptation, in: Proceedings of the IEEE Conference on Computer Vision and Pattern Recognition (CVPR), 2017, pp. 7167--7176.

\bibitem{Wilson2022SurveyDomainAdap}
G.~Wilson, D.~Cook, A survey of unsupervised domain adaptation for visual recognition, ACM Computing Surveys 55~(6) (2022) 1--37.
\newblock \href {https://doi.org/10.1145/3544792} {\path{doi:10.1145/3544792}}.

\bibitem{nichol2021improved}
A.~Nichol, P.~Dhariwal, Improved denoising diffusion probabilistic models, arXiv preprint arXiv:2102.09672 (2021).

\bibitem{lin2017feature}
T.-Y. Lin, P.~Doll{\'a}r, R.~Girshick, K.~He, B.~Hariharan, S.~Belongie, Feature pyramid networks for object detection, in: Proceedings of the IEEE conference on computer vision and pattern recognition (CVPR), 2017, pp. 2117--2125.

\bibitem{Carion2020EndToEndOD}
N.~Carion, F.~Massa, G.~Synnaeve, N.~Usunier, A.~Kirillov, S.~Zagoruyko, End-to-end object detection with transformers, in: A.~Vedaldi, H.~Bischof, T.~Brox, J.-M. Frahm (Eds.), Computer Vision -- ECCV 2020, Springer International Publishing, Cham, 2020, pp. 213--229.

\bibitem{Sun_2021_CVPR}
P.~Sun, R.~Zhang, Y.~Jiang, T.~Kong, C.~Xu, W.~Zhan, M.~Tomizuka, L.~Li, Z.~Yuan, C.~Wang, P.~Luo, Sparse r-cnn: End-to-end object detection with learnable proposals, in: Proceedings of the IEEE/CVF Conference on Computer Vision and Pattern Recognition (CVPR), 2021, pp. 14454--14463.

\bibitem{zhu2021DeformableDETR}
X.~Zhu, W.~Su, L.~Lu, B.~Li, X.~Wang, J.~Dai, \href{https://arxiv.org/abs/2010.04159}{Deformable detr: Deformable transformers for end-to-end object detection} (2021).
\newblock \href {http://arxiv.org/abs/2010.04159} {\path{arXiv:2010.04159}}.
\newline\urlprefix\url{https://arxiv.org/abs/2010.04159}

\bibitem{CarDD2023Wang}
X.~Wang, W.~Li, Z.~Wu, Cardd: A new dataset for vision-based car damage detection, IEEE Transactions on Intelligent Transportation Systems 24~(7) (2023) 7202--7214.
\newblock \href {https://doi.org/10.1109/TITS.2023.3258480} {\path{doi:10.1109/TITS.2023.3258480}}.

\bibitem{Fang_2024DataAug}
H.~Fang, B.~Han, S.~Zhang, S.~Zhou, C.~Hu, W.-M. Ye, Data augmentation for object detection via controllable diffusion models, in: Proceedings of the IEEE/CVF Winter Conference on Applications of Computer Vision (WACV), 2024, pp. 1257--1266.

\bibitem{Benaissa2025Gated}
I.~Benaissa, A.~Zitouni, S.~Sbaa, N.~Aydin, A.~C. Megherbi, A.~Z. Sellam, A.~Taleb-Ahmed, \href{https://ssrn.com/abstract=5281763}{Gated attention augmented double unet for white blood cell segmentation} (2025).
\newblock \href {https://doi.org/10.2139/ssrn.5281763} {\path{doi:10.2139/ssrn.5281763}}.
\newline\urlprefix\url{https://ssrn.com/abstract=5281763}

\bibitem{zidi2025lola}
F.~A. Zidi, D.~E. Boukhari, A.~Z. Sellam, A.~Ouafi, C.~Distante, S.~E. Bekhouche, A.~Taleb-Ahmed, \href{https://arxiv.org/abs/2506.17759}{Lola-specvit: Local attention swiglu vision transformer with lora for hyperspectral imaging} (2025).
\newblock \href {http://arxiv.org/abs/2506.17759} {\path{arXiv:2506.17759}}.
\newline\urlprefix\url{https://arxiv.org/abs/2506.17759}

\bibitem{wu2019detectron2}
Y.~Wu, A.~Kirillov, F.~Massa, W.-Y. Lo, R.~Girshick, Detectron2, \url{https://github.com/facebookresearch/detectron2} (2019).

\bibitem{Deng2009ImageNet}
J.~Deng, W.~Dong, R.~Socher, L.-J. Li, K.~Li, L.~Fei-Fei, Imagenet: A large-scale hierarchical image database, in: 2009 IEEE Conference on Computer Vision and Pattern Recognition, 2009, pp. 248--255.
\newblock \href {https://doi.org/10.1109/CVPR.2009.5206848} {\path{doi:10.1109/CVPR.2009.5206848}}.

\bibitem{Chen2022DiffusionDet}
S.~Chen, P.~Gu, J.~Zhang, X.~He, J.~He, Q.~Tian, Y.~Zhou, Y.~Qiao, {DiffusionDet}: Diffusion model for object detection, in: Proceedings of the 39th International Conference on Machine Learning (ICML), Vol. 162 of Proceedings of Machine Learning Research, 2022, pp. 3231--3245.

\bibitem{loshchilov2019AdamW}
I.~Loshchilov, F.~Hutter, \href{https://arxiv.org/abs/1711.05101}{Decoupled weight decay regularization} (2019).
\newblock \href {http://arxiv.org/abs/1711.05101} {\path{arXiv:1711.05101}}.
\newline\urlprefix\url{https://arxiv.org/abs/1711.05101}

\bibitem{Beitzel2009mAP}
S.~M. Beitzel, E.~C. Jensen, O.~Frieder, \href{https://doi.org/10.1007/978-0-387-39940-9_492}{MAP}, Springer US, Boston, MA, 2009, pp. 1691--1692.
\newblock \href {https://doi.org/10.1007/978-0-387-39940-9_492} {\path{doi:10.1007/978-0-387-39940-9_492}}.
\newline\urlprefix\url{https://doi.org/10.1007/978-0-387-39940-9_492}

\bibitem{MaskRCNN2017He}
K.~He, G.~Gkioxari, P.~Dollár, R.~Girshick, Mask r-cnn, in: 2017 IEEE International Conference on Computer Vision (ICCV), 2017, pp. 2980--2988.
\newblock \href {https://doi.org/10.1109/ICCV.2017.322} {\path{doi:10.1109/ICCV.2017.322}}.

\bibitem{CascadeRCNN2021}
Z.~Cai, N.~Vasconcelos, Cascade r-cnn: High quality object detection and instance segmentation, IEEE Transactions on Pattern Analysis and Machine Intelligence 43~(5) (2021) 1483--1498.
\newblock \href {https://doi.org/10.1109/TPAMI.2019.2956516} {\path{doi:10.1109/TPAMI.2019.2956516}}.

\bibitem{GCNET2019}
Y.~Cao, J.~Xu, S.~Lin, F.~Wei, H.~Hu, Gcnet: Non-local networks meet squeeze-excitation networks and beyond, in: 2019 IEEE/CVF International Conference on Computer Vision Workshop (ICCVW), 2019, pp. 1971--1980.
\newblock \href {https://doi.org/10.1109/ICCVW.2019.00246} {\path{doi:10.1109/ICCVW.2019.00246}}.

\bibitem{HTC2019}
K.~Chen, J.~Pang, J.~Wang, Y.~Xiong, X.~Li, S.~Sun, W.~Feng, Z.~Liu, J.~Shi, W.~Ouyang, C.~C. Loy, D.~Lin, Hybrid task cascade for instance segmentation, in: 2019 IEEE/CVF Conference on Computer Vision and Pattern Recognition (CVPR), 2019, pp. 4969--4978.
\newblock \href {https://doi.org/10.1109/CVPR.2019.00511} {\path{doi:10.1109/CVPR.2019.00511}}.

\bibitem{DCN2017}
J.~Dai, H.~Qi, Y.~Xiong, Y.~Li, G.~Zhang, H.~Hu, Y.~Wei, Deformable convolutional networks, in: 2017 IEEE International Conference on Computer Vision (ICCV), 2017, pp. 764--773.
\newblock \href {https://doi.org/10.1109/ICCV.2017.89} {\path{doi:10.1109/ICCV.2017.89}}.

\bibitem{krizhevsky2012imagenet}
A.~Krizhevsky, I.~Sutskever, G.~E. Hinton, \href{https://proceedings.neurips.cc/paper/2012/hash/c399862d3b9d6b76c8436e924a68c45b-Abstract.html}{Imagenet classification with deep convolutional neural networks}, in: Advances in Neural Information Processing Systems (NeurIPS), Vol.~25, 2012, pp. 1097--1105.
\newline\urlprefix\url{https://proceedings.neurips.cc/paper/2012/hash/c399862d3b9d6b76c8436e924a68c45b-Abstract.html}

\bibitem{dosovitskiy2021an}
A.~Dosovitskiy, L.~Beyer, A.~Kolesnikov, D.~Weissenborn, X.~Zhai, T.~Unterthiner, M.~Dehghani, M.~Minderer, G.~Heigold, S.~Gelly, J.~Uszkoreit, N.~Houlsby, \href{https://openreview.net/forum?id=YicbFdNTTy}{An image is worth 16x16 words: Transformers for image recognition at scale}, in: International Conference on Learning Representations (ICLR), 2021.
\newline\urlprefix\url{https://openreview.net/forum?id=YicbFdNTTy}

\bibitem{zhu2024visionmamba}
L.~Zhu, B.~Liao, Q.~Zhang, X.~Wang, W.~Liu, X.~Wang, \href{https://arxiv.org/abs/2401.09417}{Vision mamba: Efficient visual representation learning with bidirectional state space model}, arXiv preprint arXiv:2401.09417 (2024).
\newline\urlprefix\url{https://arxiv.org/abs/2401.09417}

\bibitem{Huynh2023VehiDE}
N.~T. ~, N.~N. Tran, A.~T. Huynh, V.-D. Hoang, H.~D. Nguyen, Vehide dataset: New dataset for automatic vehicle damage detection in car insurance, in: 2023 15th International Conference on Knowledge and Systems Engineering (KSE), 2023, pp. 1--6.
\newblock \href {https://doi.org/10.1109/KSE59128.2023.10299490} {\path{doi:10.1109/KSE59128.2023.10299490}}.

\end{thebibliography}
\end{document}